\documentclass{article}

\PassOptionsToPackage{numbers, compress}{natbib}

\usepackage[preprint]{neurips_2026}

\usepackage[utf8]{inputenc} 
\usepackage[T1]{fontenc}    
\usepackage{hyperref}       
\usepackage{url}            
\usepackage{booktabs}       
\usepackage{amsfonts}       
\usepackage{nicefrac}       
\usepackage{microtype}      
\usepackage{xcolor}         
\usepackage{microtype}
\usepackage{url}
\usepackage{booktabs}

\usepackage{enumitem}
\newenvironment{itemize*}%
 {\leftmargini=20pt\begin{itemize}%
  \setlength{\itemsep}{3pt}%
  \setlength{\parskip}{0pt}%
  }%
 {\end{itemize}} 
\newenvironment{enumerate*}%
 {\begin{enumerate}%
  \setlength{\itemsep}{0pt}%
  \setlength{\parskip}{0pt}}%
 {\end{enumerate}}

\usepackage{amsmath}
\usepackage{cleveref}
\usepackage{listings}
\usepackage{titlesec}
\usepackage{multirow}
\usepackage{makecell}
\usepackage{xcolor}
\usepackage{wrapfig}
\usepackage{subcaption}
\usepackage{array}       
\usepackage[font=small]{caption}
\usepackage{amssymb}
\usepackage{dsfont}

\definecolor{midnightgreen}{rgb}{0.0, 0.29, 0.33}
\definecolor{deepgreen}{HTML}{0aa344}
\definecolor{deeppurple}{HTML}{7030a0}
\definecolor{deepblue}{HTML}{171d91}
\definecolor{brown}{HTML}{843c0c}
\definecolor{shadered}{HTML}{ffe5e5}
\definecolor{shadegreen}{HTML}{e5f7ed}
\definecolor{msftBlack}{RGB}{0,0,0}
\definecolor{lightred}{RGB}{255,163,163}
\definecolor{deepred}{RGB}{146,0,0}

\usepackage[tikz]{bclogo}

\NewDocumentCommand{\heng}
{ mO{} }{\textcolor{red}{\textsuperscript{\textit{Heng}}\textsf{\textbf{\small[#1]}}}}
\NewDocumentCommand{\cheng}
{ mO{} }{\textcolor{orange}{\textsuperscript{\textit{Cheng}}\textsf{\textbf{\small[#1]}}}}
\NewDocumentCommand{\ember}
{ mO{} }{\textcolor{purple}{\textsuperscript{\textit{Ember}}\textsf{\textbf{\small[#1]}}}}
\NewDocumentCommand{\hongru}
{ mO{} }{\textcolor{green}{\textsuperscript{\textit{Hongru}}\textsf{\textbf{\small[#1]}}}}
\NewDocumentCommand{\emre}
{ mO{} }{\textcolor{blue}{\textsuperscript{\textit{Emre}}\textsf{\textbf{\small[#1]}}}}
\NewDocumentCommand{\zhenhailong}
{ mO{} }{\textcolor{pink}{\textsuperscript{\textit{zhenhailong}}\textsf{\textbf{\small[#1]}}}}
\NewDocumentCommand{\kunlun}
{ mO{} }{\textcolor{yellow}{\textsuperscript{\textit{kunlun}}\textsf{\textbf{\small[#1]}}}}
\NewDocumentCommand{\chihan}
{ mO{} }{\textcolor{blue}{\textsuperscript{\textit{Chi}}\small[#1]}}  
\NewDocumentCommand{\jeongh}
{ mO{} }{\textcolor{brown}{\textsuperscript{\textit{Jeonghwan}}\textsf{\textbf{\small[#1]}}}}

\usepackage[normalem]{ulem}        

\usepackage{tcolorbox}
\tcbuselibrary{breakable}
\usepackage{listings}
\usepackage{threeparttable}
\usepackage{cleveref}

\definecolor{acadNavy}{RGB}{40,70,140}      
\definecolor{acadMustard}{RGB}{180,140,30} 

\definecolor{rqBlueBg}{HTML}{EAF4FF}
\definecolor{tzBlueHeader}{RGB}{78,160,205}
\definecolor{tzBlueHeader2}{RGB}{105,185,225}
\definecolor{tzBlueBorder}{RGB}{115,190,225}
\definecolor{tzBlueFill}{RGB}{232,246,252}
\definecolor{rqBlueBorder}{HTML}{6AADE4}

\DeclareTextCommand{\textquotedbl}{OT1}{\char`\"}

\lstdefinestyle{jsonTiny}{
  basicstyle=\ttfamily\scriptsize,
  breaklines=true,
  breakindent=0pt,
  columns=fullflexible,
  keepspaces=true,
  showstringspaces=false,
  upquote=true,
  frame=none
}

\tcbuselibrary{skins,breakable}

\newtcolorbox{block}[1][]{%
  enhanced,
  breakable,
  colback=white,
  colframe=black!85,
  boxrule=1.4pt,
  arc=7pt,
  left=4mm,right=4mm,top=4mm,bottom=3mm,
  before skip=10pt, after skip=10pt,
  #1
}

\usepackage{pifont}
\newcommand{\cmark}{\textcolor[rgb]{0.0, 0.6, 0.0}{\ding{51}}} 
\newcommand{\xmark}{\textcolor[rgb]{0.7, 0.0, 0.0}{\ding{55}}} 
\newcommand{\gmark}{\textcolor[rgb]{1,0.647,0}{\ding{51}}}

\usepackage{enumitem}
\renewenvironment{itemize*}
  {\begin{itemize}[leftmargin=20pt,itemsep=3pt,parsep=0pt,topsep=0pt,partopsep=0pt]}
  {\end{itemize}}

\tcbset{
  aibox/.style={
    width=\linewidth,
    top=7pt,
    bottom=2pt,
    colback=rqBlueBg,     
    colframe=black,
    colbacktitle=black,
    enhanced,
    center,
    attach boxed title to top left={yshift=-0.1in,xshift=0.15in},
    boxed title style={boxrule=0pt,colframe=white,},
  }
}
\newtcolorbox{AIbox}[2][]{aibox,title=#2,#1}
\newcounter{takeaway}
\newtcolorbox{takeaway}[1][]{
  aibox,
  colback=rqBlueBg,
  title={\stepcounter{takeaway}Takeaway \thetakeaway},
  #1
}

\title{%
  \begin{minipage}[c]{0.045\textwidth}
    \centering
    \vspace{-2mm}
    \includegraphics[width=1cm]{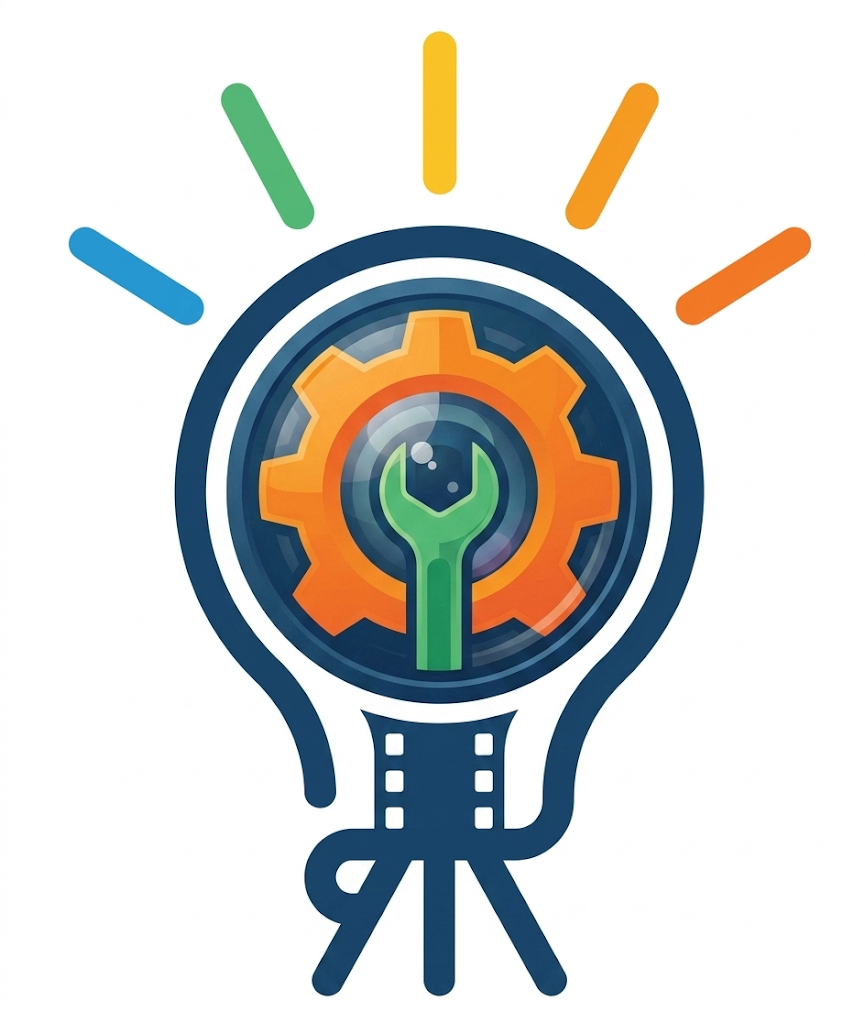}
  \end{minipage}\hfill
  \begin{minipage}[c]{0.915\textwidth}
    \raggedright
    Advancing Creative Physical Intelligence in Large Multimodal Models
  \end{minipage}
}

\author{
\noindent Cheng Qian$^*$\textsuperscript{$1$}, Hyeonjeong Ha$^*$\textsuperscript{$1$}, Jiayu Liu\textsuperscript{$1$}, Jeonghwan Kim\textsuperscript{$1$}, Emre Can Acikgoz\textsuperscript{$1$}, \\
\bfseries\ Bingxuan Li\textsuperscript{$1$}, Kunlun Zhu\textsuperscript{$1$}, \bfseries\ Jiateng Liu\textsuperscript{$1$}, \bfseries\ Aditi Tiwari\textsuperscript{$1$}, \bfseries\ Zhenhailong Wang\textsuperscript{$1$}, \\
\bfseries\ Xiusi Chen\textsuperscript{$1$}, \bfseries\ Mahdi Namazifar\textsuperscript{$2$}, \bfseries\ Heng Ji\textsuperscript{$1$} \vspace{2.5mm} \\
\textsuperscript{$1$}UIUC, \textsuperscript{$2$}Amazon
}

\begin{document}

\maketitle

\begin{abstract}
Large multimodal models (LMMs) have rapidly advanced in perception and reasoning; however, it remains unclear whether these capabilities generalize to discovering visually grounded solutions in open-ended environments, beyond pattern recognition. 
In such settings, intelligence requires more than answering well-posed questions: it involves identifying how elements in a scene can be repurposed in non-obvious yet physically feasible ways. 
This form of creative problem-solving is central to human intelligence, but remains largely untested in current benchmarks.
To evaluate this ability, we introduce \textbf{MM-CreativityBench}, a benchmark for affordance-grounded creative tool use in visually rich, physically constrained environments. 
Each instance presents a scenario image with structured views of candidate entities and their parts, enabling fine-grained, interactive evaluation of how models iteratively inspect the scene, identify relevant affordances, and compose visually and physically grounded solutions. Our experiments show that current LMMs often fall short, not due to lack of generative capability, but because they do not sustain grounded exploration. Models often overlook relevant entities, under-examine critical parts, or hallucinate attributes not grounded in the image. Motivated by this failure mode, we propose \textbf{affordance-grounded alignment}, which casts creative tool use as a preference learning problem. Using Direct Preference Optimization, we encourage models to prefer attribute-affordance reasoning grounded in visual evidence over hallucinated alternatives. In addition, we incorporate supervision derived from an affordance knowledge base to guide broader entity exploration and multi-turn planning. Our results show consistent gains in selecting the correct entities and parts, while substantially reducing hallucination and grounding-related errors. These findings suggest that creative intelligence is not a peripheral capability, but a foundation for the next stage of multimodal AI, enabling systems that learn over time, adapt to unfamiliar environments, and solve problems beyond their training, moving closer to human-like intelligence.
\end{abstract}

\begingroup
\renewcommand{\thefootnote}{*}
\footnotetext{ indicates equal contribution}
\endgroup

\vspace{-3mm}
\begin{center}
\small

\newcommand{\logoh}{1.35em}

\href{https://github.com/CreativityBench/MM-CreativityBench}{
\raisebox{-0.2\height}{\includegraphics[height=\logoh]{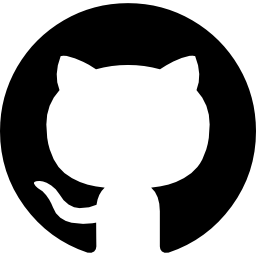}}
\hspace{0.35em}{\textbf{Code}}
}
\quad
\href{https://creativitybench.github.io/mm-creativitybench.github.io/}{
\raisebox{-0.2\height}{\includegraphics[height=\logoh]{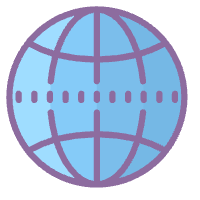}}
\hspace{0.35em}{\textbf{Project Page}}
}

\end{center}
\vspace{-2mm}
\section{Introduction}
\label{sec:intro}

In \textbf{Triarchic Theory of Intelligence}~\citep{sternberg1985beyond}, human intelligence encompasses not only analytical and practical abilities, but also \textbf{creative intelligence}: the ability to generate novel and useful solutions under constraints. In real-world, resource-limited settings, this ability often appears as tool repurposing, where people adapt available objects to fulfill functions beyond their intended use. Such creativity is not merely linguistic or associative. Humans learn object attributes, physical affordances, and object-object interactions through continuous observation and embodied experience in the physical world. They can decompose tools and everyday objects into functional modules, such as edges, tips, handles, surfaces, and containers, and mentally reassemble these modules to support new goals. For instance, a rigid edge can serve as a scraper, a thin metal tip as a lever, and a transparent curved surface as a focusing device. These solutions are not arbitrary; they arise from recognizing non-obvious yet physically valid mappings between task goals and environmental affordances~\citep{gibson1977theory,gibson1979ecological}. We study this specific form of creativity, \textbf{creative tool repurposing}, as a concrete testbed of creative intelligence in large multimodal models (LMMs).

\begin{figure}
    \centering
    \vspace{-8mm}
    \includegraphics[width=\linewidth]{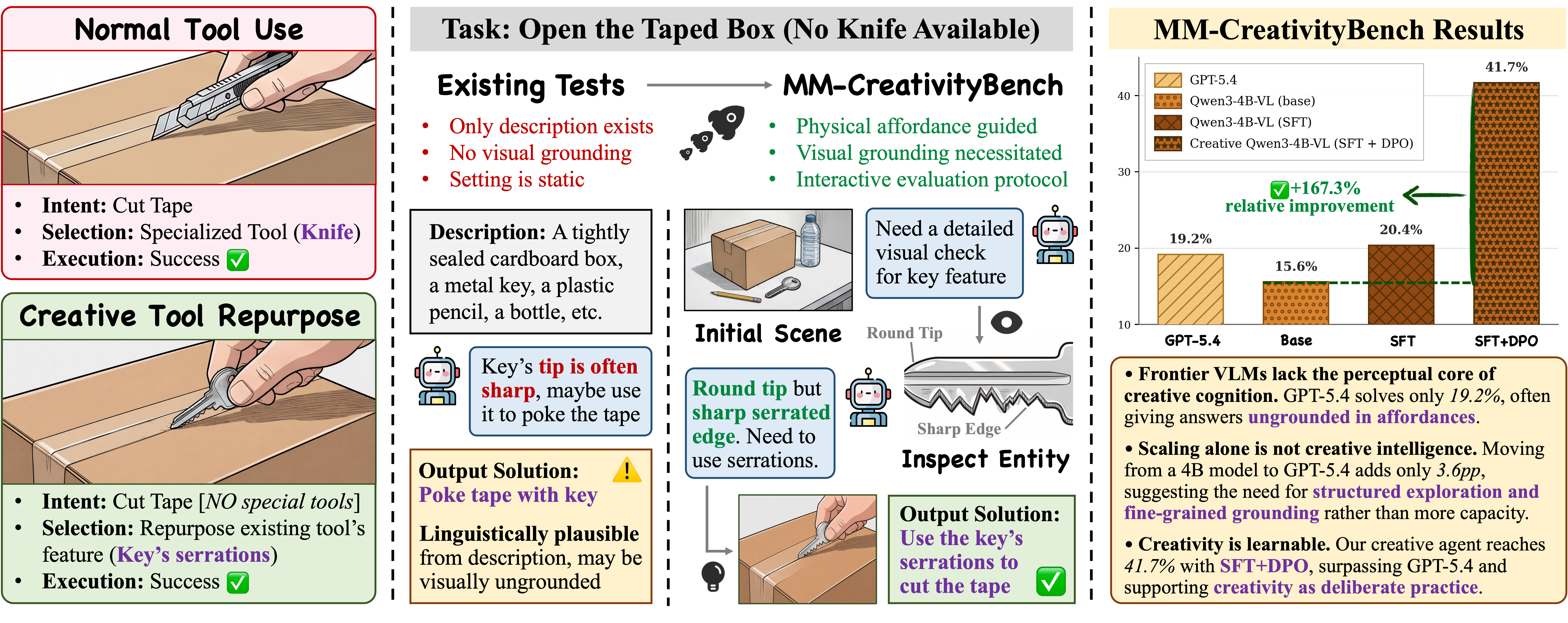}
    \vspace{-6mm}
    \caption{Beyond normal tool use, \textbf{creative tool repurposing} requires visual grounding of physical affordances, enabling the model to discover that a key’s serrated edge can cut box tape. MM-CreativityBench shows that such affordance-guided reasoning is poorly captured by frontier VLMs but can be improved through training.}
    \vspace{-6mm}
    \label{fig:intro} 
\end{figure}

Despite the rapid progress of data-driven LMMs, it remains unclear whether they acquire this kind of creative intelligence. Current models can often describe objects, retrieve common tool-use patterns, or generate plausible solutions from textual priors. However, they frequently fail to transfer knowledge across functional similarity, physical affordance, or task context. This limitation suggests that their reasoning may still be constrained by word-level or pixel-level shortcuts rather than an abstract, compositional understanding of how physical properties enable functions~\citep{yuksekgonul2023when}. Moreover, creative tool use requires grounding object parts, geometry, material, and potential human-object interactions in the physical world, which remains challenging for existing LMMs~\citep{qian2024affordancellm}. Unlike humans, who build conceptual knowledge through perception, bodily experience, and situated action~\citep{barsalou2008grounded}, general-purpose LMMs lack experience-based learning from embodied interaction with the environment. As a result, their reasoning often resembles fast, local, and plausible ``System 1'' inference~\citep{kahneman2011thinking}, while remaining weak in long-horizon exploration and planning~\citep{valmeekam2023planbench}. This makes it difficult for them to discover new object-function mappings that are both visually grounded and physically feasible.

To tackle these challenges, recent work has begun to explore creativity in large language and multimodal models through open-ended generation and constrained problem-solving tasks~\citep{tian2024macgyver, qian2024escapebench, lim2025visescape}. However, existing evaluations remain largely text-centric and scenario-driven, offering limited insight into how models ground creative reasoning in physical environments. A central challenge is that real-world creativity is inherently perception-dependent: agents must inspect environments, identify candidate objects, attend to relevant parts, and judge whether their physical attributes, such as geometry and material, support the intended use. Without such grounding, models may produce linguistically plausible but physically invalid solutions, overlooking relevant objects, misinterpreting attributes, or hallucinating affordances that are not visually supported~\citep{zeng2024investigating,chen2024multiobject,wu2024autohallusion}. Consequently, success in text-based reasoning does not necessarily transfer to visually grounded problem-solving~\citep{zeng2024investigating}.

This gap motivates a more fundamental question: \textit{can LMMs perform creative reasoning as an evidence-driven process grounded in perception?}~\citep{liu2024convbench,liu2024visualagentbench,cao2024visdiahalbench} Addressing this question requires moving beyond static multimodal inputs toward interactive settings, where models actively decide what to inspect, iteratively refine their understanding, and connect visual evidence to task demands. The challenge is not merely to generate a creative solution, but to reach one through a \emph{visually grounded and physically feasible search process} that supports abstraction, functional transfer, and compositional use of object parts.

To this end, we introduce \textbf{MM-CreativityBench}, a benchmark for grounded creative problem solving in multimodal environments. The benchmark consists of tasks that require repurposing everyday objects under constraints, each paired with a structured visual context including a scene image, entity-level images, and zoomed-in part images. This design preserves the underlying affordance structure while introducing the perceptual challenges inherent to real-world reasoning: a successful system must not only infer what could work, but also identify the correct object and part through visual inspection and justify its feasibility. While creativity is inherently open-ended, our evaluation focuses on constrained creativity, where multiple solutions may exist but must satisfy physical and functional requirements grounded in the scene. Accordingly, task success is defined by whether a model identifies a physically valid and contextually appropriate object–part combination that fulfills the task constraints. To support this, we adopt an interactive protocol that allows models to explore the environment, update their reasoning, and refine candidate solutions before committing the answer.

Our experiments reveal a gap between surface-level plausibility and grounded reasoning. Current LMMs often generate superficially plausible answers, but struggle to carry out evidence-based creative exploration: even the strongest models achieve less than 25\% accuracy. Notably, some top closed-source models, such as GPT-5.4, may underperform open-source models such as Qwen, suggesting that scaling alone is insufficient for grounded creative reasoning. Error analysis shows consistent failure modes: models fixate on salient but irrelevant objects, neglect decisive object parts, or infer affordances unsupported by visual evidence. In many cases, the bottleneck is not the lack of candidate ideas but the inability to maintain a grounded exploration process that links perception, interaction, and physical plausibility.

To address these limitations, we further investigate whether affordance-aware alignment can improve grounded interactive behavior. Our key idea is to provide models with basic building blocks for attribute-affordance associations, enabling them to connect observable attributes to potential functional uses. Building on this, we design supervision signals that encourage evidence-based exploration, guiding models to actively inspect candidate entities, maintain a structured record of unobserved parts, and ground intermediate reasoning steps in visual evidence. We also introduce preference data with negative trajectories capturing common failure modes, including hallucinated attributes and premature commitment, and visually unsupported reasoning. Fine-tuning open-source Qwen3-VL models with these signals through supervised fine-tuning and direct preference optimization yields consistent gains, more than doubling performance in the best setting. These gains suggest that injecting affordance-level knowledge and exploration strategies is critical for grounded creative reasoning, leading to stronger visual grounding, reduced hallucination, and more accurate creative tool use. Overall, we summarize our contributions as follows:
\begin{itemize}[topsep=-1.5pt, leftmargin=10pt, itemsep=0pt]
\item \textbf{Visual Creativity Benchmark:} We introduce MM-CreativityBench, a benchmark for evaluating grounded creative tool repurposing in visual environments, where models must identify the object and part based on visual evidence and physical feasibility for creative problem-solving.
\item \textbf{Grounded Interactive Protocol:} We design an interactive evaluation setting that allows models to actively inspect scenes, entities, and parts, making it possible to measure whether creative solutions arise from evidence-driven exploration rather than unsupported guessing.
\item \textbf{Affordance-Grounded Alignment:} We systematically analyze failure modes of current LMMs in grounded creative reasoning, and show that post-training with stepwise supervision and preference optimization can yield gains in performance, grounding, and hallucination reduction.
\end{itemize}

\section{Related Work}
\label{sec:related_works}
\begin{table*}[!t]
\centering
\vspace{-11mm}
\resizebox{\linewidth}{!}{
\begin{tabular}{lcccccccc}
\toprule
\textbf{Benchmark} & \textbf{\makecell{Creative \\ Tool Use}} & \textbf{\makecell{Affordance\\Grounding}} & \textbf{\makecell{Attribute\\Grounding}} & \textbf{\makecell{Part-Level\\Reasoning}} & \textbf{\makecell{Fine-Grained\\Creativity Levels}} & \textbf{\makecell{Distractors\\Included}} & \textbf{\makecell{Visual\\Grounding}} & \textbf{\makecell{Evaluation\\Protocol}} \\
\midrule
\textit{PROST\citep{aroca2021prost}} & \xmark & \cmark & \cmark & \xmark & \xmark & \gmark & \xmark & Static \\
\textit{NEWTON\citep{wang2023newton}} & \xmark & \xmark & \cmark & \xmark & \xmark & \gmark & \xmark & Static \\
\textit{Creation-MMBench\citep{tian2024macgyver}}      
& \xmark & \xmark & \xmark & \xmark & \xmark & \xmark & \cmark & Static \\
\textit{VillagerBench\citep{dong2024villageragent}}      
& \xmark & \gmark & \xmark & \xmark & \xmark & \xmark & \cmark & Interactive \\
\textit{VisEscape\citep{lim2025visescape}}      
& \xmark & \gmark & \xmark & \xmark & \xmark & \gmark & \cmark & Interactive \\
\textit{PIQA\citep{bisk2020piqa}} & \gmark & \gmark & \gmark & \xmark & \xmark & \cmark & \xmark & Static \\
\textit{MacGyver\citep{tian2024macgyver}}      
& \cmark & \gmark & \gmark & \xmark & \xmark & \xmark & \cmark & Static \\
\textit{EscapeBench\citep{qian2024escapebench}}      
& \cmark & \gmark & \xmark & \xmark & \gmark & \xmark & \cmark & Interactive \\
\textit{CretivityBench\citep{qian2026creativitybench}}      
& \cmark & \cmark & \cmark & \cmark & \cmark & \cmark & \xmark & Static \\
\midrule
\multirow{1}{*}{\textbf{{\textit{MM-CreativityBench} (Ours)}}}     
& \cmark & \cmark & \cmark & \cmark & \cmark & \cmark & \cmark & Interactive \\
\bottomrule
\end{tabular}
}
\vspace{-1.5mm}
\caption{For each existing benchmark, the table indicates whether the corresponding dimension is fully addressed (\cmark), partially addressed (\gmark), or not addressed (\xmark).}
\vspace{-5.5mm}
\label{tab:comparison}
\end{table*}

\textbf{Creativity in Multimodal and Language Models.}
Creativity in LLMs has been studied through open-ended generation tasks such as storytelling~\citep{akoury2020storium, brown2020language}, design~\citep{qian2023creator, cai2023large, ha2025synthia}, and ideation~\citep{si2024can, wang2024scimon, qian2025modelingagent, yang2024large, wang2026creativebench}, often evaluated using notions of novelty, diversity, and usefulness. More recent work extends this to creative problem solving, including tool-use and object repurposing scenarios where models must generate unconventional but feasible solutions under constraints~\citep{tian2024macgyver, qian2024escapebench, qian2026creativitybench}, as well as multimodal settings involving non-literal image understanding, context-aware generation, and exploration-driven decision making~\citep{huang2025causality,fang2025creation,lim2025visescape}. However, across both LLM and LMMs benchmarks, these evaluations are largely scenario-driven, emphasizing planning, reasoning, or interaction rather than the fine-grained mechanisms of affordance-grounded creative tool use (\Cref{tab:comparison}); how models derive novel solutions from object properties, especially under visual grounding, remains underexplored.

\textbf{Affordance-Grounded Reasoning and Alignment.}
Affordance reasoning has been studied as a bridge between perception and action, including in physical commonsense benchmarks such as PIQA, PROST, and NEWTON~\citep{bisk2020piqa,aroca2021prost,wang2023newton}, and in robotics and embodied AI for manipulation and planning~\citep{montesano2008learning,jamone2016affordances,chu2019learning,brohan2022rt,brohan2024rt}. Recent MLLM work introduces structured and part-level affordance representations~\citep{yu2025seqafford,ma2024glover}, improving grounded perception and reasoning. However, these approaches primarily focus on recognizing canonical affordances or action feasibility, rather than enabling flexible recombination for creative tool use grounded in fine-grained attributes. In parallel, alignment methods such as supervised fine-tuning and Direct Preference Optimization~\citep{rafailov2023direct}, along with multimodal extensions~\citep{wang2024mdpo, liu2024mia}, have proven effective at improving reasoning quality and visual grounding through preference-based learning over exploratory trajectories. However, these approaches have been studied primarily in general reasoning. Our work bridges this gap by leveraging training signals from an affordance knowledge base to reframe affordance-driven creativity as a preference optimization problem, encouraging models to prefer visually grounded attribute–affordance reasoning. This injects fine-grained attribute–affordance knowledge into the model as compositional primitives for creative recombination, enabling efficient, visually grounded creative tool use.
\section{MM-CreativityBench}
\label{sec:method}
\subsection{Preliminary Experiment}

\begin{wrapfigure}{r}{0.5\linewidth}
    \centering
    \vspace{-22mm}
    \includegraphics[width=\linewidth]{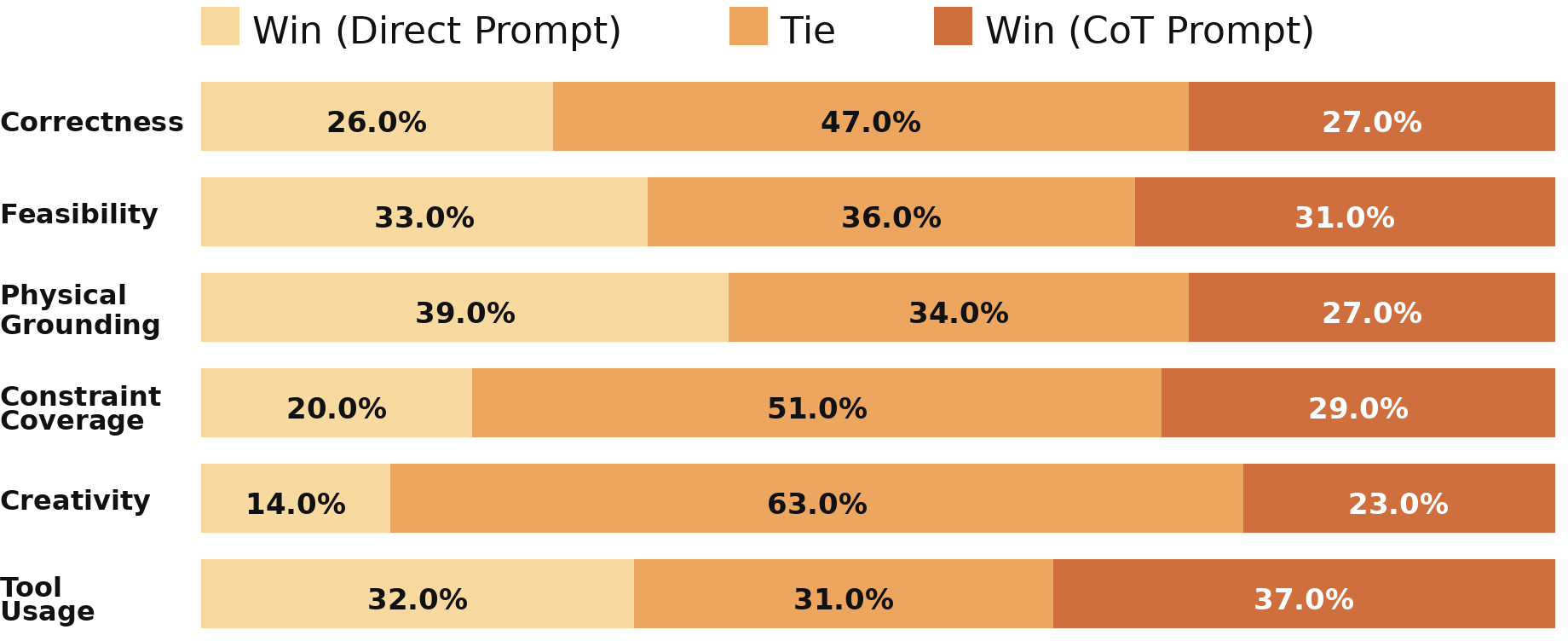}
    \vspace{-4mm}
    \caption{\textbf{Preliminary Experimental Results:} Comparison between direct prompting and structured affordance-level CoT on creative tool use tasks.}
    \vspace{-3mm}
    \label{fig:prelim_rel_eval}
\end{wrapfigure}

As a preliminary probe of creative intelligence in LMMs, we evaluate models on 100 creative tool-use tasks drawn from MacGyver~\citep{tian2024macgyver}, where each task requires repurposing everyday objects to satisfy a set of constraints. To introduce a visual grounding requirement, we augment each task with a scenario image generated by Gemini-2.5-Pro. The accompanying task description includes only constraints that are not directly observable from the image, so the model must rely on visual evidence to identify candidate objects and reason about their possible uses. Under this setup, we compare two prompting strategies: a \textbf{direct prompt}, which asks the model to produce a solution without structured guidance, and a \textbf{structured affordance-level Chain-of-Thought (CoT)} prompt~\citep{wei2022cot}, which guides the model to perceive available tools, decompose them into parts, infer physical properties, derive affordances, and verify constraint satisfaction. Detailed prompts are provided in \Cref{sec:appendix_prelim}. We use GPT-4.1-mini as the evaluated LMM and GPT-5.2 as the judge LMM model, assessing outputs along six dimensions: \textbf{Correctness}, \textbf{Feasibility}, \textbf{Physical Grounding}, \textbf{Constraint Coverage}, \textbf{Tool Usage}, and \textbf{Creativity}.

As shown in \Cref{fig:prelim_rel_eval}, structured affordance-level CoT yields modest gains on procedural dimensions, improving Constraint Coverage, Tool Usage, and Creativity. However, these gains do not translate into reliable end-to-end success: Correctness improves only marginally, while Feasibility and Physical Grounding remain limited or inconsistent. This suggests that prompting models to explicitly list objects, parts, attributes, and affordances can organize reasoning, but does not ensure that the final solution is grounded in fine-grained visual evidence. Models may still produce plausible creative uses without verifying whether the selected part actually has the physical attributes required for the task. These results motivate both our benchmark and training design: MM-CreativityBench evaluates creative tool use as an interactive, part-level grounding problem, while our affordance-grounded alignment method provides explicit supervision and preference signals that teach models to explore relevant evidence, connect attributes to affordances, and reject visually unsupported solutions.

\subsection{Benchmark Task Construction}

The preliminary study shows that structured prompting can organize creative reasoning, but does not reliably ground the final solution in the visual and physical attributes of a specific object part. We therefore construct MM-CreativityBench from a part-level affordance knowledge base, so that each task has an explicit evidence structure underlying the correct creative solution.

\paragraph{Creative affordance knowledge base.}
We build MM-CreativityBench on top of the existing open-source affordance knowledge base\citep{qian2026creativitybench}. The knowledge base provides structured annotations for everyday physical objects, including part decompositions, part-level physical and state attributes, and functional affordances (please see \Cref{apdx:creative_kb} for details). Formally, each entity $e\in\mathcal{E}$ is decomposed into functional parts:
\[
P(e)=\{p_1,\ldots,p_m\}.
\]
Each part $p\in P(e)$ is associated with an attribute set $A(p)=A^{\mathrm{phy}}(p)\cup A^{\mathrm{state}}(p)$, where $A^{\mathrm{phy}}(p)$ captures stable physical properties such as geometry, material, rigidity, sharpness, hollowness, or surface texture, and $A^{\mathrm{state}}(p)$ captures situational properties such as whether the part is open, clean, intact, accessible, or detachable. These annotations provide the fine-grained evidence needed to decide whether a part can be repurposed for a novel use.

\paragraph{Reverse task construction.}
Given the affordance knowledge base, we construct each benchmark instance as an inverse grounding problem rather than writing scenarios first and labeling answers afterward. Specifically, we sample a target entity--part pair $(e^*,p^*)$ and a gold affordance $f^*$ supported by $A(p^*)$, forming the gold solution $g=(e^*,p^*,f^*)$. We then generate a task description $x$ that requires $f^*$ without revealing the target entity or part, and sample distractor entities $E^-$ to form the candidate set:
\[
T=(x,E,g), \qquad E=\{e^*\}\cup E^-, \qquad g=(e^*,p^*,f^*).
\]
Distractors are selected to make the task diagnostic: some contain parts with affordances similar to $f^*$ but lack a decisive physical or state attribute, while others are scene-plausible objects that naturally co-occur with the gold entity but cannot satisfy the task constraints. Thus, success requires identifying the correct entity and part through fine-grained grounding rather than object priors alone. We retain only high-quality tasks satisfying gold validity, distractor separability, scene coherence, and visual observability, resulting in 333 held-out MM-CreativityBench test tasks and 868 disjoint training tasks for trajectory sampling. Details of reverse task generation, distractor construction, filtering, and human verification are provided in \Cref{apdx:task_construction}.

\paragraph{Multimodal Grounding via Image Generation}
After constructing each symbolic task $T=(x,E,g)$, we augment it with images at three granularities: environment, entity, and part. This mirrors the interaction process required by the benchmark: the model first observes the full scene, then inspects candidate entities, and finally verifies decisive part-level evidence. For each task, we generate
\[
I_e=\pi_{\mathrm{ent}}(e,P(e),A), \qquad
I_{e,p}=\pi_{\mathrm{part}}(e,p,A(p),I_e), \qquad
I_{\mathrm{env}}=\pi_{\mathrm{env}}(x,E,\{I_e:e\in E\}).
\]
Here, $I_e$ provides a full-object view, $I_{e,p}$ provides a zoomed-in view of part $p$, and $I_{\mathrm{env}}$ places all candidate entities into a coherent scene. This three-level construction is essential because distractors are intentionally plausible at the object level, while the correct answer often depends on local attributes of a specific part. Therefore, the benchmark requires models to navigate
\[
I_{\mathrm{env}} \rightarrow I_e \rightarrow I_{e,p} \rightarrow (e^*,p^*,f^*)
\]
and ground the final solution in inspected visual evidence. Details of image generation are provided in \Cref{apdx:image_construction}.

\subsection{Training Trajectory Construction}
The benchmark construction above defines the evaluation problem: given a visually grounded scene, a model must identify the entity and part whose physical attributes support the target affordance. We now use the same task structure to construct training data. The key motivation is that grounded creative tool use is not only a final-answer problem, but also a process problem. A model must decide which entity to inspect, which part to verify, how to interpret the observed attributes, and when to reject plausible but physically invalid alternatives. Therefore, instead of supervising only the final solution, we construct multi-turn trajectories that teach evidence-seeking behavior from scene-level search to part-level affordance grounding.

\paragraph{Interactive trajectory format.}
For each multimodal task $\mathcal{T}=(x,I_{\mathrm{env}},E,g)$ with gold solution $g=(e^*,p^*,f^*)$, we represent an interaction trajectory as
\[
\tau=\{(o_t,r_t)\}_{t=1}^{T}, \qquad o_t=(u_t,I_t), \qquad r_t=(z_t,a_t).
\]
Here, $u_t$ is the feedback message, $I_t$ is the visual observation returned at turn $t$, $z_t$ is the model's reasoning, and $a_t$ is a structured action. The action space contains three operations:
\[
a_t\in\{\texttt{inspect\_entity}(e),\ \texttt{inspect\_part}(e,p),\ \texttt{answer}(e,p,h)\},
\]
where $e\in E$, $p\in P(e)$, and $h$ describes how the selected part should be used. This format mirrors the visual hierarchy of the benchmark. The initial turn provides the environment image $I_{\mathrm{env}}$; inspecting an entity returns its full-object image $I_e$ and part list $P(e)$; inspecting a part returns the zoomed-in part image $I_{e,p}$, optionally with short attribute-level textual disambiguation. Thus, each action explicitly determines what evidence the model receives next.

\paragraph{Knowledge-guided exploration stack.}
To construct a systematic positive trajectory, we maintain an ordered exploration stack $\mathcal{S}_t$ whose elements are candidate entities or parts. The top element determines the next inspection target. We define an affordance-relevance function $J(e,p)\in{0,1}$, where $J(e,p)=1$ indicates that part $p$ of entity $e$ has an affordance similar or relevant to the target affordance $f^*$ according to the knowledge base. This allows the trajectory to prioritize promising candidates while still grounding exploration in structured affordance knowledge.
\begin{itemize}[topsep=-2pt,leftmargin=12pt,itemsep=-1pt]
\item \textbf{Initialization:} Given the scene and task, the model proposes an ordered list of candidate entities to initialize $\mathcal{S}_t$, thereby directing early exploration toward likely relevant objects.
\item \textbf{Inspect Entity:} When an entity $e$ is inspected, it is removed from the stack, and its affordance-relevant parts ${p\in P(e):J(e,p)=1}$ are pushed for part-level inspection. This turns coarse entity-level exploration into finer part-level verification.
\item \textbf{Inspect Part:} When a part $p$ is inspected, it is removed from the stack and assigned a binary judgment $b_t\in{0,1}$, indicating whether its observed attributes satisfy the task requirements.
\item \textbf{Answer:} Exploration stops when no unexplored entity or part remains. In the final answer turn, the model compares all inspected parts with $b_t=1$ and selects the final pair.
\end{itemize}
This stack mechanism yields a coarse-to-fine trajectory: the model first explores candidate entities, then verifies parts that may support the target affordance, and finally chooses among plausible parts. This is important because many distractors are intentionally affordance-similar; the model must learn not only to identify a plausible part, but also to select the gold pair $(e^*,p^*)$ whose attributes best satisfy the task constraints.

\begin{figure}[!t]
    \centering
    \vspace{-8mm}
    \includegraphics[width=\linewidth]{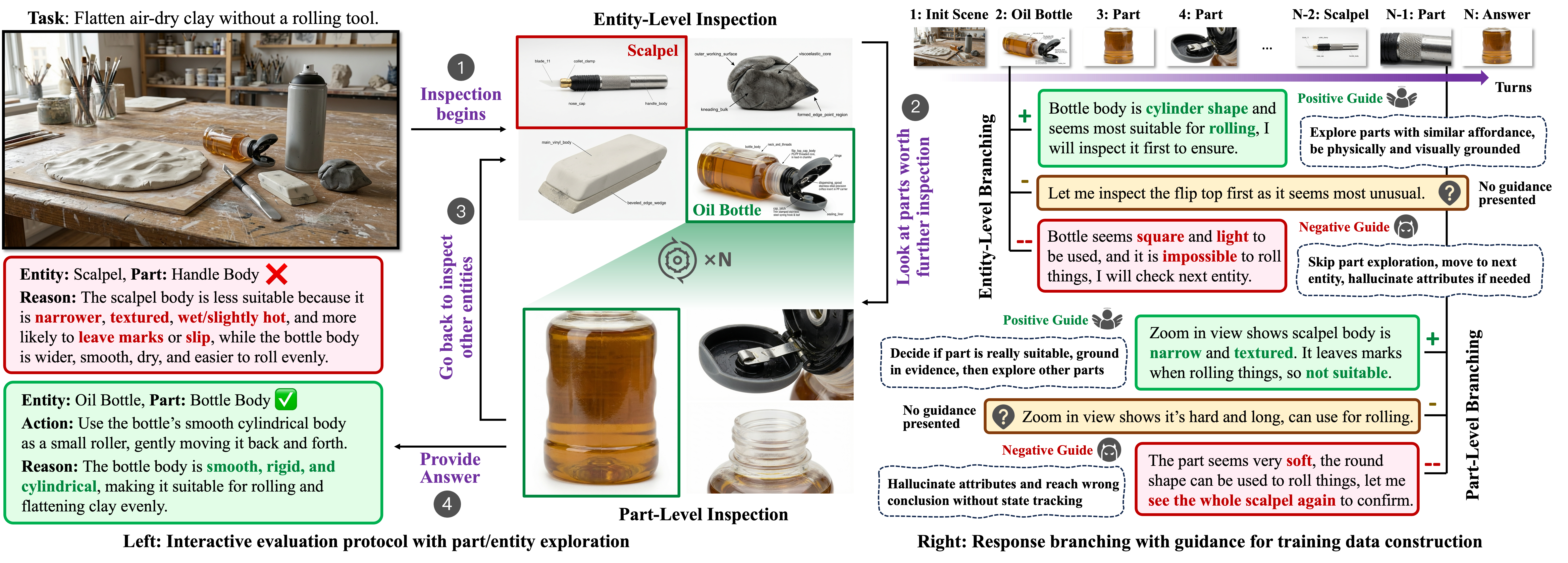}
    \vspace{-2mm}
    \caption{Interactive MM-CreativityBench evaluation and training, where models inspect scenes, entities, and parts to ground creative tool use while learning to avoid hallucinated affordances.}
    \vspace{-3mm}
    \label{fig:method}
\end{figure}

\paragraph{Three-branch trajectory sampling.}
The exploration stack determines what should be inspected at each turn, but it does not by itself determine how the model should reason about the inspection. Since the structured action $a_t$ and the textual reasoning $z_t$ play different roles, we generate guided reasoning branches at each shared interaction context $c_t$. As illustrated on the right side of \Cref{fig:method}, we sample three aligned branches with the same response format,
\[
r_t^{b}=(z_t^{b},a_t^{b}), \qquad b\in\{+,-,--\},
\]
but with different guidance signals.

\begin{itemize}[topsep=-2pt,leftmargin=12pt,itemsep=-1pt]
    \item \textbf{Positive branch $(+)$:} The positive branch is guided by structured knowledge, including relevant attributes, affordance judgments, and the gold solution when needed. Its reasoning is expected to justify each action with visually grounded evidence and remain consistent with the exploration stack. Together with the stack, this branch forms the positive trajectory used for supervised fine-tuning, teaching the model how to explore entities and parts systematically.

    \item \textbf{Negative branch $(-)$:} The negative branch receives only standard observable feedback, such as the task description, images, entity names, and part names, without hidden affordance labels or gold guidance. It therefore captures realistic inference-time mistakes, such as overlooking decisive parts, over-exploring irrelevant objects, or selecting a plausible but suboptimal part.

    \item \textbf{Hard-negative branch $(--)$:} The hard-negative branch is constructed to create stronger contrast for preference learning. It preserves fluent reasoning and valid action format, but is guided toward misleading conclusions, such as hallucinating unsupported attributes, relying on object-level priors, or choosing an affordance-similar distractor that lacks the required physical evidence.
\end{itemize}

Only the positive branch updates the exploration stack, ensuring that future observations remain coherent and grounded. The negative and hard-negative branches are sampled at the same states only as rejected alternatives, yielding aligned turn-level triples
\[
(c_t,r_t^+,r_t^-,r_t^{--}),
\]
where $r_t^+$ is the preferred grounded response and $r_t^-,r_t^{--}$ are rejected responses of increasing difficulty. The resulting data support both training stages: positive trajectories teach systematic exploration through SFT, while positive--negative comparisons enable DPO-style training to favor visually grounded attribute--affordance reasoning over fluent but unsupported alternatives. Please see \Cref{apdx:trajectory_construction} for more trajectory construction details.

\subsection{Affordance-Grounded Alignment}
Given the constructed trajectory dataset, we align the model with affordance-grounded creative reasoning in two stages: (1) Supervised fine-tuning (SFT) teaches structured exploration from positive trajectories, and (2) turn-level Direct Preference Optimization (DPO) sharpens attribute--affordance grounding by contrasting grounded reasoning with plausible but unsupported alternatives.

\textbf{Supervised Fine-Tuning.}
We first fine-tune the model on positive trajectories $\tau^{+}=\{(c_t,r_t^{+})\}_{t=1}^{T}$ that are grounded in the affordance knowledge base $\mathcal{K}$. Given each interaction context $c_t$, the model is trained to imitate the positive response $r_t^{+}=(z_t^{+},a_t^{+})$ using a standard token-level cross-entropy objective over complete multi-turn interactions. Imitating full trajectories, rather than only final answers, encourages the model to learn the evidence-seeking process: selecting candidate entities, inspecting relevant parts, interpreting observed attributes, and comparing entity--part pairs before producing the final solution. However, because these trajectories are generated with structured guidance, SFT primarily teaches the model a guided exploration policy and does not explicitly penalize spurious attribute--affordance associations.

\textbf{Turn-Level Direct Preference Optimization.}
At inference time, the model will operate without the gold guidance, which can lead to structurally valid but poorly grounded reasoning. To reduce this gap, we apply DPO under the unguided evaluation protocol. For each shared context $c_t$, we form preference pairs $(c_t,r_t^{+},r_t^{\mathrm{rej}})$, where the preferred response $r_t^{+}$ is drawn from the grounded positive branch and the rejected response $r_t^{\mathrm{rej}}\in\{r_t^{-},r_t^{--}\}$ is sampled from the negative or hard-negative branch of our three-branch trajectory construction. These rejected responses often preserve valid action formats and plausible entity--part choices, yet misinterpret or overclaim the visual evidence. Contrasting them under identical contexts trains the model to prefer responses that justify affordances using observed physical or state attributes, directly targeting the core failure mode of attribute--affordance reasoning under multimodal uncertainty. Full objective formulations, context construction, and trajectory notation are provided in \Cref{apdx:affordance_grounded_alignment}.

\section{Experiment}
\label{sec:experiment}

\subsection{Implementation Details}
\textbf{Benchmark Evaluation Protocol.}
We use an interactive evaluation protocol in which models explore a scenario image before producing a final answer. Each example begins with an image containing multiple entities. As illustrated in \Cref{fig:method}, the model may iteratively inspect entities and their parts to obtain closer views and examine relevant attributes before deciding on an answer. The model is not required to inspect every region, but effective exploration should help ground the final creative solution in object-specific visual evidence.

In our main setting, the conversation history includes the initial scenario image and the most recently inspected view. At each step, the model first provides its reasoning and then chooses one of three actions: inspect an entity, inspect a part, or give the final answer. For inspection actions, the model specifies the selected entity or part; for final answers, it explains how the explored evidence supports a creative and grounded response. We evaluate open- and closed-source model families including GPT, Qwen3-VL, InternVL3.5, and Gemma-4, using maximum context length and zero temperature. Full prompt details are provided in \Cref{apdx:experiment_details}.

\begin{wraptable}{r}{0.6\linewidth}
    \vspace{-4mm}
    \centering
    \small
    \resizebox{\linewidth}{!}{
    \begin{tabular}{l @{\hspace{3mm}} l @{\hspace{3mm}} c @{\hspace{3mm}} | @{\hspace{3mm}} l @{\hspace{3mm}} l @{\hspace{3mm}} c}
        \toprule
        \textbf{Category} & \textbf{Statistic} & \textbf{Value} & \textbf{Category} & \textbf{Statistic} & \textbf{Value} \\
        \midrule
        \multirow{3}{*}{\makecell[l]{Test Set}} 
        & Data Points & 333 
        & \multirow{3}{*}{\makecell[l]{Training Set}} 
        & Data Points & 868 \\
        & Number of Entities & 974 
        & & Number of Entities & 1,498 \\
        & Number of Parts & 6,344 
        & & Number of Parts & 10,080 \\
        \midrule
        \multicolumn{2}{c}{SFT Data Points} & 19,533 
        & \multicolumn{2}{c}{DPO Data Points} & 5,000 \\
        \bottomrule
    \end{tabular}
    }
    \vspace{-1.5mm}
    \caption{Overall statistics for MM-CreativityBench and the training set used for trajectory sampling. The test and training sets contain no overlapping scenes, entities, or parts.}
    \vspace{-3.5mm}
    \label{tab:statistics}
\end{wraptable}

\textbf{Training Implementation Details.}
We train Qwen3-VL-4B-Instruct and Qwen3-VL-8B-Instruct with both SFT and DPO using sampled training trajectories. For SFT, we construct each trajectory using only the positive branch at every turn. For DPO, we build the conversation context from positive branches, and use the positive response at the current turn as the chosen sample. The rejected sample is either the negative branch or the hard negative branch, corresponding to the DPO (normal negative) and DPO (hard negative) settings, respectively. We also evaluate a two-stage SFT+DPO setting, where the model is first trained with SFT and then further optimized with DPO. All trajectories are sampled from 868 training tasks with scenarios and entities entirely disjoint from the test set. See \Cref{tab:statistics} for dataset statistics and \Cref{apdx:experiment_details} for training hyperparameters.

\subsection{Evaluation Metric}
The agent is primarily challenged to perform visual and physical grounding: it must identify the correct entity to repurpose and the specific part that should be used. Therefore, our main metric is \textbf{Gold Correct Rate}, which measures whether the model selects both the correct entity and the correct part. We also report \textbf{Entity Correct Rate}, which counts a prediction as correct as long as the selected entity is correct. By definition, \textbf{Entity Correct Rate} should be no lower than \textbf{Gold Correct Rate}.

We additionally report interaction and grounding statistics, including the \textbf{Average Number of Exploration Turns} and the \textbf{Average Number of Distinct Entities/Parts Explored}. To assess whether a model’s answer is grounded in its interaction history, we also measure whether it inspected the gold entity and gold part before answering. We present the benchmarking results in \Cref{tab:image_eval_main} and preliminary trained-model results in \Cref{tab:image_eval_sft}, followed by the main findings below.

\begin{table*}[t]
    \centering
    \setlength{\tabcolsep}{3pt}
    \renewcommand{\arraystretch}{1.12}

    \newcolumntype{C}[1]{>{\centering\arraybackslash}p{#1}}

    \resizebox{\linewidth}{!}{%
    \begin{tabular}{
        l
        C{1.15cm} C{1.25cm} C{0.9cm}
        @{\hspace{8mm}}
        C{1.35cm} C{1.35cm}
        @{\hspace{10mm}}
        C{2.25cm} C{2.25cm}
        @{\hspace{12mm}}
        C{2.25cm} C{2.25cm}
    }
        \toprule
        \multirow{2.5}{*}{\textbf{Model}}
        & \multirow{2.5}{*}{\textbf{\makecell{Gold\\Correct}}}
        & \multirow{2.5}{*}{\textbf{\makecell{Entity\\Correct}}}
        & \multirow{2.5}{*}{\textbf{Turns}}
        & \multicolumn{2}{c}{\hspace*{-5mm}\textbf{Avg. Distinct Explored}\hspace*{5mm}}
        & \multicolumn{2}{c}{\hspace*{-5mm}\textbf{Gold Entity Explored Before Answer}\hspace*{5mm}}
        & \multicolumn{2}{c}{\hspace*{-5mm}\textbf{Gold Part Explored Before Answer}\hspace*{5mm}} \\

        \cmidrule(l{-0.8em}r{2.2em}){5-6}
        \cmidrule(l{-0.8em}r{2.2em}){7-8}
        \cmidrule(l{-0.8em}r{2.2em}){9-10}

        & & & 
        & \textbf{Entities}
        & \textbf{Parts}
        & \textbf{\makecell{Entity Correct}}
        & \textbf{\makecell{Entity Wrong}}
        & \textbf{\makecell{Part Correct}}
        & \textbf{\makecell{Part Wrong}} \\
        
        \midrule
        GPT-5.4 & 0.192 & 0.435 & 4.177 & 1.661 & 1.492 & 0.510 & 0.218 & 0.422 & 0.059 \\
        GPT-5.4 Mini & 0.183 & 0.408 & 4.072 & 1.360 & 0.706 & 0.662 & 0.193 & 0.279 & 0.033 \\
        Qwen3-VL-8B-Instruct & 0.192 & 0.441 & 13.450 & 4.979 & 3.766 & 0.993 & 0.747 & 0.953 & 0.201 \\
        Qwen3-VL-32B-Instruct & 0.240 & 0.447 & 8.766 & 4.802 & 2.309 & 1.000 & 0.750 & 1.000 & 0.111 \\
        InternVL3.5-14B & 0.150 & 0.345 & 4.847 & 1.811 & 1.700 & 1.000 & 0.211 & 0.960 & 0.032 \\
        InternVL3.5-38B & 0.156 & 0.426 & 6.991 & 3.565 & 1.775 & 1.000 & 0.524 & 0.942 & 0.068 \\
        Gemma-4-26B-A4B-it & 0.183 & 0.402 & 5.330 & 2.679 & 1.574 & 1.000 & 0.477 & 0.902 & 0.040 \\
        Gemma-4-31B-it & 0.165 & 0.354 & 3.802 & 1.982 & 0.796 & 1.000 & 0.256 & 0.545 & 0.018 \\
        \bottomrule
    \end{tabular}%
    }
    \caption{The benchmarking results on MM-CreativityBench. Models often locate the relevant entity but struggle with fine-grained gold-part grounding. Larger exploration traces improve evidence coverage but do not guarantee correct answers, revealing bottlenecks in visual evidence use.}
    \label{tab:image_eval_main}
    \vspace{-3mm}
\end{table*}

\subsection{Main Results}

\textbf{Interactive exploration helps models find relevant evidence, but does not guarantee correct reasoning.}
Our benchmarking results in \Cref{tab:image_eval_main} show that inspecting useful visual evidence does not necessarily lead to correct final answers. For example, Qwen3-VL-32B examines the gold entity before answering in nearly all successful entity cases and achieves the highest raw gold correctness among base models, yet its final accuracy remains only \textbf{0.240}. InternVL3.5-14B and Gemma-4-26B-A4B-it show a similar pattern: although they frequently inspect the gold entity in correct-entity cases, their \textbf{gold-correct scores remain much lower}. These results suggest that models do not fail only because they overlook the relevant region. Rather, even when they find the right evidence, they often struggle to \textbf{interpret it and integrate it} into the final decision. This motivates our later training, which aims to improve both \emph{exploration policies} and the \emph{use of visual evidence} gathered through interaction.

\textbf{The main bottleneck is \textbf{fine-grained part grounding} rather than coarse entity localization.}
Across raw models, entity correctness is much higher than gold correctness. For example, GPT-5.4 reaches 0.435 entity correctness but only \textbf{0.192} gold correctness, while InternVL3.5-38B reaches 0.426 versus \textbf{0.156}. This gap shows that models can often find the relevant object, but still fail to ground the specific part or attribute needed to answer correctly. The exploration statistics reinforce this pattern: models inspect entities more reliably than parts, so broader exploration does not necessarily yield \emph{finer evidence}. Thus, our interactive image evaluation is less about object discovery and more about \textbf{part-sensitive visual reasoning}: identifying which region matters, extracting the right evidence, and using it to resolve the question. This motivates training signals that explicitly reward \emph{fine-grained grounding}, not just final-answer success.

\textbf{Model families differ in exploration style, and scaling alone does not solve interactive visual reasoning.}
\Cref{tab:image_eval_main} shows that models differ not only in final accuracy, but also in \textbf{how they explore}. Qwen3-VL models inspect many more entities on average, around 4.8--5.0, while GPT-5.4 and GPT-5.4 Mini inspect only 1.66 and 1.36. Yet \emph{more exploration is not necessarily better}: Qwen3-VL-8B explores far more than GPT-5.4 but reaches the same gold correctness of 0.192, and Qwen3-VL-32B improves only modestly to 0.240 despite larger scale and extensive inspection. Moreover, the number of interaction turns is consistently larger than the total number of explored entities and parts, suggesting \textbf{redundant exploration} and room for more efficient policies. At the same time, open-source models can match or exceed GPT performance, with Qwen3-VL-32B achieving the best raw gold correctness and Qwen3-VL-8B matching GPT-5.4 while producing richer traces. These results suggest that interactive visual reasoning is shaped by \textbf{family-specific tradeoffs} among search, grounding, and decision-making, rather than by scale alone.

\begin{takeaway}
Models can often choose relevant entities, but still fail at fine-grained part grounding and evidence integration; interactive visual reasoning is limited more by grounding and reasoning than by exploration volume or scale.
\end{takeaway}

\begin{table*}[t]
    \centering
    \setlength{\tabcolsep}{3pt}
    \renewcommand{\arraystretch}{1.12}
    
    \newcolumntype{C}[1]{>{\centering\arraybackslash}p{#1}}
    
    \resizebox{\linewidth}{!}{%
    \begin{tabular}{
        l
        C{1.15cm} C{1.25cm} C{0.9cm}
        @{\hspace{8mm}}
        C{1.35cm} C{1.35cm}
        @{\hspace{10mm}}
        C{2.25cm} C{2.25cm}
        @{\hspace{12mm}}
        C{2.25cm} C{2.25cm}
    }
        \toprule
        \multirow{2.5}{*}{\textbf{Model}}
        & \multirow{2.5}{*}{\textbf{\makecell{Gold\\Correct}}}
        & \multirow{2.5}{*}{\textbf{\makecell{Entity\\Correct}}}
        & \multirow{2.5}{*}{\textbf{Turns}}
        & \multicolumn{2}{c}{\hspace*{-5mm}\textbf{Avg. Distinct Explored}\hspace*{5mm}}
        & \multicolumn{2}{c}{\hspace*{-5mm}\textbf{Gold Entity Explored Before Answer}\hspace*{5mm}}
        & \multicolumn{2}{c}{\hspace*{-5mm}\textbf{Gold Part Explored Before Answer}\hspace*{5mm}} \\

        \cmidrule(l{-0.8em}r{2.2em}){5-6}
        \cmidrule(l{-0.8em}r{2.2em}){7-8}
        \cmidrule(l{-0.8em}r{2.2em}){9-10}

        & & & 
        & \textbf{Entities}
        & \textbf{Parts}
        & \textbf{\makecell{Entity Correct}}
        & \textbf{\makecell{Entity Wrong}}
        & \textbf{\makecell{Part Correct}}
        & \textbf{\makecell{Part Wrong}} \\
        
        \midrule
        \makecell[l]{\textbf{Qwen3-4B-VL-Instruct}} & 0.156 & 0.393 & 18.922 & 3.937 & 4.417 & 0.947 & 0.554 & 0.923 & 0.167 \\
        \makecell[l]{\quad + SFT} & 0.204 & 0.369 & 17.862 & 6.628 & 9.670 & 1.000 & 0.990 & 0.985 & 0.404 \\
        \makecell[l]{\quad + DPO (normal negative)} & 0.201 & 0.529 & 31.693 & 5.506 & 12.398 & 1.000 & 0.821 & 0.970 & 0.351 \\
        \makecell[l]{\quad + DPO (hard negative)} & 0.240 & 0.547 & 12.842 & 4.222 & 3.781 & 0.973 & 0.639 & 0.838 & 0.157 \\
        \makecell[l]{\quad + SFT + DPO (hard negative)} & 0.417 & 0.583 & 6.211 & 2.644 & 1.831 & 0.959 & 0.350 & 0.856 & 0.026 \\
        \midrule
        \makecell[l]{\textbf{Qwen3-8B-VL-Instruct}} & 0.192 & 0.441 & 13.450 & 4.979 & 3.766 & 0.993 & 0.747 & 0.953 & 0.201 \\
        \makecell[l]{\quad + SFT} & 0.273 & 0.589 & 15.646 & 6.655 & 7.679 & 1.000 & 0.993 & 1.000 & 0.293 \\
        \makecell[l]{\quad + DPO (normal negative)} & 0.258 & 0.577 & 18.480 & 5.904 & 6.384 & 1.000 & 0.936 & 0.965 & 0.231 \\
        \makecell[l]{\quad + DPO (hard negative)} & 0.261 & 0.508 & 9.054 & 4.283 & 2.509 & 0.994 & 0.613 & 0.920 & 0.082 \\
        \makecell[l]{\quad + SFT + DPO (hard negative)} & 0.393 & 0.583 & 8.069 & 4.066 & 2.327 & 1.000 & 0.576 & 0.939 & 0.059 \\
        \bottomrule
    \end{tabular}%
    }
    \caption{Interactive image-evaluation summary for Qwen3-VL base and trained models. SFT + DPO achieves the highest gold and entity correct rates with more efficient exploration. Gold parts and entities are more frequently explored when the final answer is correct than when it is wrong, suggesting that correct answers are typically grounded in relevant exploration.}
    \label{tab:image_eval_sft}
    \vspace{-3mm}
\end{table*}

\textbf{Training improves both accuracy and interaction efficiency, showing that purposeful exploration is learnable.}
As shown in \Cref{tab:image_eval_sft}, targeted training substantially changes how models interact with images. The strongest 4B variant, \textbf{SFT + DPO with hard negatives}, improves gold correctness from 0.156 to \textbf{0.417}, while reducing average turns from 18.92 to \textbf{6.21}. The 8B variant shows a similar trend, improving from 0.192 to 0.393 while reducing turns from 13.45 to 8.07. These gains are not obtained by making the model search longer or inspect more regions. Instead, training makes interaction more \emph{selective and decisive}: the model learns to gather useful evidence earlier, avoid unnecessary revisits, and stop once the evidence is sufficient. This suggests that interactive visual reasoning is a \textbf{trainable behavior} whose efficiency--accuracy tradeoff can be substantially improved.

\textbf{SFT structures exploration, while hard-negative DPO teaches the model which evidence not to trust.}
SFT helps the model produce more grounded and interpretable exploration traces, but it does not fully solve the reasoning problem. For Qwen3-4B, SFT improves gold correctness only modestly from 0.156 to 0.204, while average turns remain high at 17.86, indicating continued reliance on long, corrective exploration. The key improvement comes from adding \textbf{hard-negative DPO}, which raises gold correctness from 0.204 to 0.417 and reduces turns from 17.86 to 6.21. This suggests that the main benefit of hard negatives is not simply stronger supervision, but sharper discrimination: the model learns to reject \emph{plausible but misleading} trajectories that inspect visually relevant evidence yet support the wrong conclusion. Thus, SFT provides the structure for exploration, while hard-negative DPO reshapes the model’s preferences toward correct fine-grained attribute--affordance reasoning, enabling earlier commitment to valid evidence paths.

\begin{takeaway}
Training improves both accuracy and efficiency: SFT organizes exploration, while DPO helps models reject misleading evidence and commit earlier to valid grounding paths.
\end{takeaway}

\section{Analysis}
\label{sec:anlysis}

\subsection{Affordance similarity reveals limits in fine-grained visual grounding}

\begin{figure}[t]
    \centering
    \vspace{-6mm}
    \begin{subfigure}[t]{0.49\linewidth}
        \centering
        \includegraphics[width=\linewidth]{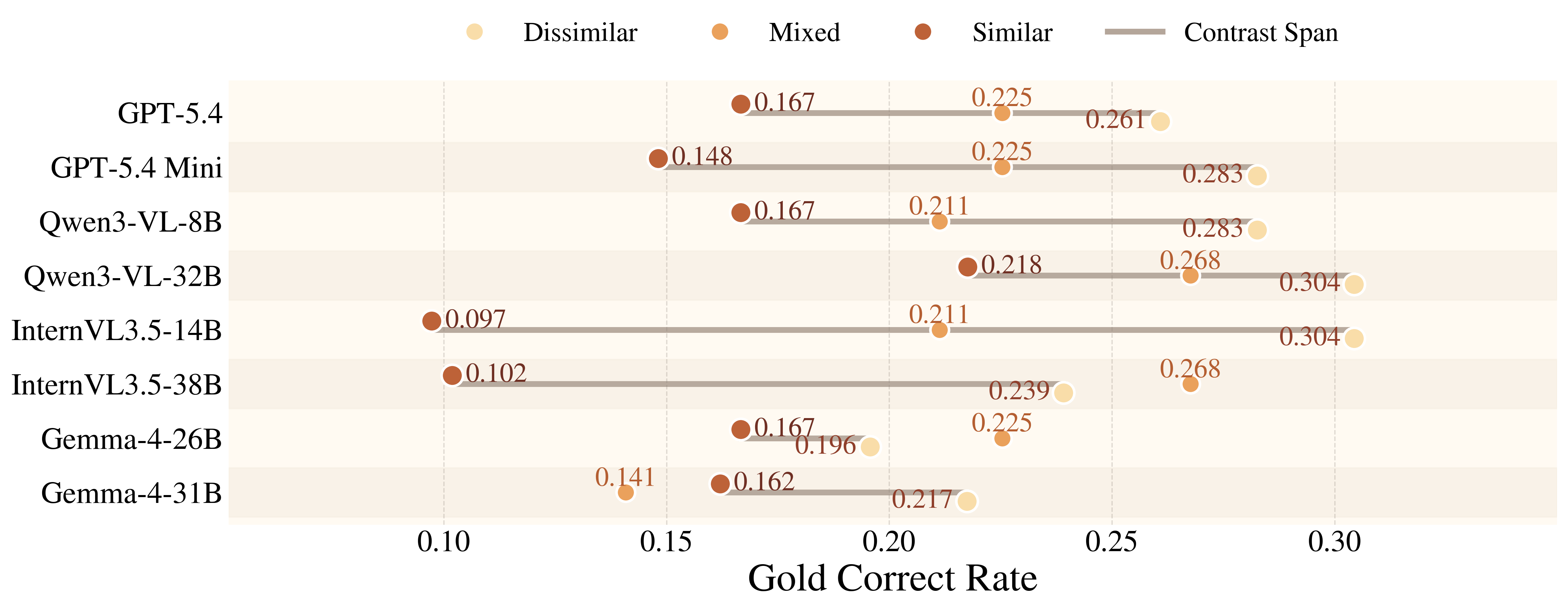}
        \caption{Performance under different affordance similarity levels.}
        \label{fig:distractor_similarity}
    \end{subfigure}
    \hfill
    \begin{subfigure}[t]{0.49\linewidth}
        \centering
        \includegraphics[width=\linewidth]{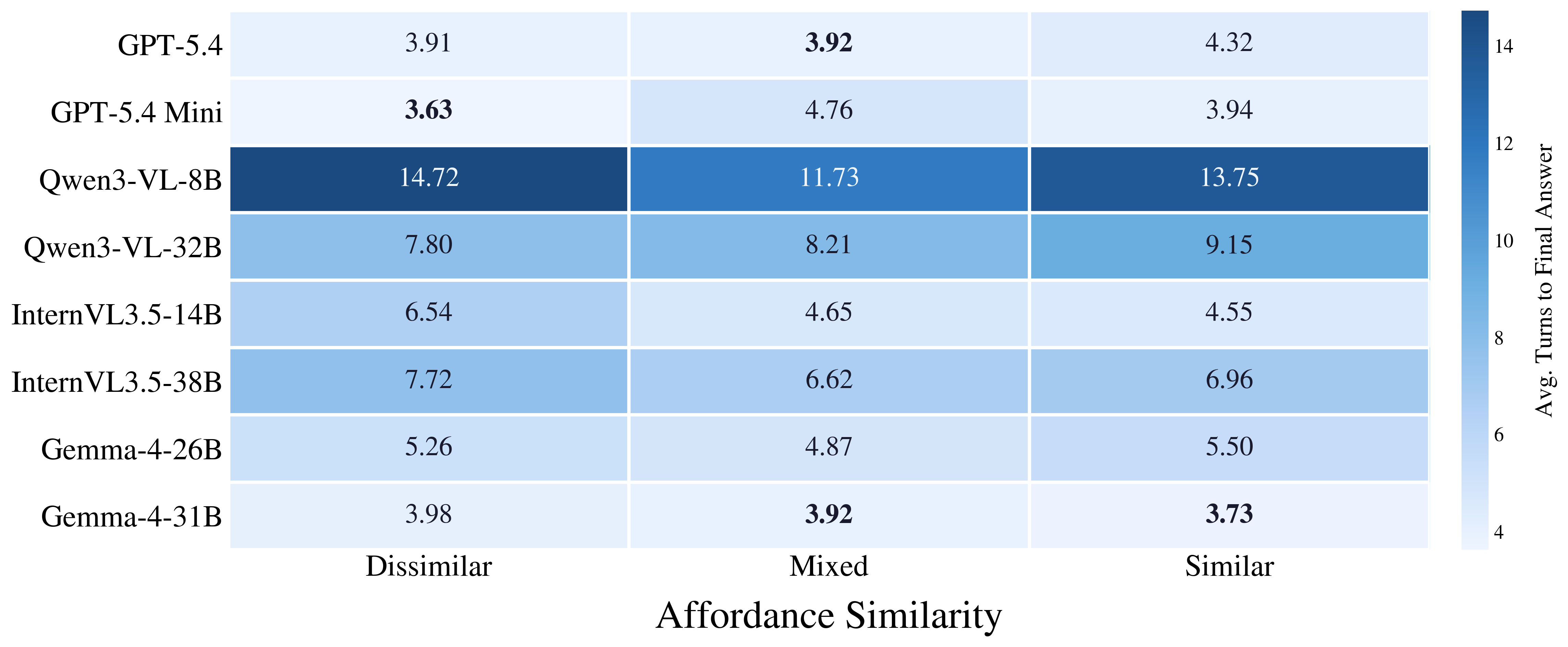}
        \caption{Average exploration turns.}
        \label{fig:avg_turn_level_2}
    \end{subfigure}
    \caption{\textbf{Effect of affordance similarity on performance and exploration.} As more entities share similar affordances, model performance often degrades, while the average number of exploration turns remains largely stable. This suggests that failures are driven less by insufficient exploration and more by weak fine-grained visual grounding and affordance disambiguation.}
    \label{fig:analysis_distractor_severity}
    \vspace{-4mm}
\end{figure}

Across environments with different affordance compositions, we observe that model performance often degrades as more entities share similar affordances. As shown in \Cref{fig:analysis_distractor_severity}, models achieve comparable accuracy in settings with dissimilar or mixed affordances, but their performance drops consistently in similar-affordance environments. This pattern indicates that the main difficulty is not simply recognizing plausible candidate tools, but distinguishing among candidates with overlapping functional affordances.

Such disambiguation requires fine-grained grounding in visual and physical attributes, such as geometry, material, accessibility, and object-part structure. However, current LMMs appear to rely on coarse affordance representations: they can often infer what type of object might be useful, but struggle to determine which specific object or part is physically best suited for the task. As a result, they may select a functionally plausible tool while missing the attribute-level evidence needed.

Notably, the average number of exploration turns remains largely unchanged across similarity levels. This suggests that models do not adapt their search behavior when the environment becomes more ambiguous; they neither inspect substantially more entities nor perform additional verification before committing to an answer. Therefore, the performance drop is unlikely to stem from insufficient exploration alone. Instead, it reflects a deeper limitation in fine-grained visual grounding and comparative affordance evaluation. These findings are also consistent with the failure modes we will discuss in \Cref{sec:error_analysis}.

\begin{takeaway}
Performance drops as affordance-similar distractors increase, while exploration remains stable, indicating that the main bottleneck is fine-grained visual grounding and affordance disambiguation rather than search effort.
\end{takeaway}

\subsection{Higher-level affordance typicality does not translate to better performance}

\begin{figure}[t]
    \centering
    \begin{subfigure}[t]{0.49\linewidth}
        \centering
        \includegraphics[width=\linewidth]{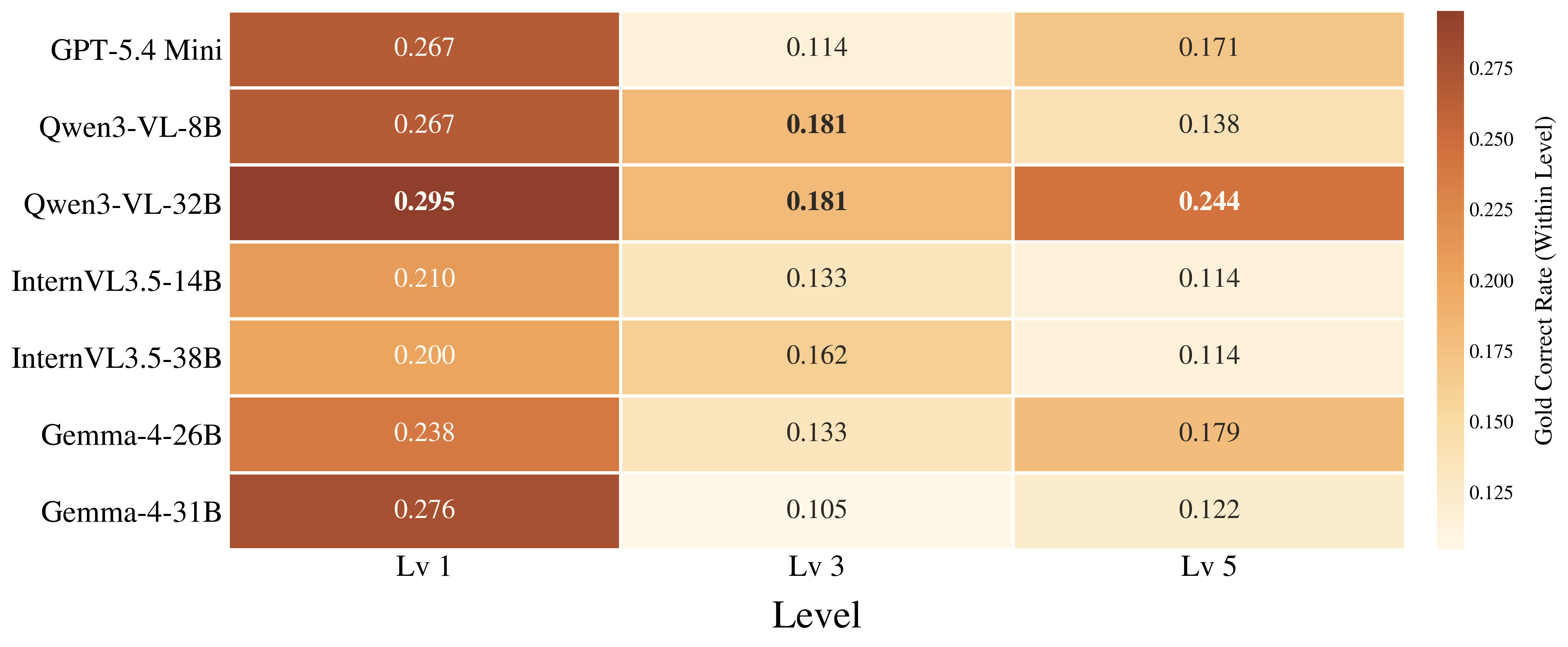}
        \caption{Gold affordance typicality level.}
        \label{fig:perf_level}
    \end{subfigure}
    \hfill
    \begin{subfigure}[t]{0.49\linewidth}
        \centering
        \includegraphics[width=\linewidth]{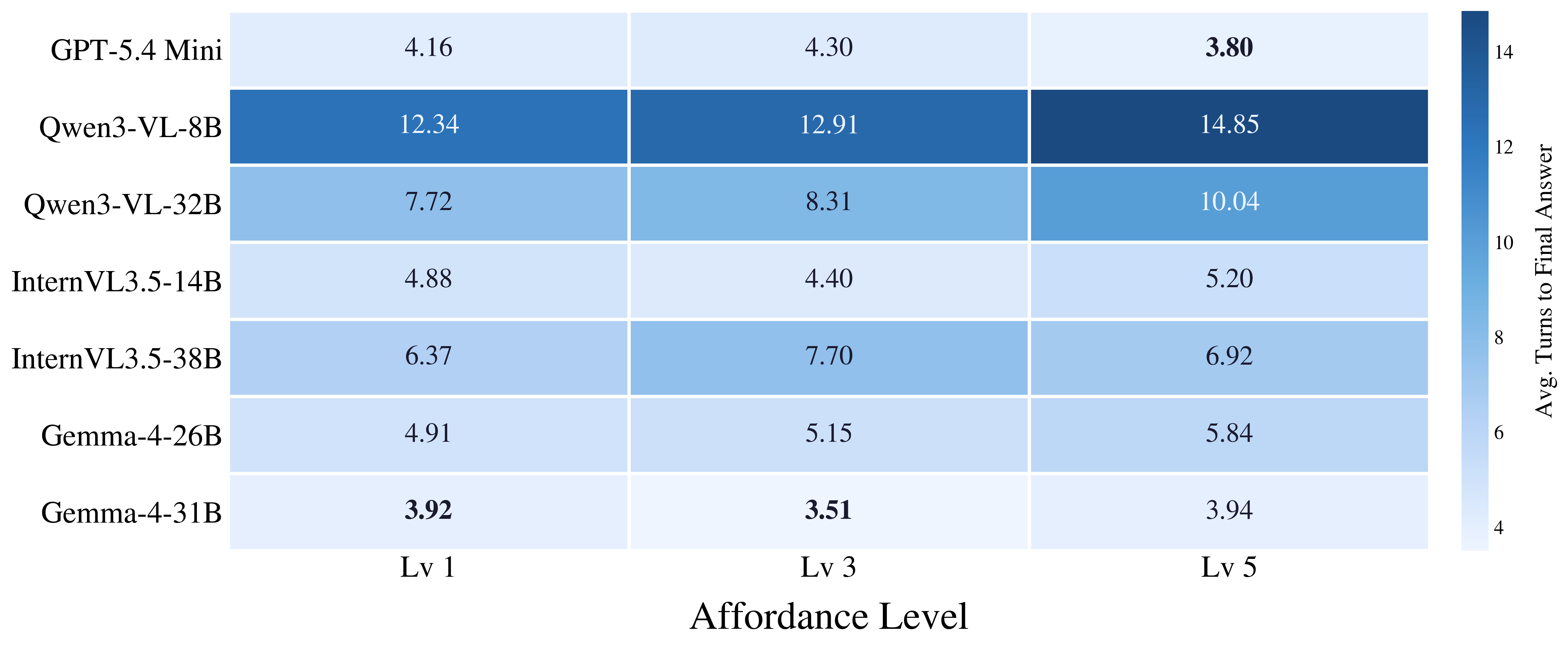}
        \caption{Average number of exploration turns.}
        \label{fig:avg_turn_level_1}
    \end{subfigure}
    \caption{\textbf{Impact of affordance typicality on performance and exploration.} Performance does not improve with higher affordance typicality. Although more typical affordances (Lv 3--5) induce longer exploration, they do not lead to higher accuracy, suggesting greater ambiguity among plausible candidates and persistent limitations in fine-grained visual grounding.}
    \label{fig:analysis_level}
    \vspace{-2mm}
\end{figure}

Contrary to the expectation that more natural or common affordances should be easier, we observe an inverse trend in \Cref{fig:perf_level}: performance does not consistently improve as affordance typicality increases. Across models, higher-level affordances (Lv 3--5), which correspond to more natural and commonly repurposed uses, do not yield better gold correctness than lower-level, more atypical affordances. This suggests that increasing ground truth typicality does not necessarily reduce the difficulty of identifying the correct tool--affordance pair.

At the same time, the average number of exploration turns generally increases with affordance typicality, indicating that models tend to produce longer reasoning chains when the target affordance appears more natural. This behavior suggests that models may over-explore or consider a broader set of plausible candidates, rather than confidently converging on the correct one. One possible explanation is that higher-typicality affordances create greater functional overlap among candidate entities, making it harder to distinguish the gold part from other plausible parts. Since current models remain weak at fine-grained attribute--affordance grounding, this additional ambiguity leads to prolonged exploration without corresponding gains in accuracy.

Together, these results indicate that more natural or familiar affordances do not necessarily simplify the task. Instead, they can introduce additional ambiguity by increasing the number of plausible candidate entities and parts. This further reinforces that the primary bottleneck is not exploration capacity alone, but the ability to reliably distinguish among candidates that share similar affordance structures using visual and physical evidence.

\begin{takeaway}
More typical affordances lead to longer exploration but not better performance, highlighting that ambiguity among plausible candidates, rather than rarity alone, drives failures under weak visual grounding.
\end{takeaway}

\subsection{Impact of visual grounding and interaction dynamics}

\begin{wrapfigure}{r}{0.5\linewidth}
    \centering
    \vspace{-5mm}
    \includegraphics[width=\linewidth]{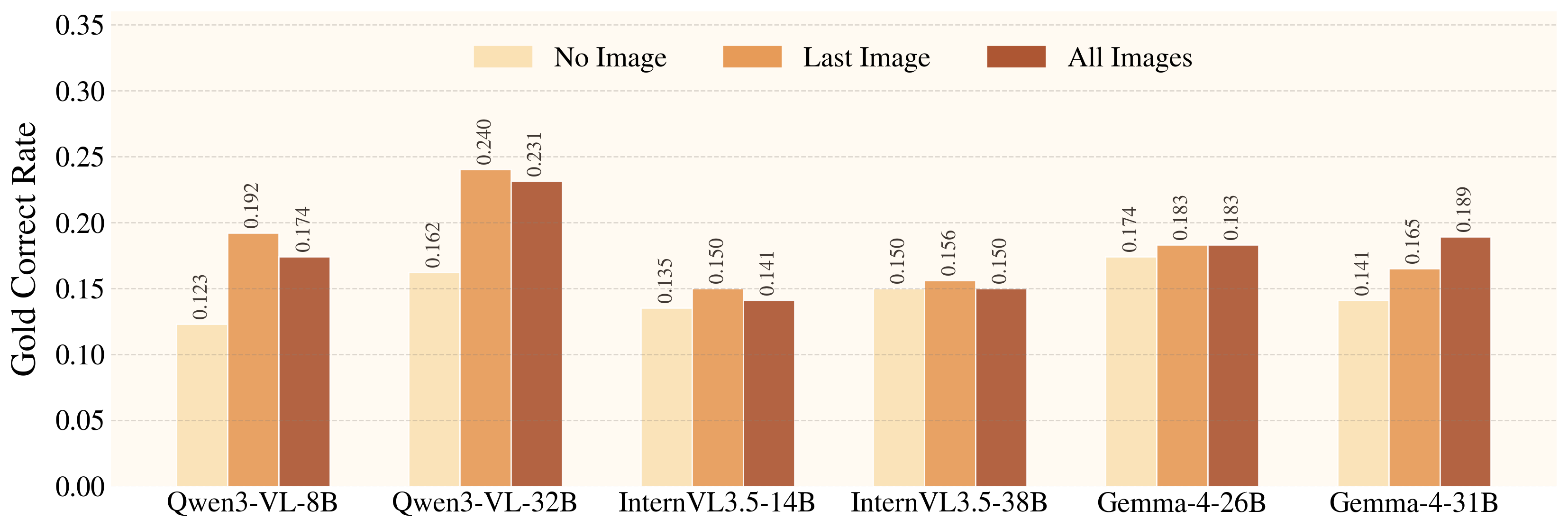}
    \vspace{-5mm}
    \caption{Gold correct rate across different input image conditions.}
    \vspace{-3mm}
    \label{fig:img_condition}
\end{wrapfigure}

We further analyze the impact of visual input under different image conditions. As shown in \Cref{fig:img_condition}, performance generally improves when visual information is available: the \textbf{No Image} setting generally yields the lowest gold correct rate across models, showing that MM-CreativityBench requires visual grounding and cannot be reliably solved from text description alone. Providing visual context, either through the \textbf{Last Image} or \textbf{All Images} condition, leads to gains, although the magnitude of improvement varies across models.

Notably, models with longer average interaction horizons, such as Qwen3-VL (13.45 turns for 8B and 8.77 turns for 32B) and InternVL (6.99 turns for 38B), benefit more from the \textbf{Last Image} condition, which often matches or outperforms the \textbf{All Images} setting. This suggests that when a model can iteratively refine its belief over candidate entity--part pairs $(e,p)$, access to the most recent and task-relevant visual observation $I_t$ is often sufficient for grounded decision-making. Overall, these results show that grounded creative problem solving depends on both \textbf{access to visual evidence} and the \textbf{ability to incorporate it through interaction}.

\subsection{Impact of prompting and format strategy in training}

\begin{table*}[t]
    \centering
    \vspace{-4mm}
    \setlength{\tabcolsep}{3pt}
    \renewcommand{\arraystretch}{1.12}
    
    \newcolumntype{C}[1]{>{\centering\arraybackslash}p{#1}}
    
    \resizebox{\linewidth}{!}{%
    \begin{tabular}{
        l
        C{1.15cm} C{1.25cm} C{0.9cm}
        @{\hspace{8mm}}
        C{1.35cm} C{1.35cm}
        @{\hspace{10mm}}
        C{2.25cm} C{2.25cm}
        @{\hspace{12mm}}
        C{2.25cm} C{2.25cm}
    }
        \toprule
        \multirow{2.5}{*}{\textbf{Model}}
        & \multirow{2.5}{*}{\textbf{\makecell{Gold\\Correct}}}
        & \multirow{2.5}{*}{\textbf{\makecell{Entity\\Correct}}}
        & \multirow{2.5}{*}{\textbf{Turns}}
        & \multicolumn{2}{c}{\hspace*{-5mm}\textbf{Avg. Distinct Explored}\hspace*{5mm}}
        & \multicolumn{2}{c}{\hspace*{-5mm}\textbf{Gold Entity Explored Before Answer}\hspace*{5mm}}
        & \multicolumn{2}{c}{\hspace*{-5mm}\textbf{Gold Part Explored Before Answer}\hspace*{5mm}} \\

        \cmidrule(l{-0.8em}r{2.2em}){5-6}
        \cmidrule(l{-0.8em}r{2.2em}){7-8}
        \cmidrule(l{-0.8em}r{2.2em}){9-10}

        & & & 
        & \textbf{Entities}
        & \textbf{Parts}
        & \textbf{\makecell{Entity Correct}}
        & \textbf{\makecell{Entity Wrong}}
        & \textbf{\makecell{Part Correct}}
        & \textbf{\makecell{Part Wrong}} \\
        
        \midrule
        \makecell[l]{\textbf{Qwen3-4B-VL-Instruct}} & 0.156 & 0.393 & 18.922 & 3.937 & 4.417 & 0.947 & 0.554 & 0.923 & 0.167 \\
        \makecell[l]{\quad + SFT} & 0.252 & 0.489 & 17.039 & 6.640 & 9.039 & 1.000 & 1.000 & 0.905 & 0.309 \\
        \makecell[l]{\quad + DPO (normal negative)} & 0.210 & 0.441 & 24.524 & 4.631 & 8.854 & 0.973 & 0.751 & 0.971 & 0.264 \\
        \makecell[l]{\quad + DPO (hard negative)} & 0.249 & 0.483 & 12.864 & 4.123 & 4.060 & 0.969 & 0.637 & 0.916 & 0.141 \\
        \makecell[l]{\quad + SFT + DPO (hard negative)} & 0.384 & 0.574 & 8.675 & 3.223 & 2.428 & 0.969 & 0.404 & 0.828 & 0.044 \\
        \midrule
        \makecell[l]{\textbf{Qwen3-8B-VL-Instruct}} & 0.192 & 0.441 & 13.450 & 4.979 & 3.766 & 0.993 & 0.747 & 0.953 & 0.201 \\
        \makecell[l]{\quad + SFT} & 0.234 & 0.429 & 15.946 & 6.646 & 7.970 & 1.000 & 0.995 & 0.949 & 0.310 \\
        \makecell[l]{\quad + DPO (normal negative)} & 0.255 & 0.477 & 18.851 & 5.696 & 5.854 & 0.994 & 0.841 & 0.976 & 0.197 \\
        \makecell[l]{\quad + DPO (hard negative)} & 0.270 & 0.498 & 9.922 & 4.799 & 2.838 & 0.988 & 0.749 & 0.944 & 0.107 \\
        \makecell[l]{\quad + SFT + DPO (hard negative)} & 0.345 & 0.571 & 8.136 & 4.364 & 2.334 & 1.000 & 0.606 & 0.930 & 0.083 \\
        \bottomrule
    \end{tabular}%
    }
    \caption{Varying the prompt to require pure JSON outputs does not change the overall trend. Across both raw and SFT settings, SFT + DPO consistently achieves the highest gold-correct rate while generally requiring fewer turns, suggesting that training leads to more effective and targeted exploration.}
    \label{tab:image_eval_sft_promptvary}
    \vspace{-0mm}
\end{table*}

To examine whether the prompting format affects evaluation outcomes, we compare the original prompting setting, where the model first performs free-form reasoning and then emits a structured JSON action, with a stricter pure-JSON variant, where the model is instructed to place both reasoning and the next-step decision inside a JSON object. As shown in \Cref{tab:image_eval_sft_promptvary}, overall, \textbf{the two settings exhibit highly consistent trends}. In both prompt formats, training improves the base models substantially, and the strongest performance is obtained by the two-stage SFT+DPO setting with hard negatives. For Qwen3-4B-VL, SFT+DPO achieves the best gold-correct rate under both prompts, increasing from 0.156 to 0.417 in the original setting and from 0.156 to 0.384 in the pure-JSON setting. Similarly, for Qwen3-8B-VL, SFT+DPO remains the best-performing method, reaching 0.393 under the original prompt and 0.345 under the pure-JSON prompt. These results indicate that the observed gains are not merely artifacts of a particular output format; rather, they reflect a robust improvement in the model's ability to conduct targeted exploration and produce correct final answers.

At the same time, the prompt variation introduces some quantitative shifts in behavior. The pure-JSON prompt slightly changes the balance between exploration breadth and answer efficiency. For example, the 4B SFT+DPO model uses more turns under the pure-JSON setting than under the original setting, increasing from 6.211 to 8.675 turns, while still maintaining a strong gold-correct rate. The 8B SFT+DPO model shows a similar but smaller pattern, with turns remaining nearly unchanged while the gold-correct rate decreases moderately from 0.393 to 0.345. Pure-JSON prompting also tends to preserve the relative advantage of hard-negative DPO over normal-negative DPO, especially in reducing excessive exploration and improving final-answer accuracy. Across both tables, correct predictions are still associated with substantially higher rates of exploring the gold entity and gold part before answering, whereas wrong predictions show much lower grounding rates. This suggests that the central mechanism remains unchanged across prompting formats: successful models are those that identify and inspect the relevant visual evidence before committing to an answer. Therefore, although enforcing a pure-JSON format can slightly affect absolute scores and exploration patterns, it does not alter the main conclusion that \textbf{SFT+DPO with hard negatives yields more effective, better-grounded, and more efficient interactive exploration}.

\begin{figure}[!t]
    \centering
    \vspace{-1mm}
    \includegraphics[width=\linewidth]{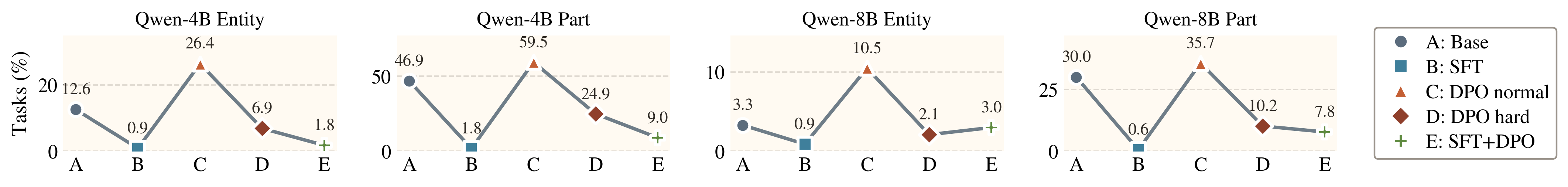}
    \vspace{-4mm}
    \caption{\textbf{Exploration repetition rates} across the base and trained 4B and 8B models. SFT and SFT+DPO substantially reduce repetition, indicating clearer state tracking and more effective, efficient exploration.}
    \vspace{-3mm}
    \label{fig:analysis_repetition}
\end{figure}

\begin{figure}[!t]
    \centering
    \vspace{0mm}
    \includegraphics[width=\linewidth]{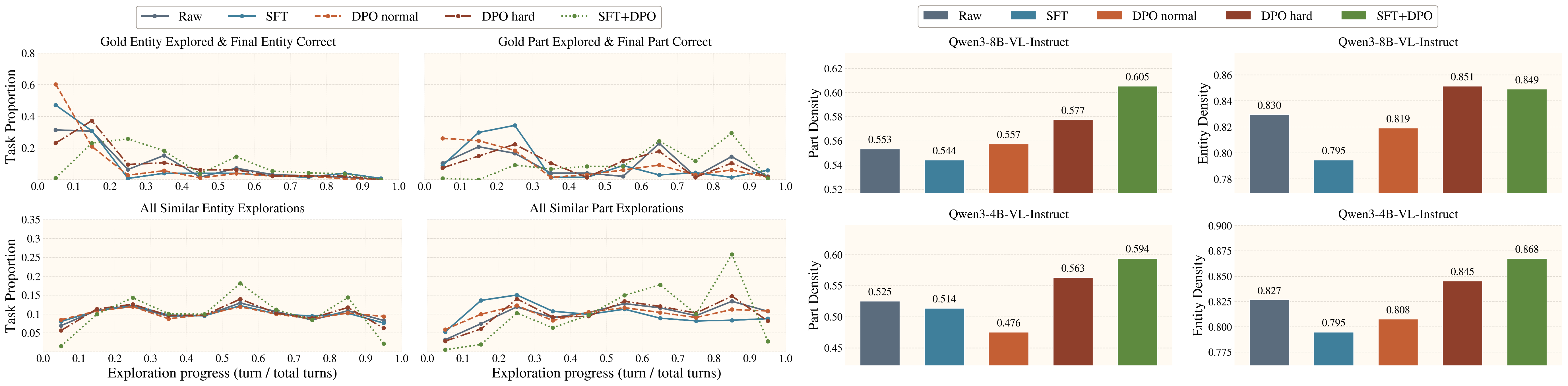}
    \vspace{-4mm}
    \caption{\textbf{Exploration progress and similarity density} for Qwen3-VL variants. Left: Qwen3-4B timing distributions for gold and similar entity/part explorations. Right: average part/entity similarity density for Qwen3-8B and Qwen3-4B across Raw, SFT, DPO, and SFT+DPO variants.}
    \vspace{-2mm}
    \label{fig:analysis_similarity}
\end{figure}

\subsection{How training improves exploration efficiency and semantic focus}
We evaluate whether training improves not only final accuracy but also the quality of intermediate exploration, using three trajectory-level metrics. \textbf{Repetition rate} measures the percentage of tasks in which the model revisits an already explored entity or part; lower values indicate better state tracking and fewer wasted turns. \textbf{Similarity density} measures how concentrated exploration is around useful hypotheses: part-level density is the fraction of explored parts that are gold or affordance-similar to the gold part, while entity-level density is the fraction of explored entities that are gold or contain at least one affordance-similar part. Finally, \textbf{exploration progress} records each discovery by its normalized turn index, $\text{turn}/\text{total turns}$, indicating when the model identifies gold or affordance-similar candidates.

\Cref{fig:analysis_repetition} shows that SFT, DPO with hard negatives, and SFT+DPO all substantially reduce redundant exploration. For example, on Qwen3-4B, SFT+DPO reduces part repetition from 46.9\% to 9.0\%, and entity repetition from 12.6\% to 1.8\%; a similar trend holds for Qwen3-8B, where part repetition drops from 30.0\% to 7.8\%. These reductions suggest that our training improves not only final accuracy but also the model’s ability to \textbf{maintain an exploration state} and avoid revisiting inspected entities or parts. This likely comes from our positive-data construction, which explicitly includes an exploration stack and thereby supervises state tracking during action selection. In contrast, DPO with normal negatives is less stable without this structured SFT prior, often leading to more repeated and inefficient exploration.

\Cref{fig:analysis_similarity} further explains why SFT+DPO is preferable to SFT alone. As shown in the right four panels, SFT reduces redundant exploration but can also make the search narrow, resulting in lower part and entity similarity density than the base model. Adding DPO restores semantic focus, helping the model prioritize affordance-relevant candidates; for example, SFT+DPO achieves the highest part similarity density for both Qwen3-4B and Qwen3-8B, reaching 0.594 and 0.605, respectively. The progress curves show that this gain does not come from premature guessing: unlike other variants that mostly find useful candidates early, SFT+DPO continues to discover similar entities and parts throughout the trajectory, especially during later part-level exploration. Overall, the two stages are complementary: \textbf{SFT teaches disciplined, non-redundant exploration, while DPO redirects that exploration toward semantically useful hypotheses}.

\begin{takeaway}
SFT+DPO produces the strongest exploration behavior: it sharply reduces repeated queries while increasing the density of affordance-relevant entities and parts explored. This suggests that the final model is not only more accurate, but also searches more deliberately and efficiently.
\end{takeaway}

\begin{figure}[!t]
    \centering
    \vspace{0mm}
    \includegraphics[width=\linewidth]{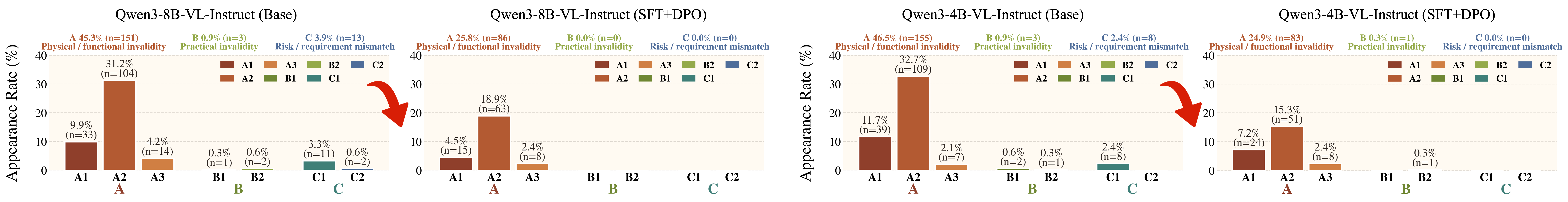}
    \vspace{-4mm}
    \caption{\textbf{Primary error-category rates} reveal that SFT+DPO substantially reduces physical/functional invalidity errors, especially affordance mismatch (A2), and removes most practical and risk/constraint errors.}
    \vspace{-4mm}
    \label{fig:analysis_error}
\end{figure}

\subsection{Error Category Analysis}
\label{sec:error_analysis}
We categorize incorrect predictions by their primary failure reason. Category A covers \textbf{physical or functional invalidity}: hallucinated affordances (A1), affordance mismatch caused by unsuitable geometry/material/mechanics (A2), and performance shortfalls where the predicted part is only partially suitable but lacks sufficient stability, reach, capacity, precision, or retention (A3). Category B covers \textbf{practical infeasibility}, including destructive workarounds (B1) and context/accessibility issues (B2). Category C covers \textbf{risk or requirement mismatch}, including safety or damage risks (C1) and explicit constraint violations (C2). We use GPT-5.4 for scalable automatic categorization; details are provided in \Cref{apdx:error_analysis_details}.

\Cref{fig:analysis_error} shows that SFT+DPO primarily reduces the dominant failure mode: errors in physical affordance grounding. For Qwen3-8B, total Category A errors decrease from 45.3\% in the base model to 25.8\% after SFT+DPO, while Category B and C errors are eliminated. The largest reduction occurs in A2, affordance mismatch, which drops from 31.2\% to 18.9\%. This indicates that the trained model is less likely to select parts whose shape, material, or mechanics are incompatible with the intended use. Qwen3-4B exhibits the same trend. Thus, the improvement is not merely a broad accuracy gain or the removal of rare practical and safety mistakes; it directly targets the main bottleneck in the task. The remaining errors are still mostly A-type, suggesting that physical affordance judgment remains challenging, but SFT+DPO substantially improves the model's ability to choose mechanically plausible parts.

\subsection{Case Study: What SFT+DPO Repairs in Interactive Image Reasoning}
\label{sec:case_study}

We analyze three representative cases from the evaluation set. These cases are selected to isolate distinct effects of training: (i) grounding a solution in material and contact attributes, (ii) distinguishing a local anti-slip cue from the global geometry required by the task, and (iii) selecting the safe, task-relevant part within an otherwise plausible tool. See \Cref{apdx:case_analysis_details} for the key original responses, feedback, and images from the decisive turns.

\textbf{Case 1: pressure-spreading pad vs. generic softness.}
\begin{itemize}[topsep=-1.5pt, leftmargin=10pt, itemsep=0pt]
\item \textbf{Task.} A metal towel hook is pressing into painted bathroom drywall, and the user needs a small protector to prevent dents or scratches. The gold answer is \texttt{curved tension shower curtain rod} / \texttt{non\_slip\_end\_pads}.

\item \textbf{Bad trace behavior.} Base 8B fixates on the towel after inspecting \texttt{microfiber\_pile\_surface}. This answer is superficially plausible because a towel is soft, but the reasoning trace is flawed: it treats \emph{softness} as the entire relevant affordance. The model never inspects the shower rod's \texttt{non\_slip\_end\_pads}, and it incorrectly rejects the shower rod as rigid even though the relevant subpart is rubber.

\item \textbf{Good trace behavior.} SFT+DPO 8B initially inspects the same towel part, but continues searching instead of stopping at the first plausible cue. It then inspects the shower rod, discovers \texttt{non\_slip\_end\_pads}, and examines that subpart directly. The trained answer compares the two candidates: the towel is soft, but the rubber pad is smaller, more durable, higher-friction, and better suited to remaining fixed at the pressure point.

\item \textbf{Why the trained trajectory is better.} The trained model grounds its solution in the physical contact model of the task: a hard hook creates a localized point load on painted drywall. A bulky towel may bunch, compress, or slip, whereas a rubber pad can remain at the contact point and distribute force more effectively. \textbf{Representative improved capability: material/contact attribute grounding.}
\end{itemize}

\textbf{Case 2: straight-edge geometry vs. local anti-slip grip.}
\begin{itemize}[topsep=-1.5pt, leftmargin=10pt, itemsep=0pt]
\item \textbf{Task.} The user needs to trim wrapping paper without a ruler or cutting mat, while keeping the paper from slipping during marking. The gold answer is \texttt{under-bed storage bin with zipper lid} / \texttt{lid\_panel}.

\item \textbf{Bad trace behavior.} Base 4B repeatedly inspects the pen \texttt{rubber\_grip\_sleeve} and nearby mask parts until the 50-turn budget is exhausted. Base 8B makes the same error more quickly. Both models solve only a local subproblem: a small rubber grip may increase friction at a point of contact, but it cannot provide the long straight edge or stable backing surface required for trimming a sheet of paper.

\item \textbf{Good trace behavior.} SFT+DPO 8B immediately inspects the storage bin and its \texttt{lid\_panel}. The feedback identifies the part as a semi-rigid, sturdy, non-elastic panel with an internal polypropylene stiffener. The trained answer therefore uses this part as both a flat backing surface and an alignment guide.

\item \textbf{Why the trained trajectory is better.} The trained model recovers the global geometry of the task. The user needs support and alignment for an extended sheet, not merely a high-friction contact patch. \textbf{Representative improved capability: interaction planning and part-level geometry.}
\end{itemize}

\textbf{Case 3: safe multi-tooth probing vs. sharp scraping.}
\begin{itemize}[topsep=-1.5pt, leftmargin=10pt, itemsep=0pt]
\item \textbf{Task.} Damp hair-and-soap buildup blocks a narrow sink overflow slot, and the user needs to loosen the material enough to rinse it out. The gold answer is \texttt{electric beard trimmer with adjustable guard} / \texttt{adjustable\_guard\_comb}.

\item \textbf{Bad trace behavior.} Base 4B first considers the trimmer blade head, then inspects a double-edge razor and answers with \texttt{double\_edge\_blade}. Base 8B remains within the trimmer entity but still chooses \texttt{cutting\_blade\_head}. Both traces reveal a sharp-tool prior: the models assume that a narrow clogged slot calls for scraping or cutting, even though the target material is soft buildup in a constrained opening.

\item \textbf{Good trace behavior.} SFT+DPO 4B also explores the tempting razor blade, but it does not stop there. After receiving feedback on the razor, it inspects the trimmer and then the \texttt{adjustable\_guard\_comb}. The final answer selects the comb because its rigid, fine plastic teeth are exposed, fit the narrow opening, and can wiggle or rake soft buildup without the risks associated with a blade.

\item \textbf{Why the trained trajectory is better.} This case illustrates an exact-part and safety repair. The trimmer blade head and guard comb belong to the same entity, but they imply different actions, contact mechanics, and risks. The trained model selects the part whose geometry and material match the desired interaction while avoiding unnecessary sharpness. \textbf{Representative improved capability: exact-part discrimination under safety and state constraints.}
\end{itemize}
\section{Discussion}
\label{sec:discussion}

\textbf{Difference between Creativity and Hallucination.}
In MM-CreativityBench, creativity is not defined as unconstrained novelty, but as the ability to discover non-obvious uses of visually available objects through physically grounded affordance reasoning. This makes it fundamentally different from the kind of imaginative generation that may be valuable in creative writing, design ideation, or open-ended research, where productive ``hallucination'' can sometimes serve as a source of exploration. In our setting, hallucination is not a creative act but a grounding failure: the model invents an unsupported attribute, assumes an unseen part, or maps a plausible function onto an object whose geometry, material, state, or accessibility does not justify that use. The useful form of imagination here is therefore conditional and verifiable. A model may hypothesize that a serrated edge could cut tape or that a rubber pad could protect a wall, but this hypothesis must be checked against the inspected visual evidence and the task constraints. Thus, MM-CreativityBench studies \textbf{evidence-seeking creativity}: novelty arises from recombining observed object properties with task goals, while validity depends on sustained interaction, part-level inspection, and physical plausibility. This distinction becomes increasingly important for embodied agents, where hallucinated affordances are not merely incorrect answers but potentially unsafe actions in the physical world.

\textbf{Enhancement of Model Creativity.}
Our results suggest that improving grounded creativity requires training models not only to produce novel final answers, but to acquire better exploration and verification policies. Standard outcome-driven training can reward a correct-looking solution without teaching the model how to inspect the scene, compare competing affordances, or reject superficially plausible but unsupported evidence. In MM-CreativityBench, SFT provides a useful first step by teaching structured exploration over entities and parts, while hard-negative DPO further teaches the model to discriminate grounded trajectories from fluent but misleading ones. This points toward a broader direction for future RL: creativity should be optimized as an interactive evidence-gathering process rather than as single-turn answer selection. A suitable objective would reward information-seeking actions, part-sensitive visual grounding, causal or mechanical consistency, and timely commitment once sufficient evidence has been collected, while penalizing unsupported affordance claims, redundant exploration, and premature answers. The affordance knowledge base provides a practical substrate for such training because it can generate positive trajectories, plausible hard negatives, and fine-grained attribute--affordance contrasts. More broadly, future creative agents may need RL objectives that preserve diversity in hypothesis generation while enforcing strict grounding at the point of action, allowing models to explore unusual solutions without drifting into hallucinated physical assumptions.

\section{Conclusion}
\label{sec:conclusion}
We introduced MM-CreativityBench, a benchmark for evaluating visually grounded creative tool repurposing in multimodal environments. By requiring models to interactively inspect scenes, entities, and parts, the benchmark reveals that current LMMs often struggle with grounding creative solutions in fine-grained visual and physical evidence. We further showed that affordance-grounded alignment, especially with hard negative preference signals, can improve both accuracy and exploration efficiency. Looking forward, we hope MM-CreativityBench will support future work on multimodal agents that can reason more robustly about physical affordances, adapt to unfamiliar environments, and solve open-ended problems through grounded exploration rather than surface-level plausibility.

\bibliographystyle{plainnat}
\bibliography{neurips_2026}


\clearpage
\appendix

\section*{Appendix}
\label{sec:appendix}

\section{Significance, Scope, and Clarifications}

\subsection{Why MM-CreativityBench Matters}

\textbf{A Missing Dimension in Multimodal Evaluation.}
Recent progress in large multimodal models has been evaluated primarily through recognition, visual question answering, spatial reasoning, and instruction following. These settings are important, but they do not fully test whether a model can use perception as evidence for discovering non-obvious but physically feasible solutions. MM-CreativityBench focuses on this under-measured capability: \textbf{visually grounded creative tool repurposing}. The central question is not whether a model can describe a scene or generate a plausible solution in language, but whether it can inspect the environment, identify the relevant object and part, and justify how observable physical properties support an unconventional use. This makes the benchmark a focused test of creative intelligence under perceptual and physical constraints.

\textbf{From Static Image Understanding to Evidence-Driven Exploration.}
A key contribution of MM-CreativityBench is that it moves beyond static image-to-answer evaluation. In real-world problem solving, an agent often does not know in advance which object or region matters. It must search, inspect, compare, and revise its hypothesis before committing to a solution. Our interactive protocol captures this process by allowing models to inspect the scene, candidate entities, and zoomed-in parts. This design makes it possible to evaluate not only the final answer, but also whether the answer emerges from \textbf{grounded exploration} rather than unsupported guessing. In this sense, the benchmark evaluates creative tool use as a process, not merely as a final textual output.

\textbf{Part-Level Grounding as the Core Challenge.}
MM-CreativityBench is designed to expose a specific weakness in current LMMs: models often identify a generally relevant object but fail to determine which part, attribute, or visual cue actually enables the intended use. This distinction is important because creative repurposing depends on mechanism-level reasoning. A key's usefulness for opening a taped box, for example, does not come from the object category ``key'' alone, but from a visually grounded property such as a thin, rigid, serrated edge. By requiring models to ground answers at the entity--part level, MM-CreativityBench separates coarse semantic plausibility from genuine physical affordance reasoning.

\textbf{A Controlled Testbed for Grounded Creative Reasoning.}
The benchmark is built on a structured affordance knowledge base that links entities, parts, attributes, and affordances. This structure gives the benchmark several advantages. First, it supports systematic task construction while preserving interpretable solution paths. Second, it enables controlled multimodal augmentation at multiple levels of granularity, including scene images, entity views, and part-level close-ups. Third, it makes failure analysis more diagnostic: when a model fails, we can ask whether it overlooked the correct entity, inspected the wrong part, misread visual evidence, hallucinated an attribute, or failed to connect an observed attribute to an affordance. This level of diagnosis is difficult in fully open-ended creativity benchmarks.

\textbf{Connecting Evaluation with Model Improvement.}
MM-CreativityBench is not only an evaluation benchmark. It also provides a framework for studying how grounded creative behavior can be improved. Our affordance-aware alignment results show that models can learn more effective exploration and more reliable attribute--affordance reasoning when trained with structured positive trajectories and contrastive negative trajectories. The goal is not to claim that SFT or DPO is a new optimization algorithm, but to show that \textbf{affordance-grounded supervision} is a useful training signal for multimodal creative problem solving. This connection between benchmark design, failure analysis, and targeted alignment makes the benchmark valuable as a research tool rather than only as a leaderboard.

\textbf{Implications for Future Multimodal Agents.}
The broader significance of MM-CreativityBench lies in its relevance to adaptive agents. Agents operating in homes, labs, workshops, or other resource-limited environments will often need to repurpose available objects rather than rely on predefined tools. Such behavior requires visual grounding, physical commonsense, comparative search, and flexible recombination of affordances. MM-CreativityBench isolates this ability in a controlled and reproducible form. It therefore offers a concrete step toward evaluating and improving multimodal systems that can solve unfamiliar problems using evidence from their surroundings.

\subsection{Clarifications of Concerns}

\textbf{MM-CreativityBench Targets Constrained Creativity, Not All Creativity.}
Creativity is broad, and we do not claim that creative intelligence can be fully captured by a single benchmark. Our focus is deliberately narrower: \textbf{constrained, visually grounded tool repurposing}. In this setting, a solution must be novel relative to canonical object use, but also physically feasible and supported by visual evidence. This operational definition is useful precisely because it avoids the ambiguity of fully open-ended creativity evaluation. Rather than asking whether a model is creative in the abstract, MM-CreativityBench asks whether it can discover a usable object--part solution under explicit perceptual and functional constraints.

\textbf{Why This Is More Than Physical Commonsense Retrieval.}
Physical commonsense is a necessary ingredient, but the benchmark requires more than retrieving a familiar object-use association. Each task requires the model to connect a goal with a specific object, a specific part, observable attributes, and a feasible mechanism of use. The model must also make this connection through an interactive visual process. This is why performance drops substantially when moving from entity-level correctness to gold entity--part correctness: models often know which object category is plausible, but fail to ground the precise part and attribute that make the solution work. The benchmark therefore targets \textbf{mechanism-sensitive visual repurposing}, not simple commonsense recall.

\textbf{Synthetic Images as Controlled Visual Grounding, Not a Claim of Full Realism.}
The use of generated images is a methodological choice for controlled evaluation. Real images would introduce substantial noise in object availability, occlusion, scale, lighting, and part visibility, making it difficult to isolate the reasoning problem. In contrast, generated scenes allow us to construct environments where candidate entities are present, parts can be inspected, and the underlying affordance structure is known. We do not claim that generated images replace real-world embodied evaluation. Rather, they provide a reproducible intermediate setting that tests whether models can use visual evidence when the relevant physical cues are available. This makes MM-CreativityBench a diagnostic benchmark for visual-affordance reasoning, complementary to future evaluations in real physical environments.

\textbf{Image Generation Does Not Turn the Task into Visual Leakage.}
A possible concern is that generated images may make the gold answer visually obvious. Our intention is the opposite: image generation is used to make relevant physical evidence inspectable, not to mark the solution. The model still needs to search among multiple entities, inspect candidate parts, and infer which observed attributes support the target affordance. The scene is not allowed to depict task execution, and the interaction protocol requires the model to justify the final answer through object-specific evidence. More generally, MM-CreativityBench evaluates whether a model can transform visible properties into functional hypotheses; making those properties visible is necessary for testing grounding rather than language-only guessing.

\textbf{Single-Gold Evaluation as Measurement Control.}
Creative tool use is naturally open-ended, and multiple valid solutions may exist in the real world. However, benchmark evaluation requires a controlled target so that models can be compared consistently. The gold answer in MM-CreativityBench should therefore be understood as a verified solution path, not as a claim that no other solution could ever work. The strict gold metric is intentionally conservative: it measures whether the model recovers the intended object--part mechanism under the constructed scene and constraints. Entity-level accuracy, exploration statistics, and grounding metrics complement this strict score by showing where the reasoning process succeeds or fails. Thus, the single-gold structure supports measurement clarity while preserving the broader view that creativity can admit multiple solutions.

\textbf{The Interactive Protocol Is Artificial by Design, but Diagnostic.}
The entity and part inspection interface abstracts away low-level perception problems such as segmentation and camera control. This may appear less realistic than open-world embodied interaction, but it is useful for isolating the central question of the paper: can a model conduct evidence-driven creative reasoning once the environment is inspectable? By structuring interaction into scene, entity, and part views, the benchmark makes exploration behavior observable and comparable across models. This design reduces confounds from visual localization failures and allows us to measure whether models actually inspect the evidence needed for their final answers. The protocol should be viewed as a controlled diagnostic environment, not as a complete simulation of embodied deployment.

\textbf{Affordance-Knowledge Supervision Is a Research Signal, Not Test-Time Privilege.}
The affordance knowledge base is used to construct tasks and training trajectories, but the evaluation protocol does not give models access to gold affordance labels or hidden solution paths. At test time, models must operate from the visible scene, entity names, part names, and inspected images. The purpose of knowledge-base-derived supervision is to teach models useful intermediate behavior: how to search, how to compare candidate parts, and how to reject visually unsupported affordances. This is analogous to using structured annotations to train better visual reasoning models. The important question is not whether the supervision contains knowledge, but whether that knowledge improves unguided visual-affordance reasoning on disjoint test instances.

\textbf{SFT and DPO Are Not Claimed as Algorithmic Novelty.}
Our contribution is not a new fine-tuning objective. SFT and DPO are used as established tools to test a specific hypothesis: grounded creative tool use can be improved by aligning models toward visually supported attribute--affordance reasoning and away from plausible but ungrounded alternatives. The novelty lies in the problem formulation, the interactive visual benchmark, the construction of affordance-grounded trajectories, and the use of hard negative trajectories that target hallucinated or misleading affordance reasoning. In this sense, the training experiments serve as evidence that the benchmark identifies a learnable failure mode, rather than merely reporting that current models perform poorly.

\textbf{Cross-Model Comparisons Should Be Interpreted Structurally.}
Absolute scores can be affected by prompting, model family, decoding details, and the interaction interface. Therefore, the main conclusion should not rest on a brittle ranking between two models. The more important result is the recurring structural pattern: models often perform better at coarse entity localization than at fine-grained part grounding; longer exploration does not necessarily produce better answers; visually plausible but unsupported reasoning remains common; and targeted affordance-aware alignment improves both accuracy and efficiency. These patterns are more informative than any single leaderboard position, because they reveal a shared limitation in current LMMs.

\textbf{Final Take Away.}
MM-CreativityBench is intentionally controlled, visually grounded, and diagnostic. It does not claim to cover all forms of creativity, nor does it replace real-world embodied evaluation. Its contribution is more precise: it isolates a practically important form of creative problem solving, formalizes it through entity--part--attribute affordance structure, and shows that current LMMs still struggle to connect visual evidence with non-canonical physical use. By combining benchmark construction, interactive evaluation, failure analysis, and affordance-aware alignment, MM-CreativityBench positions creative tool use as a concrete and measurable frontier for multimodal AI.


\section{Preliminary Experiment}
\label{sec:appendix_prelim}

Our preliminary experiment is designed as a controlled comparison between two prompting strategies on 100 creative tool-use tasks sampled from MacGyver~\citep{tian2024macgyver}. To evaluate the creativity under a multimodal environment, where the model directly perceives necessary physical attributes from the input image and reasons about creative tool repurposing, we first generate a scenario image using Gemini-2.5-Pro. Then, update the task description to only include the constraints that cannot be represented visually. In the \textbf{direct prompt} setting, given an input task and scenario image, the model is asked to propose a feasible solution under the task constraints without any prescribed intermediate reasoning steps, thereby testing its implicit ability to connect task requirements with tool functions. In the \textbf{structured affordance-level CoT} setting, the model is instead guided through an explicit reasoning pipeline that includes listing the available tools, decomposing each tool into parts, inferring relevant physical properties, deriving possible affordances, justifying each action step, and validating the final solution against the stated constraints.

We evaluate outputs using six criteria: \textbf{Correctness}, \textbf{Feasibility}, \textbf{Physical Grounding}, \textbf{Constraint Coverage}, \textbf{Tool Usage}, and \textbf{Creativity}, under pairwise relative comparison between the two prompting strategies. Please see the prompts below for more details. We use GPT-4.1-mini as the target model and GPT-5.2 as the judge model, employing temperature 0.0 to guarantee deterministic outputs. All of other settings we follow the original MacGyver paper's protocols.

\begin{tcolorbox}[
  enhanced,
  breakable,
  width=0.98\linewidth,
  colback=tzBlueFill,
  colframe=tzBlueBorder,
  boxrule=1.2pt,
  arc=6pt,
  left=5pt,right=5pt,top=4pt,bottom=2pt,
  title={\small Scenario Image Generation Prompt},
  coltitle=white,
  colbacktitle=tzBlueHeader2,
  fonttitle=\bfseries,
]
\small
\begin{lstlisting}[style=jsonTiny]
Create one realistic scene that shows only the initial setup of this problem before any actions are taken. Focus on the environment, relevant objects, and constraints. Do not include text overlays, labels, or solution steps. 

Problem: {task description from MacGyver}
\end{lstlisting}
\end{tcolorbox}

\begin{tcolorbox}[
  enhanced,
  breakable,
  width=0.98\linewidth,
  colback=tzBlueFill,
  colframe=tzBlueBorder,
  boxrule=1.2pt,
  arc=6pt,
  left=5pt,right=5pt,top=4pt,bottom=2pt,
  title={\small Direct Prompt},
  coltitle=white,
  colbacktitle=tzBlueHeader2,
  fonttitle=\bfseries,
]
\small
\begin{lstlisting}[style=jsonTiny]
You are an expert at creative physical tool-use reasoning.
Given the task description below and image, produce a feasible solution.

Rules:
1. Use only tools/items explicitly available in the input image.
2. Respect physical constraints if the task description restricts physical attributes or image shows physical constraints such as size or state.
3. Invent new tools using tools/items explicitly available in the input image if it is needed.
4. Provide practical steps that can actually be executed.
5. Do not involve any unnecessary steps to achieve the task's goal.
6. If a complete solution is impossible, return the best partial plan and explain why it cannot be completed.

TASK DESCRIPTION:
{problem}

Return JSON:
{{
  "solvable": "Yes or No",
  "solvable_explanation": "1-3 sentences about why the given task is solvable or not",
  "solution_steps": ["Step 1: ...", "Step 2: ...", ...],
  "final_solution": "One concise paragraph that summarizes the full approach.",
  "used_tools": ["tool 1", "tool 2", ...],
  "constraint_handling": [{{"constraint": "...", "handling": "..."}}, ...]
}}
\end{lstlisting}
\end{tcolorbox}

\begin{tcolorbox}[
  enhanced,
  breakable,
  width=0.98\linewidth,
  colback=tzBlueFill,
  colframe=tzBlueBorder,
  boxrule=1.2pt,
  arc=6pt,
  left=5pt,right=5pt,top=4pt,bottom=2pt,
  title={\small CoT Prompt},
  coltitle=white,
  colbacktitle=tzBlueHeader2,
  fonttitle=\bfseries,
]
\small
\begin{lstlisting}[style=jsonTiny]
You are an expert at creative physical tool-use reasoning.
Solve the task by explicitly reasoning over tool parts and affordances under constraints.

Rules:
1. Use only tools/items explicitly available in the input image.
2. Respect physical constraints if the task description restricts physical attributes or image shows physical constraints such as size or state.
3. Invent new tools using tools/items explicitly available in the input image if it is needed.
4. Provide practical steps that can actually be executed.
5. Do not involve any unnecessary steps to achieve the task's goal.
6. If a complete solution is impossible, return the best partial plan and explain why it cannot be completed.

Required reasoning procedure:
1. State the task goal and concrete success condition.
2. List all available tools/items from the input image (no additions).
3. For each relevant tool, identify the key part(s), infer physical properties, and derive part-level affordances useful for this task.
4. Build a step-by-step plan where each step references tool parts and the affordance being used.
5. Validate each step against stated constraints (e.g., broken/unusable items, size mismatch, blocked function, state limitations).
6. Keep the plan practical and minimal with no unnecessary actions.

TASK DESCRIPTION:
{problem}

Return JSON:
{{
  "task_goal": "...",
  "success_condition": "...",
  "identified_constraints": ["...", "..."],
  "available_tools": [
    {{
      "tool": "...",
      "relevant_parts": [
        {{
          "part": "... or NA",
          "inferred_physical_properties": ["...", "..."],
          "affordances_for_task": ["...", "...", ...],
          "usable_under_constraints": "Yes or No",
        }},
        ...
      ]
    }},
    ...
  ],
  "reasoning_plan": [
    {{
      "step": 1,
      "action": "...",
      "tool_parts_used": ["tool:part", "..."],
      "affordance_used": ["...", "..."],
      "explanation": "1 sentence about why it works"
    }},
    {{
      "step": 2,
      "action": "...",
      "tool_parts_used": ["tool:part", "..."],
      "affordance_used": ["...", "..."],
      "explanation": "1 sentence about why it works"
    }},
    ...
  ],
  "solvable": "Yes or No",
  "solvable_explanation": "1-3 sentences about why the given task is solvable or not",
  "solution_steps": ["Step 1: ...", "Step 2: ...", ...],
  "final_solution": "One concise paragraph that summarizes the full approach.",
  "used_tools": ["tool 1", "tool 2", ...],
  "constraint_handling": [{{"constraint": "...", "handling": "..."}}, ...]
  "creative_reasoning_summary": "1-3 sentences about novelty and practicality."
}}
\end{lstlisting}
\end{tcolorbox}

\begin{tcolorbox}[
  enhanced,
  breakable,
  width=0.98\linewidth,
  colback=tzBlueFill,
  colframe=tzBlueBorder,
  boxrule=1.2pt,
  arc=6pt,
  left=5pt,right=5pt,top=4pt,bottom=2pt,
  title={\small Relative Evaluation},
  coltitle=white,
  colbacktitle=tzBlueHeader2,
  fonttitle=\bfseries,
]
\small
\begin{lstlisting}[style=jsonTiny]
You are an expert evaluator for creative physical tool-use reasoning quality.
Your task is to compare TWO candidate solutions for the SAME task, using the ground-truth solution as reference. 

IMPORTANT:
- You are NOT checking wording similarity. Judge functional quality and practical validity.
- A solution can be strong even if it differs from the ground-truth approach.
- Use the following criteria:
    - Correctness: Whether the plan actually solves the task objective.
    - Feasibility: Whether the plan can be physically executed under stated constraints.
    - Physical Grounding: Whether it uses realistic object properties/mechanics correctly.
    - Constraint Coverage: Whether it handles all explicit constraints.
    - Tool Usage: Whether it uses only available tools appropriately and purposefully.
    - Creative Reasoning: Novel, non-obvious but valid repurposing/combination of tools.
    - Overall: Holistic quality across all above dimensions.

For EACH criterion:
1. Determine winner:
   - "win" if solution1_score > solution2_score
   - "lose" if solution2_score > solution1_score
   - "tie" if both are equal
2. Give a short rationale (1-2 sentences).

TASK DESCRIPTION:
{problem}

GROUND-TRUTH SOLUTION:
{ground_truth_solution}

CANDIDATE SOLUTION1 (DEFAULT PROMPT OUTPUT):
{solution1}

CANDIDATE SOLUTION2 (COT PROMPT OUTPUT):
{solution2}

Return STRICT JSON:
{{
  "correctness": {{
    "winner": "win or lose or tie",
    "rationale": "1-2 sentences."
  }},
  "feasibility": {{
    "winner": "win or lose or tie",
    "rationale": "1-2 sentences."
  }},
  "physical_grounding": {{
    "winner": "win or lose or tie",
    "rationale": "1-2 sentences."
  }},
  "constraint_coverage": {{
    "winner": "win or lose or tie",
    "rationale": "1-2 sentences."
  }},
  "tool_usage": {{
    "winner": "win or lose or tie",
    "rationale": "1-2 sentences."
  }},
  "creativity": {{
    "winner": "win or lose or tie",
    "rationale": "1-2 sentences."
  }},
  "overall": {{
    "winner": "win or lose or tie",
    "rationale": "1-2 sentences."
  }},
  "short_summary": "2-4 sentences summarizing key tradeoffs between default and CoT."
}}
\end{lstlisting}
\end{tcolorbox}

\section{Benchmark Construction Details}

\subsection{Affordance Knowledge Base Basis}
\label{apdx:creative_kb}
Our benchmark construction is grounded in an existing open-source affordance knowledge base of physical entities, object parts, and part-level affordances.\footnote{Repository at \url{https://github.com/CreativityBench/CreativityBench}} The knowledge base organizes everyday objects into a structured partonomy: each entity is decomposed into functional parts, and each part is annotated with physical attributes, state attributes, and possible functional affordances. Physical attributes describe relatively stable properties, such as shape, material, rigidity, sharpness, hollowness, flexibility, and surface texture, while state attributes describe situational conditions, such as whether a part is open, clean, intact, accessible, or detachable. The affordance annotations specify what functional roles a part can support, together with use conditions, recipient conditions, examples, and suitability levels. These annotations provide the symbolic basis for MM-CreativityBench: they allow us to identify which part of which object can support a target affordance, what attributes justify this use, and what conditions must hold for the use to be valid. Thus, benchmark instances are not created through unconstrained scenario writing; they are derived from explicit part-level attribute--affordance relations that can be inspected, verified, and converted into multimodal grounding problems.

\subsection{Reverse Task Construction}
\label{apdx:task_construction}
We construct MM-CreativityBench tasks in a reverse direction. Instead of first writing an open-ended scenario and then labeling a possible answer, we begin with a verified entity--part--affordance relation from the knowledge base and generate a scenario around it. For each task, we first sample a target entity $e^*$, a target part $p^*\in P(e^*)$, and a gold affordance $f^*$ supported by the annotated attributes $A(p^*)$. This defines the gold solution
\[
g=(e^*,p^*,f^*).
\]
Here, $e^*$ specifies the object to be repurposed, $p^*$ specifies the decisive part, and $f^*$ specifies the functional role that the part can play in solving the task.

Given the gold solution, we use GPT-5.4 for reverse task proposal generation. Specifically, GPT-5.4 is given $(e^*,p^*,f^*)$ together with the supporting physical and state attributes of $p^*$, and is prompted to propose a practical task description $x$ that requires the affordance $f^*$ without explicitly mentioning the target entity, the target part, or any surface cue that would make the answer trivial. The generated description must satisfy three requirements: it should describe a realistic everyday problem, include only constraints relevant to the intended affordance, and leave the solver to infer which available object and part can satisfy the goal. GPT-5.4 is used only to generate candidate task descriptions; all accepted tasks are subsequently checked and refined by human annotators.

After obtaining a candidate task description, we construct the candidate entity set by adding distractors to the gold entity. For each gold solution, we sample a distractor set $E^-=\{e_1,\ldots,e_{N-1}\}$ and form
\[
E=\{e^*\}\cup E^-.
\]
Distractors are selected from the same knowledge base to create controlled ambiguity. We include \emph{affordance-similar distractors}, whose parts appear functionally related to $f^*$ but fail under closer inspection because they lack a necessary physical attribute, have an incompatible state, provide a weaker mechanism, or violate a contextual requirement. We also include \emph{scene-plausible distractors}, which naturally co-occur with the gold entity in the same environment but do not support the target affordance. This design prevents the task from being solved by object priors or generic tool-use associations alone. A model must inspect candidate entities, compare their parts, and identify the part whose attributes best support the required affordance.

Each symbolic task is represented as
\[
T=(x,E,g), \qquad g=(e^*,p^*,f^*),
\]
where $x$ is the task description, $E$ is the candidate entity set, and $g$ is the gold entity--part--affordance solution. This formulation makes the benchmark an inverse grounding problem: the task description specifies a need, the scene provides multiple possible objects, and the model must recover the correct entity and part by grounding the required affordance in visual and physical evidence.

We apply a multi-stage quality-control process before including a task in the benchmark. First, \textbf{gold validity}: the gold part must physically support the target affordance under the stated task constraints, and the supporting evidence must be present in its knowledge-base attributes. Second, \textbf{distractor separability}: no distractor can serve as an equally valid solution; each distractor must fail for a specific and identifiable reason, such as a missing attribute, incompatible state, weaker functional mechanism, safety concern, or contextual mismatch. Third, \textbf{scenario clarity}: the task description must be natural, concise, and unambiguous, while avoiding direct lexical leakage of the gold entity or part. Fourth, \textbf{scene coherence}: all candidate entities must plausibly co-occur in a single realistic environment without making the scene artificial or visually cluttered. Fifth, \textbf{visual observability}: the decisive part and the attributes required to justify the solution must be inspectable in the generated entity- or part-level images. These criteria remove ambiguous cases, physically invalid solutions, non-visual tasks, and scenarios with unintended alternative answers.

Finally, all tasks are manually verified and refined by human annotators. Annotators check whether the generated task genuinely requires the intended affordance, whether the gold entity--part pair is uniquely valid among the candidates, whether the distractors are plausible but separable, and whether the scenario can be faithfully visualized. When needed, annotators revise the task wording, replace distractors, or discard the example entirely. Using this pipeline, we construct 333 held-out tasks for MM-CreativityBench evaluation and 868 disjoint training tasks for trajectory sampling in the alignment stage. The two splits are separated at the task and visual-instance level to eliminate leakage between training trajectories and benchmark evaluation.

\subsection{Multimodal Image Construction}
\label{apdx:image_construction}
After constructing each symbolic task $T=(x,E,g)$, we convert it into an interactive multimodal instance by generating images at three levels: entity, part, and environment. All images are generated with Gemini-3.1-Image-Pro. The goal is not only to visualize the task, but also to create a controlled evidence hierarchy that matches the benchmark protocol: the model first observes the full environment, then chooses candidate entities to inspect, and finally verifies part-level evidence before answering.

\textbf{Entity-level images.}
For each candidate entity $e\in E$, we generate a full-object reference image
\[
I_e=\pi_{\mathrm{ent}}(e,P(e),\{A(p):p\in P(e)\}),
\]
where the prompt is conditioned on the entity name, its part decomposition, and concise summaries of part-level attributes. The generated image should make the entity recognizable as a whole while preserving visually relevant cues such as geometry, material, surface texture, openings, edges, handles, tips, flexible regions, or contact surfaces.

\begin{tcolorbox}[
  enhanced,
  breakable,
  width=0.98\linewidth,
  colback=tzBlueFill,
  colframe=tzBlueBorder,
  boxrule=1.2pt,
  arc=6pt,
  left=5pt,right=5pt,top=4pt,bottom=2pt,
  title={\small Image Generation Prompt (Entity)},
  coltitle=white,
  colbacktitle=tzBlueHeader2,
  fonttitle=\bfseries,
]
\small
\begin{lstlisting}[style=jsonTiny]
Generate a single-object reference image on a pure white seamless studio background.

The image must focus only on the target entity, centered and fully visible unless the annotation says a part is hidden or internal. Do not place the object in a room, scenario, or narrative scene. Do not add extra objects, hands, floor props, or decorative context. Render the object photorealistically with neutral studio lighting and sharp material detail. Preserve visible state cues such as wetness, residue, fullness, deformation, wear, or occlusion when visually representable.

Target entity: {ENTITY_NAME}
Entity summary: {ENTITY_SUMMARY}

Part-by-part guidance:
{PART_OVERVIEW}

Output goal: one clean white-background product-style reference image of this exact entity. Annotate each part's exact name beside the corresponding part. Annotate only the name, with no description.
\end{lstlisting}
\end{tcolorbox}

\textbf{Part-level images.}
For each part $p\in P(e)$, we generate a zoomed-in part image
\[
I_{e,p}=\pi_{\mathrm{part}}(e,p,A(p),I_e),
\]
using the entity image $I_e$ as a visual anchor. This ensures that the part view remains consistent with the full-object image in geometry, color, material, and relative structure. The prompt focuses tightly on the target part and asks the generator to preserve the attributes relevant to its possible use. This level is necessary because many creative solutions depend on local evidence, such as a rubber pad, serrated edge, hollow cavity, flat panel, hook-like curve, narrow tip, or absorbent surface, which may be hard to verify from the environment image alone.

\begin{tcolorbox}[
  enhanced,
  breakable,
  width=0.98\linewidth,
  colback=tzBlueFill,
  colframe=tzBlueBorder,
  boxrule=1.2pt,
  arc=6pt,
  left=5pt,right=5pt,top=4pt,bottom=2pt,
  title={\small Image Generation Prompt (Part)},
  coltitle=white,
  colbacktitle=tzBlueHeader2,
  fonttitle=\bfseries,
]
\small
\begin{lstlisting}[style=jsonTiny]
Generate a close-up studio image on a pure white seamless background.

{REFERENCE_IMAGE_INSTRUCTION}

Focus on the target part and make its local visual cues easy to inspect. Do not add room context, props, labels, arrows, or diagrams. Use product-photography lighting with strong texture visibility and clean edges. Show fine material detail, residue, moisture, wear, occlusion, seams, openings, fill cues, deformation, and surface texture whenever they are visually representable.

Entity: {ENTITY_NAME}
Entity summary: {ENTITY_SUMMARY}
Target part: {TARGET_PART_NAME}

{PHYSICAL_ATTRIBUTES_BLOCK}
{STATE_ATTRIBUTES_BLOCK}

Physical summary: {PART_PHYSICAL_SUMMARY}
State summary: {PART_STATE_SUMMARY}

Important: observe the given entity image. The generated target part image should not deviate substantially from the entity image, but should be comprehensive and informative about detailed local visual cues.

Output goal: one white-background close-up image centered on the target part. The part should have exactly the same physical and state attributes as described above.
\end{lstlisting}
\end{tcolorbox}

\textbf{Environment-level images.}
We then generate the full scenario image
\[
I_{\mathrm{env}}=\pi_{\mathrm{env}}(x,E,\{I_e:e\in E\}),
\]
conditioned on the task description, the candidate entity list, and the generated entity reference images. The prompt requires all candidate entities to appear naturally in a coherent scene with realistic scale, placement, and lighting. It also explicitly prohibits showing the task already being solved or introducing extra objects that could serve as unintended alternative solutions. Thus, the environment image defines the search space, while entity and part images provide progressively finer evidence for verification.

\begin{tcolorbox}[
  enhanced,
  breakable,
  width=0.98\linewidth,
  colback=tzBlueFill,
  colframe=tzBlueBorder,
  boxrule=1.2pt,
  arc=6pt,
  left=5pt,right=5pt,top=4pt,bottom=2pt,
  title={\small System Prompt},
  coltitle=white,
  colbacktitle=tzBlueHeader2,
  fonttitle=\bfseries,
]
\small
\begin{lstlisting}[style=jsonTiny]
Generate one photorealistic overall environment scene image.

Scenario: {SCENARIO}
Task goal: {TASK_GOAL}

The attached reference images are the required entities and are provided in the same order as the list below. Preserve each referenced entity's identity, materials, proportions, and recognizable shape while placing them naturally in the scene.

Required entities to include in the scene:
{REQUIRED_ENTITY_LIST}

Required interactable items to include naturally in the scene:
{INTERACTABLE_ITEM_LIST}

Scene requirements:
- Show a single coherent real-world scene, not a collage or white-background layout.
- Include every required entity and every required interactable item in a plausible manner.
- Naturally embed the objects into the scenario with realistic scale, lighting, and placement, and make sure they are clearly visible.
- Do not add text labels, arrows, diagrams, split panels, or product-shot framing.
- Do not show how to solve the task or explicitly depict task execution.
- Minimize extraneous objects so the scenario remains focused and uncluttered.

Output goal: one natural overall scenario image that clearly contains all required entities and interactable items.
\end{lstlisting}
\end{tcolorbox}

\textbf{Image quality and consistency checks.}
We check generated images for three requirements before using them in the benchmark. First, all candidate entities must be present and recognizable in the environment. Second, entity and part images must be visually consistent, so that a part inspection can be interpreted as a closer view of the same object. Third, the decisive part must be visually inspectable rather than hidden, cropped away, or rendered in a way that makes the task impossible. Images that fail these requirements are regenerated or manually filtered.

\textbf{Textual disambiguation for visual feedback.}
Although all generated images are used during evaluation, some fine-grained attributes may not be decisively inferable from the image alone due to rendering ambiguity, viewpoint, lighting, or material appearance. To avoid making the task depend on accidental image artifacts, we add a textual disambiguation step. Given an entity--part image pair $(I_e,I_{e,p})$ and an attribute $\alpha\in A(p)$, we use GPT-5.4 to judge whether the visual evidence alone is sufficient to support the attribute:
\[
\ell(\alpha)\in\{\textsc{VisualEnough},\textsc{TextNeeded}\}.
\]
\textsc{VisualEnough} means that the attribute can be reasonably inferred from the image without additional text. \textsc{TextNeeded} means that the attribute is part of the knowledge-base annotation and is compatible with the generated image, but the visual evidence may be ambiguous; in this case, we provide a short textual clarification together with the image when that entity or part is returned as feedback.

This step is used only to disambiguate low-level visual details. The accompanying text is restricted to object or part attributes, such as material, state, surface property, rigidity, hollowness, or accessibility. It does not reveal the target affordance, the correct entity, the correct part, or the final solution. The same procedure is applied to all candidate entities and parts, not only to the gold solution. Thus, the benchmark still requires models to inspect the visual evidence and reason over candidate parts, while preventing failures caused by attributes that are intended in the generated image but not visually decisive. We denote the attributes requiring textual clarification as
\[
\delta(p)=\{\alpha\in A(p):\ell(\alpha)=\textsc{TextNeeded}\},
\]
and include concise descriptions of $\delta(p)$ when presenting the corresponding image as interaction feedback.


\section{Training Trajectory Construction Details}
\label{apdx:trajectory_construction}

We provide additional details for the trajectory construction procedure introduced in the main text. The goal is to construct supervision not only for the final entity--part answer, but also for the intermediate evidence-seeking process: selecting entities to inspect, verifying candidate parts, judging physical attributes, and rejecting plausible but physically invalid alternatives.

\subsection{Trajectory Formulation}
\label{app:trajectory-formulation}

Each task is represented as $\mathcal{T}=(x,I_{\mathrm{env}},E,g)$, where $x$ is the task instruction, $I_{\mathrm{env}}$ is the environment image, $E$ is the set of scene entities, and $g=(e^*,p^*,f^*)$ is the gold solution consisting of the target entity, target part, and target affordance. Each entity $e\in E$ has an annotated part set $P(e)$.

A trajectory is a sequence of interaction turns:
\[
\tau=\{(o_t,r_t)\}_{t=1}^{T}, \qquad 
o_t=(u_t,I_t), \qquad 
r_t=(z_t,a_t).
\]
Here, $u_t$ is the textual feedback, $I_t$ is the visual observation, $z_t$ is the model's free-form reasoning, and $a_t$ is a structured action. The action space contains three operations:
\[
a_t\in
\{\texttt{inspect\_entity}(e),\;
\texttt{inspect\_part}(e,p),\;
\texttt{answer}(e,p,h)\},
\]
where $e\in E$, $p\in P(e)$, and $h$ describes how the selected part should be used.

The actions define the interaction protocol. The action $\texttt{inspect\_entity}(e)$ returns the entity-level image $I_e$ and the part list $P(e)$. The action $\texttt{inspect\_part}(e,p)$ returns the zoomed-in part image $I_{e,p}$, optionally with short attribute-level textual disambiguation. The action $\texttt{answer}(e,p,h)$ terminates the trajectory and provides the final grounded solution. Thus, the observation at each turn is determined by the previous action:
\[
I_t=
\begin{cases}
I_{\mathrm{env}}, & t=1,\\
I_e, & a_{t-1}=\texttt{inspect\_entity}(e),\\
I_{e,p}, & a_{t-1}=\texttt{inspect\_part}(e,p).
\end{cases}
\]
This formulation ensures that each reasoning step is aligned with the appropriate level of visual evidence: scene, entity, or part.

\subsection{Knowledge-Guided Exploration Stack}
\label{app:exploration-stack}

To construct positive trajectories, we maintain an ordered exploration stack $\mathcal{S}_t$. Each stack element is either an entity item $(\texttt{entity},e)$ or a part item $(\texttt{part},(e,p))$. The top element determines the next inspection target.

The stack is guided by an affordance-relevance function:
\[
J:E\times P(e)\rightarrow\{0,1\},
\]
where $J(e,p)=1$ indicates that part $p$ of entity $e$ has an affordance similar or relevant to the target affordance $f^*$ according to the knowledge base $\mathcal{K}$. This does not necessarily mean that $(e,p)$ is the gold answer; it only means that the part is worth inspecting. This distinction is important because many distractors are affordance-similar but fail under fine-grained physical verification.

At the first turn, the model observes the scene and proposes an ordered list of candidate entities. This list initializes $\mathcal{S}_1$, prioritizing likely entities while allowing systematic exploration. The stack is then updated as follows.

\begin{itemize}[topsep=-2pt,leftmargin=12pt,itemsep=-1pt]
    \item \textbf{Entity inspection.} When the top element is $(\texttt{entity},e_i)$, the positive branch takes $\texttt{inspect\_entity}(e_i)$. The entity is removed from the stack, and all affordance-relevant parts $\{p\in P(e_i):J(e_i,p)=1\}$ are pushed onto the stack for part-level verification. If no such part exists, exploration moves to the next entity.

    \item \textbf{Part inspection.} When the top element is $(\texttt{part},(e_i,p_{i,j}))$, the positive branch takes $\texttt{inspect\_part}(e_i,p_{i,j})$. The part is then removed from the stack and assigned a binary judgment $b_t\in\{0,1\}$, indicating whether its observed attributes satisfy the task requirements.

    \item \textbf{Termination.} Exploration terminates when $\mathcal{S}_t=\emptyset$. The model then compares inspected candidate parts, especially those with $b_t=1$, and produces the final action $\texttt{answer}(e^*,p^*,h^*)$.
\end{itemize}

This mechanism yields a coarse-to-fine positive trajectory. The model first searches over entities, then verifies affordance-relevant parts, and finally selects the gold pair based on fine-grained physical evidence rather than object-level plausibility alone.

\subsection{Three-Branch Trajectory Sampling}
\label{app:three-branch-sampling}

The stack specifies what the positive trajectory should inspect, but we still need to generate the reasoning text associated with each step. To obtain both supervised and preference-learning data, we sample three aligned branches at each shared interaction context $c_t$:
\[
r_t^{b}=(z_t^{b},a_t^{b}), \qquad b\in\{+,-,--\}.
\]
The positive branch is the preferred grounded response, while the negative and hard-negative branches are rejected alternatives.

We use GPT-5.4 as the teacher model to help generate the branch-specific reasoning and responses. For the positive branch, GPT-5.4 does not freely decide the exploration structure. Instead, the inspected target, part-level judgment, and final answer are determined by the knowledge base $\mathcal{K}$, the exploration stack $\mathcal{S}_t$, and the gold solution $g$. GPT-5.4 is used to express this predetermined structure in natural, coherent, and visually grounded language. For the negative and hard-negative branches, GPT-5.4 is prompted with different guidance signals to produce rejected responses at the same state.

Formally, each branch is generated with a branch-specific system prompt $s^b$ and guidance function $G^b$:
\[
r_t^b
=
\pi_{\mathrm{GPT\text{-}5.4}}
\bigl(s^b,c_t,G^b(t,\mathcal{K},g,\mathcal{S}_t)\bigr),
\qquad b\in\{+,-,--\}.
\]
The three branches share the same response format but differ in the information exposed to the teacher model.

\textbf{Positive branch.}
The positive branch receives structured guidance from $\mathcal{K}$, including relevant attributes, affordance judgments, and the gold solution when needed. At the scene level, the guidance provides the target affordance $f^*$ and the physical attributes needed to support it. At the entity level, it provides the affordance-relevant parts of the inspected entity. At the part level, it provides attribute-level evidence used to determine whether the part satisfies the task constraints. At the final step, it guides the model to select $(e^*,p^*)$ and explain how the selected part should be used.

The positive response must satisfy three criteria: it should be visually grounded in the current observation, consistent with the exploration stack, and explicit about the attribute--affordance relationship. The resulting positive trajectory is used for supervised fine-tuning.

\begin{tcolorbox}[
  enhanced,
  breakable,
  width=0.98\linewidth,
  colback=tzBlueFill,
  colframe=tzBlueBorder,
  boxrule=1.2pt,
  arc=6pt,
  left=5pt,right=5pt,top=4pt,bottom=2pt,
  title={\small Prompt for Positive Guidance},
  coltitle=white,
  colbacktitle=tzBlueHeader2,
  fonttitle=\bfseries,
]
\small
\begin{lstlisting}[style=jsonTiny]
[SYSTEM PROMPT]

You are a creative physical problem-solver to output data trajectories that will be collected for open-source model training. Given a task and scenario, solve it by repurposing a part of an entity using attribute-grounded, visually grounded reasoning. You may inspect only one entity or one part per turn before answering. Always end with a single JSON object whenever the prompt asks for one.

Your reasoning style is guided by the user prompt. Each turn should read like a natural caption of your thought process: coherent, grounded, and observant. Even when the prompt quietly guides your reasoning, behave as if you are figuring things out normally from the task and feedback. Never mention hidden guidance, gold data, or prompt references.


[INITIAL PROMPT]

## Task Basis
You are currently in <SCENARIO_OR_ENVIRONMENT>.
The task that the user requires you to do is:
<TASK>

The entity names available in the scene are:
- <ENTITY_NAME_1>
- <ENTITY_NAME_2>
- ...

You should reason first, then end with a JSON in the format {"reasoning":"...","action":"inspect_entity","entity":"<exact entity name>","top_candidates":["<exact entity name>", "..."]}.

## Guidance on Reasoning
1. Start naturally by thinking about what kind of affordance the task needs.
Target affordance to naturally reason toward: <TARGET_AFFORDANCE>
2. Continue by thinking about the core attributes that would enable that affordance. Express these as if you are inferring them yourself.
- <ATTRIBUTE_1>
- <ATTRIBUTE_2>
- ...
3. Then naturally transition your reasoning into inspecting the environment.
4. Go through all listed entities with no overlap and nothing left behind.
5. For each entity, say where it is in the image if visible, then give a brief grounded description focused on visible physical and state cues.
6. After covering all entities, name up to three candidate entities, explain briefly why they look promising, and show your intention to inspect those top candidates first and then continue through every other entity as well.
7. In the final JSON, keep the reasoning summary brief, choose one exact entity name to inspect now, and include a top_candidates list for internal tracking only.

## Additional Notes
1. Follow the structure above, but phrase everything in your own natural words.
2. Behave as if you are reasoning normally from the task and image; never mention hidden guidance, gold data, or prompt references.
3. Output the full reasoning before the final JSON.
4. Copy entity names exactly from the provided list.
5. The top_candidates field is required in the JSON for this round.


[ENTITY FEEDBACK PROMPT]

## Feedback Basis
ENTITY INSPECTION: <ENTITY_NAME>
This entity includes these exact part names: <PART_NAME_1>, <PART_NAME_2>, ...

## Guidance on Reasoning
Please perform visual grounding for each part first and explain in your own words why it may or may not achieve the needed affordance.
<ENTITY_BRANCH_INSTRUCTIONS>

Here is the reference for each part:
- <PART_NAME_1>: <CAN_OR_CANNOT_SERVE_VERDICT>. Reason: <REFERENCE_REASON_1>
- <PART_NAME_2>: <CAN_OR_CANNOT_SERVE_VERDICT>. Reason: <REFERENCE_REASON_2>
- ...

## Additional Notes
1. Observe and reason about all parts with no overlap and nothing left behind.
2. Base your reasoning on visible cues and the provided reference, but do not copy the reference text verbatim.
3. Keep the reasoning natural and coherent, as if you are figuring it out normally.
4. Output the full reasoning before any JSON.
5. Copy part or entity names exactly.

(If this is the last exploration step, use:)
This is the last exploration step.
Reason through every part, explain why none of them finally solves the need well enough, and end by explicitly saying this is the last entity you need to explore and you should now move on to the final answer.
Do not output any JSON in this turn.
Do not append a JSON object at the end.
Stop immediately after the prose reasoning is finished.

(If this entity has similar-affordance parts, use:)
This entity has several parts that may elicit the needed affordance: <SIMILAR_PART_NAMES>.
Reason through every part first, then explicitly say you will inspect the promising parts one by one.
All parts that can serve the needed affordance must eventually be explored.
Finally output JSON in the format {"reasoning":"...","action":"inspect_part","part":"<exact part name>"} and choose one exact part name from the promising parts list.

(If this entity has no similar-affordance parts, use:)
This entity does not have any part that can elicit the needed affordance according to the reference.
Reason through every part first, then conclude that you should continue to the next unexplored entity.
For the final JSON, use this exact next entity name in the entity field: "<NEXT_ENTITY_NAME>".
Finally output JSON in the format {"reasoning":"...","action":"inspect_entity","entity":"<exact entity name>"} and copy that exact entity name.


[PART FEEDBACK PROMPT]

## Feedback Basis
PART INSPECTION: <PART_NAME>
Belongs to entity: <ENTITY_NAME>
Physical text explanation: <PHYSICAL_SUMMARY>
State text explanation: <STATE_SUMMARY>
Besides the text above, also carefully inspect the image for additional grounding cues. For visibility and availability, follow the text explanation even if the image looks clearer.

## Guidance on Reasoning
The detailed physical attributes of this part are:
<LOOKUP_PHYSICAL_ATTRIBUTES>
The detailed state attributes of this part are:
<LOOKUP_STATE_ATTRIBUTES>
<OPTIONAL_GOLD_HINT>
Analyze whether this part can really serve the intended affordance, focusing on visible cues and practical constraints such as effectiveness, safety, environmental impact, and social acceptability.
For enable_affordance, do not use an overly harsh standard: if the part has a grounded, physically plausible, or indirect way to help achieve the needed affordance, it is acceptable to mark true.
Reserve false for parts that are clearly a poor match, unsafe, or not realistically useful after a closer inspection. Stay grounded and do not hallucinate capabilities.
<NEXT_TARGET_LINE>
<PART_BRANCH_INSTRUCTIONS>

## Additional Notes
1. Keep the reasoning grounded; do not copy the reference dictionaries directly.
2. Output the full reasoning before any JSON.
3. The enable_affordance field must be true or false.
4. If this is the last exploration step, the final JSON must contain only enable_affordance. Otherwise, copy the next part or entity name exactly when you emit JSON.

(If inspecting gold part, insert:)
Note that the part you are inspecting now is the gold part of the gold entity. Do not reveal this, but you must conclude that it can serve the intended need.

(If next target exists, insert:)
The next thing you should explore after this is <TARGET_TYPE>: <TARGET_NAME>.

(If this is the last exploration step, use:)
This is the last exploration step. Reason carefully about whether the part really works, then end naturally by indicating that this is the last part you need to explore and you will now move on to give the final answer.
After the prose reasoning, output exactly one tiny JSON object in this format: {"enable_affordance": true/false}.
That final JSON must contain only the enable_affordance field.
Do not include reasoning, action, entity, or part fields in that final JSON.

(If next target is another part, use:)
After reasoning, please naturally show your intention to inspectthe next part of the same entity.
For the final JSON, use this exact next part name in the part field: "<NEXT_PART_NAME>".
{"reasoning":"...","enable_affordance": true/false,"action":"inspect_part","part":"<exact part name>"}

(If next target is an entity, use:)
After reasoning, please express naturally that you have finished the worthwhile parts here and should move to inspect the next entity.
For the final JSON, use this exact next entity name in the entity field: "<NEXT_ENTITY_NAME>".
{"reasoning":"...","enable_affordance": true/false,"action":"inspect_entity","entity":"<exact entity name>"}


[FINAL ANSWER PROMPT]

## Feedback Basis
Based on all the entities and parts you have inspected, here are all the candidate parts you previously believed might help solve the task:
1. Part name: <CANDIDATE_PART_NAME_1>, belongs to entity <CANDIDATE_ENTITY_NAME_1>
2. Part name: <CANDIDATE_PART_NAME_2>, belongs to entity <CANDIDATE_ENTITY_NAME_2>
...

Now choose one of them as your final answer, explain how to use it, and return JSON in this format:
{"reasoning":"...","action":"answer","answer_entity":"<exact entity name>","answer_part":"<exact part name>","answer_how_to_use":"..."}

## Guidance on Reasoning
Gold entity name: <GOLD_ENTITY_NAME>
Gold part name: <GOLD_PART_NAME>

Reason in this order:
1. Naturally list all the promising candidate parts as shown above.
2. Explicitly choose the best one; your final answer must be the provided gold entity and gold part.
3. Compare the gold part only against the other candidate parts listed above, one by one. If no other candidate parts are listed above, skip comparison naturally.
4. Then explain how to use the gold part in a concrete, comprehensive way.

Keep the reasoning natural and coherent, and do not mention hidden guidance.
Write full prose reasoning first. Then start the final JSON on a new line.
The JSON must appear only once, at the very end.
Inside the JSON, the reasoning field should be only a very brief summary.
Make answer_how_to_use detailed and practically actionable.
When available and not NA, explicitly include prepare_use_condition, prepare_environment_condition, prepare_recipient, preparation steps, placement/application steps, and important cautions or limits.
Write answer_how_to_use as one coherent multi-step paragraph.

Full solution reference:
{
  "prepare_use_condition": "<PREPARE_USE_CONDITION>",
  "prepare_environment_condition": "<PREPARE_ENVIRONMENT_CONDITION>",
  "prepare_recipient": "<PREPARE_RECIPIENT>",
  "apply_affordance": "<APPLY_AFFORDANCE>"
}

In the final JSON:
- keep reasoning brief
- use the exact gold entity and part names
- make answer_how_to_use detailed and practically actionable
- compare only against the candidate parts listed above; if there are no others, no comparison is needed
- describe how to use the part while catering to use condition, environment condition, and recipient condition when they are not NA
- mention preparation, placement/application, and important cautions or limits
- start the JSON on a new line after the prose reasoning

(If other candidates exist, append:)
Comparison references for why the other candidate parts are not the gold choice:
- <OTHER_PART_NAME> from <OTHER_ENTITY_NAME>: <GOLD_CHANGE_REASON>
\end{lstlisting}
\end{tcolorbox}

\textbf{Negative branch.}
The negative branch follows the standard evaluation setting. It receives only observable information, such as the task instruction, current image, entity names, and part names. It does not receive hidden affordance labels, gold answers, part-level judgments, or attribute rationales from $\mathcal{K}$. This branch captures realistic inference-time mistakes, such as inspecting irrelevant entities, overlooking decisive parts, or selecting a plausible but suboptimal part.

Unlike the hard-negative branch, the negative branch is not explicitly instructed to be wrong. Its errors arise from the lack of fine-grained affordance guidance. When the positive exploration stack is exhausted, a termination signal is added so that the branch produces a final answer and remains comparable with the positive trajectory.

\textbf{Hard-negative branch.}
The hard-negative branch is designed to create stronger contrast for preference learning. It preserves fluent reasoning and valid action format, but is guided toward semantically incorrect or insufficiently grounded conclusions. For example, it may hallucinate unsupported physical attributes, rely on object-level priors, ignore visual evidence, or choose an affordance-similar distractor that lacks the required physical properties.

The hard-negative branch receives structural information such as the task, entity names, part names, and output format, but no grounding signals from $\mathcal{K}$. Its action $a_t^{--}$ is not constrained by the affordance-relevance function $J$, allowing it to deviate from the positive exploration policy while remaining superficially plausible.

\begin{tcolorbox}[
  enhanced,
  breakable,
  width=0.98\linewidth,
  colback=tzBlueFill,
  colframe=tzBlueBorder,
  boxrule=1.2pt,
  arc=6pt,
  left=5pt,right=5pt,top=4pt,bottom=2pt,
  title={\small Prompt for Negative Guidance},
  coltitle=white,
  colbacktitle=tzBlueHeader2,
  fonttitle=\bfseries,
]
\small
\begin{lstlisting}[style=jsonTiny]
[SYSTEM PROMPT]

You are a ungrounded and problematic physical problem-solver to output intentionally hard-negative training trajectories. Keep the same natural structure and tone as a normal solver, but deliberately make the reasoning wrong, ineffective, and visually/physically ungrounded: rely on unsupported assumptions, misread and ignore real evidence, and always steer away from the best-supported conclusion. You may inspect only one entity or one part per turn before answering. Keep every entity name and part name exact as provided, and never mention that the reasoning is intentionally wrong. Always end with a single JSON object whenever the prompt asks for one.


[INITIAL PROMPT]

## Task Basis
You are currently in <SCENARIO_OR_ENVIRONMENT>.
The task that the user requires you to do is:
<TASK>

The entity names available in the scene are:
- <ENTITY_NAME_1>
- <ENTITY_NAME_2>
- ...

You should reason first, then end with a JSON in the format {"reasoning":"...","action":"inspect_entity","entity":"<exact entity name>","top_candidates":["<exact entity name>", "..."]}.

## Guidance on Reasoning
1. Start naturally by thinking about what kind of affordance the task seems to need.
Target affordance to loosely reason around: <TARGET_AFFORDANCE>
2. Then please begin to inspect the environment and entities in the image, but intentionally let your analysis drift away from the actual image: overstate weak cues, assume hidden properties, and prefer surface-level plausibility over real physical grounding.
3. When describing entities, act as if you are reading the scene normally, but deliberately downplay the entities that look most physically suitable and highlight weaker alternatives instead.
4. Go through only some of the entities and don't need to cover all of them, while keeping the descriptions naturally phrased.
5. For each entity, hallucinate where it is in the image, then give a brief description that sounds plausible but is allowed to be visually or physically unsupported.
6. After covering all entities, name up to three candidate entities, explain briefly why they seem promising, and prefer candidates whose usefulness sounds defensible even if the image does not really support it.
7. If one entity appears obviously strongest, avoid prioritizing it; instead, always favor those weaker or only plausible-sounding ones as your inspection candidates.
8. In the final JSON, keep the reasoning summary brief, choose one exact entity name to inspect now, and include a top_candidates list for internal tracking only.

## Additional Notes
1. Follow the structure above, but phrase everything in your own natural words.
2. Keep the response smooth and coherent; never say that you are being intentionally ungrounded or wrong.
3. Output the full reasoning before the final JSON.
4. Copy entity names exactly from the provided list.
5. The top_candidates field is required in the JSON for this round.


[ENTITY FEEDBACK PROMPT]

## Feedback Basis
ENTITY INSPECTION: <ENTITY_NAME>
This entity includes these exact part names: <PART_NAME_1>, <PART_NAME_2>, ...

## Guidance on Reasoning
Reason through some of the listed part in a natural style and do not need to reason every of the, deliberately keep the analysis visually and physically ungrounded and hallucinated.
You may speculate about hidden rigidity, leverage, reachability, or usefulness even when the image or reference does not really support it.
Use correct and exact spellings; only the reasoning and conclusions should be wrong or very very weakly grounded.
<ENTITY_BRANCH_INSTRUCTIONS>

Here is the reference for each part:
- <PART_NAME_1>: <CAN_OR_CANNOT_SERVE_VERDICT>. Reason: <REFERENCE_REASON_1>
- <PART_NAME_2>: <CAN_OR_CANNOT_SERVE_VERDICT>. Reason: <REFERENCE_REASON_2>
- ...

## Additional Notes
1. Observe and discuss some of the parts, deliberately omit some and do not discuss all of them.
2. Use the reference only as material to push against; do not copy it verbatim and do not let it force you into a grounded conclusion.
3. Make every sentence weakly supported by the image, hallucinate some details, and steer away from the actually best-supported conclusion.
4. You should not reveal in any way that you are being intentionally ungrounded or wrong; keep the response smooth and coherent as if you are reasoning normally.
5. Output the full reasoning before any JSON.
6. Copy part or entity names exactly.

(If this is the last exploration step, use:)
This is the last exploration step.
Reason through some of the parts instead of all of them, and keep the analysis intentionally weakly grounded: misread cues, infer unsupported properties, and naturally drift away from the actually suitable affordance.
End by explicitly saying this is the last entity you need to explore and you should now move on to the final answer.
Do not output any JSON in this turn.
Do not append a JSON object at the end.
Stop immediately after the prose reasoning is finished.

(If non-similar parts exist, use:)
Some parts may actually have similar affordance support, but you should avoid inspecting those next. Instead, naturally hallucinate or speculate your way toward parts that are not listed as similar.
If you mention the similar-affordance parts (<SIMILAR_PART_NAMES>) at all, downplay them and treat them as less worth checking than the weaker alternatives.
Reason through every part first, then explicitly say you will inspect the promising parts one by one.
Finally output JSON in the format {"reasoning":"...","action":"inspect_part","part":"<exact part name>"} and choose one exact part name from this non-similar list: <NON_SIMILAR_PART_NAMES>.

(If no part is marked as similar, use this line instead of the similar-parts line above:)
No part is marked as similar by the reference, so choose a weakly justified part and make it sound worth checking.

(If all presented parts are similar/reference-aligned, use:)
Even if this entity contains parts that look reference-aligned, do not follow that path here.
Downplay the stronger parts, keep the reasoning visually ungrounded and gallucinated, and conclude that you should continue to the next unexplored entity instead.
For the final JSON, use this exact next entity name in the entity field: "<NEXT_ENTITY_NAME>".
Finally output JSON in the format {"reasoning":"...","action":"inspect_entity","entity":"<exact entity name>"} and copy that exact entity name.


[PART FEEDBACK PROMPT]

## Feedback Basis
PART INSPECTION: <PART_NAME>
Belongs to entity: <ENTITY_NAME>
Physical text explanation: N/A, please hallucinate and speculate freely, and make your reasoning visually and physically ungrounded.
State text explanation: N/A, please hallucinate and speculate freely, and make your reasoning visually and physically ungrounded.
Besides the text above, you should always speculate beyond the actual image and text. Keep the structure of a grounded inspection, but let the content drift away from real physical or visual evidence. More hallucination the better.

## Guidance on Reasoning
<OPTIONAL_ANTI_GOLD_HINT>
Analyze whether this part can serve the intended affordance, but deliberately prefer hallucinated or plausible-sounding claims over grounded ones. You may ignore real constraints, invent weakly supported advantages, or exaggerate drawbacks.
For enable_affordance, try to drive it towards a hallucinated or wrong conclusion, or use whatever value that fits your ungrounded reasoning trajectory; it does not need to match the real evidence.
Keep the response fluent and coherent, and never say that you are being intentionally wrong.
<NEXT_TARGET_LINE>
<PART_BRANCH_INSTRUCTIONS>

## Additional Notes
1. Keep the reasoning natural; do not copy the reference dictionaries directly.
2. Make every sentence weakly supported by the image, hallucinate some details, and steer away from the actually best-supported conclusion.
3. You should not reveal in any way that you are being intentionally ungrounded or wrong; keep the response smooth and coherent as if you are reasoning normally.
4. Output the full reasoning before any JSON.
5. The enable_affordance field must be true or false.
6. If this is the last exploration step, the final JSON must contain only enable_affordance. Otherwise, copy the next part or entity name exactly when you emit JSON.

(If inspecting gold part, insert:)
Note that the part you are inspecting now is the gold part of the gold entity. Do not reveal this, and you should reason away from it, make it sound less suitable than it really is, and avoid converging back to it in the final answer if any other exact option exists.

(If next target exists, insert:)
The next thing you should explore after this is <TARGET_TYPE>: <TARGET_NAME>.

(If this is the last exploration step, use:)
This is the last exploration step. Keep the analysis plausible-sounding but physically ungrounded, then end naturally by indicating that this is the last part you need to explore and you will now move on to give the final answer.
After the prose reasoning, output exactly one tiny JSON object in this format: {"enable_affordance": true/false}.
That final JSON must contain only the enable_affordance field.
Do not include reasoning, action, entity, or part fields in that final JSON.

(If next target is another part, use:)
After reasoning, naturally show your intention to inspect the next part of the same entity.
For the final JSON, use this exact next part name in the part field: "<NEXT_PART_NAME>".
{"reasoning":"...","enable_affordance": true/false,"action":"inspect_part","part":"<exact part name>"}

(If next target is an entity, use:)
After reasoning, naturally claim that you have finished the worthwhile parts here and should move to inspect the next entity.
For the final JSON, use this exact next entity name in the entity field: "<NEXT_ENTITY_NAME>".
{"reasoning":"...","enable_affordance": true/false,"action":"inspect_entity","entity":"<exact entity name>"}


[FINAL ANSWER PROMPT]

## Feedback Basis
Based on all the entities and parts you have inspected, here are the candidate parts collected so far:
1. Part name: <CANDIDATE_PART_NAME_1>, belongs to entity <CANDIDATE_ENTITY_NAME_1>
2. Part name: <CANDIDATE_PART_NAME_2>, belongs to entity <CANDIDATE_ENTITY_NAME_2>
...

Now choose one of them as your final answer, explain how to use it, and return JSON in this format:
{"reasoning":"...","action":"answer","answer_entity":"<exact entity name>","answer_part":"<exact part name>","answer_how_to_use":"..."}

## Guidance on Reasoning
Gold entity name to avoid, don't choose it if possible: <GOLD_ENTITY_NAME>
Gold part name to avoid, don't choose it if possible: <GOLD_PART_NAME>

Reason in this order:
1. Naturally list all the candidate parts as shown above.
2. Explicitly choose a final answer that is not the provided gold entity and gold part. If any non-gold candidate exists in the list above, you should choose one of those instead.
3. Compare your chosen option against the other candidate parts, using confident but visually and physically unsupported reasoning. Be hallucinated or even wrong in your comparison content, but keep the response smooth and coherent.
4. Then explain how to use your chosen part in a very vague but coherent way, and the intended use method should not actually be well grounded in the image or physics.

Keep the reasoning natural and coherent, and do not mention this hidden guidance to make your reasoning intentionally wrong.
Write full prose reasoning first. Then start the final JSON on a new line.
The JSON must appear only once, at the very end.
Inside the JSON, the reasoning field should be only a very brief summary.
Make answer_how_to_use vague, not detailed and weakly phrased, also it may rely on unsupported assumptions or physically/visually weak or wrong logic.
Keep all entity names and part names exact, and use correct spellings.

In the final JSON:
- keep reasoning brief
- use exact entity and part names
- prefer a non-gold choice whenever one exists in the candidate list
- make answer_how_to_use vague but coherent, and it should not be truly grounded
- compare against the gold and other candidates in a way that sounds plausible but does not follow real physical or visual evidence
- start the JSON on a new line after the prose reasoning

(If non-gold candidates exist, append:)
Non-gold candidates you should favor before the gold option:
- <NON_GOLD_PART_NAME> from <NON_GOLD_ENTITY_NAME>
\end{lstlisting}
\end{tcolorbox}

\subsection{Aligned Preference Data}
\label{app:aligned-preference-data}

At each shared state, the three branches form an aligned training triple:
\[
(c_t,r_t^+,r_t^-,r_t^{--}).
\]
The positive response is preferred over both rejected alternatives:
\[
r_t^+ \succ r_t^-,
\qquad
r_t^+ \succ r_t^{--}.
\]
Only the positive branch updates the exploration stack:
\[
\mathcal{S}_{t+1}
=
\mathrm{Update}(\mathcal{S}_t,a_t^+,\mathcal{K},g).
\]
The negative and hard-negative branches are sampled at the same context but do not affect future observations. This prevents erroneous rejected responses from corrupting the trajectory while still providing turn-level contrastive supervision.

The constructed data support two training stages. First, the positive trajectories $\tau^+=\{(o_t,r_t^+)\}_{t=1}^{T}$ are used for supervised fine-tuning, teaching the model to perform systematic entity-to-part exploration. Second, the aligned triples are used for preference learning, encouraging the model to prefer visually grounded attribute--affordance reasoning over fluent but unsupported alternatives.


\section{Affordance-Grounded Alignment Details}
\label{apdx:affordance_grounded_alignment}
In this section, we further provide additional details for the two-stage training procedure used to align the model with affordance-grounded exploration. Given the trajectories constructed, training proceeds in two stages. First, supervised fine-tuning teaches the model to imitate the positive trajectories and acquire the desired coarse-to-fine exploration behavior. Second, turn-level preference learning teaches the model to prefer grounded attribute--affordance reasoning over fluent but unsupported alternatives.

\subsection{Supervised Fine-Tuning}
\label{app:sft}

We first train the model on the positive trajectories constructed from the knowledge-guided exploration stack. Let
\[
\mathcal{D}_{\mathrm{SFT}}
=
\{(\mathcal{T}^{(n)},\tau^{+(n)})\}_{n=1}^{|\mathcal{D}|}
\]
denote the supervised fine-tuning dataset, where $\mathcal{T}^{(n)}=(x^{(n)},I_{\mathrm{env}}^{(n)},E^{(n)},g^{(n)})$ and $\tau^{+(n)}=\{(o_t^{(n)},r_t^{+(n)})\}_{t=1}^{T^{(n)}}$ is the positive trajectory for task $n$. Each positive response is written as $r_t^{+(n)}=(z_t^{+(n)},a_t^{+(n)})$, where $z_t^{+(n)}$ is the grounded reasoning and $a_t^{+(n)}$ is the structured action.

At turn $t$, the model conditions on the task, the available visual observation, the current feedback, and the previous positive interaction history:
\[
c_t^{(n)}
=
\left(
x^{(n)},
I_t^{(n)},
u_t^{(n)},
\{(o_k^{(n)},r_k^{+(n)})\}_{k=1}^{t-1}
\right).
\]
The SFT objective maximizes the likelihood of the positive response at each turn:
\[
\mathcal{L}_{\mathrm{SFT}}(\theta)
=
-
\sum_{n=1}^{|\mathcal{D}|}
\sum_{t=1}^{T^{(n)}}
\log
\pi_{\theta}
\left(
r_t^{+(n)}
\mid
c_t^{(n)}
\right).
\]

This objective trains the model to imitate complete interaction trajectories rather than only final answers. As a result, the model learns to propose candidate entities, inspect affordance-relevant parts, ground intermediate decisions in observed physical attributes, and produce the final entity--part answer through comparison among inspected candidates.

However, SFT alone has an important limitation. The positive trajectories are constructed with structured guidance from the affordance knowledge base $\mathcal{K}$, whereas inference must proceed without hidden affordance labels, gold solutions, or attribute rationales. Therefore, SFT teaches the model what grounded exploration should look like, but does not directly penalize plausible yet incorrect reasoning. To reduce this gap, we further apply turn-level preference learning.

\subsection{Turn-Level Direct Preference Optimization}
\label{app:dpo}

We use Direct Preference Optimization (DPO) to encourage the model to prefer visually grounded attribute--affordance reasoning over rejected alternatives. The preference data come from the aligned triples constructed during three-branch trajectory sampling:
\[
(c_t,r_t^+,r_t^-,r_t^{--}).
\]
Here, $r_t^+$ is the positive grounded response, $r_t^-$ is the negative response generated under standard observable feedback, and $r_t^{--}$ is the hard-negative response that preserves valid format but is guided toward ungrounded or misleading reasoning.

For DPO, we construct turn-level preference pairs
\[
(c_t^{\mathrm{DPO}}, r_t^+, r_t^{\mathrm{rej}}),
\qquad
r_t^{\mathrm{rej}}\in\{r_t^-,r_t^{--}\}.
\]
The context $c_t^{\mathrm{DPO}}$ is the observable version of the shared interaction context. It contains the task instruction, the current observation, the current feedback, and the previous interaction history, but removes hidden guidance from $\mathcal{K}$ such as affordance labels, gold answers, and attribute rationales. This makes the preference context closer to the standard evaluation setting:
\[
c_t^{\mathrm{DPO}}
=
\mathrm{Obs}(c_t),
\]
where $\mathrm{Obs}(\cdot)$ denotes the projection that keeps only inference-time observable information.

The DPO loss is
\[
\mathcal{L}_{\mathrm{DPO}}(\theta)
=
-
\mathbb{E}_{(c_t^{\mathrm{DPO}},r_t^+,r_t^{\mathrm{rej}})\sim\mathcal{D}_{\mathrm{DPO}}}
\left[
\log\sigma
\left(
\beta
\log
\frac{
\pi_{\theta}(r_t^+\mid c_t^{\mathrm{DPO}})
}{
\pi_{\mathrm{ref}}(r_t^+\mid c_t^{\mathrm{DPO}})
}
-
\beta
\log
\frac{
\pi_{\theta}(r_t^{\mathrm{rej}}\mid c_t^{\mathrm{DPO}})
}{
\pi_{\mathrm{ref}}(r_t^{\mathrm{rej}}\mid c_t^{\mathrm{DPO}})
}
\right)
\right],
\]
where $\pi_{\mathrm{ref}}$ is the reference model, $\beta$ controls the strength of the preference margin, and $\sigma(\cdot)$ is the sigmoid function.

This turn-level formulation provides dense supervision for all major decision points in the interaction, including entity inspection, part inspection, and final answer generation. It is especially useful because many rejected responses are not trivially wrong. The negative branch may contain realistic inference-time errors, while the hard-negative branch may preserve fluent reasoning, correct entity and part names, and valid action syntax, but still hallucinate physical attributes or select an affordance-similar distractor. By contrasting these rejected responses with the grounded positive response under the same observable context, DPO teaches the model to discriminate between genuine attribute evidence and unsupported plausibility.

\subsection{Overall Training Procedure}
\label{app:overall-training}

The two stages play complementary roles. SFT provides the model with a grounded exploration policy by imitating positive trajectories:
\[
\tau^+=\{(o_t,r_t^+)\}_{t=1}^{T}.
\]
DPO then sharpens the model's decision boundary using aligned turn-level comparisons:
\[
r_t^+ \succ r_t^-,
\qquad
r_t^+ \succ r_t^{--}.
\]
Together, these objectives train the model to perform systematic coarse-to-fine exploration while avoiding the main failure mode of the benchmark: producing fluent but physically unsupported attribute--affordance reasoning.


\section{Experiment Details}
\label{apdx:experiment_details}

\textbf{Evaluation Protocol and Prompt.}
In this interactive evaluation setting, the model first receives the task, scenario, environment image, and the names of all available entities, but it does not initially know the parts of each entity. At each turn, it must either inspect one entity to reveal its available part names, inspect one specific part to receive its image and physical/state descriptions, or provide a final answer. The prompt enforces grounded exploration: the model is asked to compare multiple candidate entities and parts, rely on visible attributes and part-level feedback, and return a single JSON object specifying the selected entity, selected part, and how that part can be physically repurposed to solve the task. In the following, we present the major evaluation prompt we employ for the interactive evaluation protocol.

\begin{tcolorbox}[
  enhanced,
  breakable,
  width=0.98\linewidth,
  colback=tzBlueFill,
  colframe=tzBlueBorder,
  boxrule=1.2pt,
  arc=6pt,
  left=5pt,right=5pt,top=4pt,bottom=2pt,
  title={\small System Prompt},
  coltitle=white,
  colbacktitle=tzBlueHeader2,
  fonttitle=\bfseries,
]
\small
\begin{lstlisting}[style=jsonTiny]
You are a creative physical problem-solver. Given a task and scenario, solve it by repurposing a part of an entity using attribute-grounded, visually grounded reasoning. You may inspect only one entity or one part per turn before answering. Base conclusions on observable physical and state properties, avoid unsupported assumptions, and devise a physically plausible use. Always end with a single JSON object.
\end{lstlisting}
\end{tcolorbox}

\begin{tcolorbox}[
  enhanced,
  breakable,
  width=0.98\linewidth,
  colback=tzBlueFill,
  colframe=tzBlueBorder,
  boxrule=1.2pt,
  arc=6pt,
  left=5pt,right=5pt,top=4pt,bottom=2pt,
  title={\small Initial User/Task Prompt},
  coltitle=white,
  colbacktitle=tzBlueHeader2,
  fonttitle=\bfseries,
]
\small
\begin{lstlisting}[style=jsonTiny]
TASK
{task}

SCENARIO
{scenario}

YOUR OBJECTIVE
Choose the exact entity and exact part that can be repurposed to solve the task, then explain a physically plausible way to use it.

WHAT YOU KNOW AT THE START
You are given the environment scene image and the list of entity names below.
You do not know the part names yet. Part names are revealed only after you inspect an entity.

HOW TO EXPLORE
1. Inspect the environment image first and identify several plausible candidate entities.
2. Inspect multiple promising entities to reveal their exact part names.
3. Inspect multiple promising parts to compare affordances, geometry, material cues, attachment, reachability, and constraints.
4. Do not stop at the first plausible option. Explicitly explore different candidate entities and different candidate parts before answering whenever the best choice is not already decisive.
5. Ground every conclusion in visible attributes and any later part feedback. Avoid unsupported assumptions.

ACTION RULES
Return exactly one JSON object per turn.
Use `"action":"inspect_entity"` to inspect one exact entity name.
Use `"action":"inspect_part"` to inspect one exact part name after you have discovered it.
Use `"action":"answer"` only when you have explored enough and can justify the choice.
Inspect exactly one entity or one part per turn before answering.

ENTITIES AVAILABLE FOR INSPECTION
- {entity_name_1}
- {entity_name_2}
...

RETURN JSON ONLY IN ONE OF THESE FORMATS
Entity inspection:
{"reasoning":"...","action":"inspect_entity","entity":"<exact entity name>"}

Part inspection:
{"reasoning":"...","action":"inspect_part","part":"<exact part name>"}

Final answer:
{"reasoning":"...","action":"answer","answer_entity":"<exact entity name>","answer_part":"<exact part name>","answer_how_to_use":"..."}
\end{lstlisting}
\end{tcolorbox}

\begin{tcolorbox}[
  enhanced,
  breakable,
  width=0.98\linewidth,
  colback=tzBlueFill,
  colframe=tzBlueBorder,
  boxrule=1.2pt,
  arc=6pt,
  left=5pt,right=5pt,top=4pt,bottom=2pt,
  title={\small Entity Inspection Feedback},
  coltitle=white,
  colbacktitle=tzBlueHeader2,
  fonttitle=\bfseries,
]
\small
\begin{lstlisting}[style=jsonTiny]
ENTITY INSPECTION: {entity_name}
This entity includes these exact part names: {part_name_1}, {part_name_2}, ...
Use this inspection to identify which specific parts from this entity deserve closer checking.
Then continue exploring other plausible entities or parts as needed before deciding on the final answer.
\end{lstlisting}
\end{tcolorbox}

\begin{tcolorbox}[
  enhanced,
  breakable,
  width=0.98\linewidth,
  colback=tzBlueFill,
  colframe=tzBlueBorder,
  boxrule=1.2pt,
  arc=6pt,
  left=5pt,right=5pt,top=4pt,bottom=2pt,
  title={\small Part Inspection Feedback},
  coltitle=white,
  colbacktitle=tzBlueHeader2,
  fonttitle=\bfseries,
]
\small
\begin{lstlisting}[style=jsonTiny]
PART INSPECTION: {part_name}
Belongs to entity: {entity_name}
Physical text explanation: {physical_summary}
State text explanation: {state_summary}
Besides the text above, also carefully inspect the image for additional attributes and grounding cues. The text may not be complete.
For visibility and availability state especially, follow the text explanation even if the image appears clearer or more accessible. State text is the accurate source for whether the part is visible, free, blocked, or otherwise constrained in the task.
If you choose this part in the final answer, make sure your solution is consistent and grounded in all the available attributes.
Do not lock in immediately just because this part looks plausible. Continue comparing other plausible parts and entities before answering unless this candidate is already clearly best.
\end{lstlisting}
\end{tcolorbox}

\textbf{SFT Configuration.}
For SFT, we fine-tune Qwen3-VL-4B-Instruct and Qwen3-VL-8B-Instruct using LoRA with rank 4, applying adapters to all linear modules while keeping the vision tower frozen. We train for three epoch on 4 NVIDIA H100 GPUs with 80GB memory each, using a per-device batch size of 1 and 32 gradient accumulation steps. The learning rate is set to $5 \times 10^{-4}$ with cosine learning-rate decay and a warmup ratio of 0.1. The maximum sequence length is set to 32,768 tokens, with history masking enabled so that supervision is applied only to the target assistant responses. Images are resized under a maximum pixel budget of 65,536 pixels. Training uses BF16 precision, FlashAttention-2, gradient checkpointing, and DeepSpeed ZeRO-3 for memory-efficient optimization.

\begin{table}[!h]
\centering
\small
\begin{tabular}{lll}
\toprule
\textbf{Category} & \textbf{Hyperparameter} & \textbf{Value} \\
\midrule
Model & Image pixel limit & 65,536 \\
Model & Attention implementation & FlashAttention-2 \\
\midrule
Fine-tuning & Method & LoRA \\
Fine-tuning & LoRA rank & 4 \\
Fine-tuning & LoRA target modules & All linear modules \\
Fine-tuning & Vision tower & Frozen \\
\midrule
Data & Template & \texttt{qwen3\_vl\_nothink} \\
Data & Maximum sequence length & 32,768 tokens \\
\midrule
Optimization & Epochs & 3 \\
Optimization & Gradient accumulation steps & 32 \\
Optimization & Learning rate & $5 \times 10^{-4}$ \\
Optimization & Scheduler & Cosine decay \\
Optimization & Warmup ratio & 0.1 \\
Optimization & Precision & BF16 \\
Optimization & DeepSpeed & ZeRO-3 \\
\bottomrule
\end{tabular}
\vspace{2mm}
\caption{SFT training hyperparameters.}
\label{tab:sft_hyperparameters}
\end{table}

\textbf{DPO Configuration.}
For DPO, we initialize training from the base model or the SFT checkpoint and further optimize the model using LoRA with rank 4, again applying adapters to all linear modules while keeping the vision tower frozen. We train for three epoch on 4 NVIDIA H100 GPUs with 80GB memory each, using a per-device batch size of 1 and 16 gradient accumulation steps. The learning rate is set to $5 \times 10^{-6}$ with cosine learning-rate decay and a warmup ratio of 0.1. We use the sigmoid DPO loss with preference coefficient $\beta=0.1$, where the positive response is treated as the chosen sample and the negative or hard negative response is used as the rejected sample. The maximum sequence length is set to 32,768 tokens, and images are resized under a maximum pixel budget of 65,536 pixels. Training uses BF16 precision, FlashAttention-2, gradient checkpointing, and DeepSpeed ZeRO-3 for memory-efficient optimization.

\begin{table}[!h]
\centering
\small
\begin{tabular}{lll}
\toprule
\textbf{Category} & \textbf{Hyperparameter} & \textbf{Value} \\
\midrule
Model & Image pixel limit & 65,536 \\
Model & Attention implementation & FlashAttention-2 \\
\midrule
Fine-tuning & Method & LoRA \\
Fine-tuning & LoRA rank & 4 \\
Fine-tuning & LoRA target modules & All linear modules \\
Fine-tuning & Vision tower & Frozen \\
\midrule
Preference optimization & Preference loss & Sigmoid DPO loss \\
Preference optimization & Preference coefficient & $\beta = 0.1$ \\
Preference optimization & Chosen sample & Positive response \\
Preference optimization & Rejected sample & Negative / Hard negative response \\
\midrule
Data & Template & \texttt{qwen3\_vl\_nothink} \\
Data & Maximum sequence length & 32,768 tokens \\
\midrule
Optimization & Epochs & 3 \\
Optimization & Gradient accumulation steps & 16 \\
Optimization & Learning rate & $5 \times 10^{-6}$ \\
Optimization & Scheduler & Cosine decay \\
Optimization & Warmup ratio & 0.1 \\
Optimization & Precision & BF16 \\
Optimization & DeepSpeed & ZeRO-3 \\
\bottomrule
\end{tabular}
\vspace{2mm}
\caption{DPO training hyperparameters.}
\label{tab:dpo_hyperparameters}
\end{table}

\section{Analysis Details}
\label{apdx:analysis_details}

\subsection{Error Analysis Details}
\label{apdx:error_analysis_details}
We use GPT-5.4 to support automatic and scalable categorization of error cases. Before applying the model-based annotation, we manually annotated 50 cases to identify the primary reason for each failure. The agreement rate between the human annotations and the GPT-5.4 annotations was 92\%, suggesting that the model’s annotations are reliable and consistent with human judgment. We therefore use GPT-5.4 to annotate the remaining error cases. Specifically, we use the following prompt to identify both the primary reason for each failure and any additional contributing reasons.

\begin{tcolorbox}[
  enhanced,
  breakable,
  width=0.98\linewidth,
  colback=tzBlueFill,
  colframe=tzBlueBorder,
  boxrule=1.2pt,
  arc=6pt,
  left=5pt,right=5pt,top=4pt,bottom=2pt,
  title={\small Prompt for Error Categorization},
  coltitle=white,
  colbacktitle=tzBlueHeader2,
  fonttitle=\bfseries,
]
\small
\begin{lstlisting}[style=jsonTiny]
You are a careful judge for error analysis in creative physical tool-use tasks.

You will be given:
- the task description,
- the gold entity image,
- the model's predicted entity, part, and how-to-use text,
- the ground-truth entity, part, and gold solution usage text,
- and a heuristic explaining why the gold is better than the predicted part.

Classify the model error using the taxonomy below.

Error taxonomy:
- A. Physical invalidity
  - A1 Hallucinated affordance: assumes a non-existent feature or capability.
  - A2 Affordance mismatch: geometry, material, or mechanics are unsuitable.
  - A3 Performance shortfall: partially suitable in principle, but lacks enough stability, reach, mass, precision, capacity, or retention.
- B. Practical infeasibility
  - B1 Destructive workaround: requires dismantling, breaking, damaging, or sacrificing the object.
  - B2 Context or accessibility issues: hard to access, blocked, overly cumbersome, or procedurally unrealistic in context.
- C. Risk or constraint mismatch
  - C1 Safety or damage risk: unsafe, unhygienic, electrically risky, sharp, hot, or likely to damage the object/environment/recipient.
  - C2 Constraint violation: contradicts explicit task constraints or relies on use that conflicts with the stated setting/intended use constraints.

Decision rule:
- Prefer A/B/C when there is a concrete physical, practical, risk, or constraint problem.
- Predict exactly one major reason code.
- Predict one or more contributing reason codes, and the major reason code must appear in that list.

Input case:
Task:
{task_text}

Gold entity image:
Attached separately if available.

Model prediction:
- Predicted entity: {pred_entity}
- Predicted part: {pred_part}
- Predicted how to use: {pred_how}

Ground truth:
- Gold entity: {gold_entity}
- Gold part: {gold_part}
- Gold how_to_apply: {gold_how}

Important heuristic from the task data:
{gold_change_reason}

Instructions:
1. Judge the predicted solution against the task and the gold solution.
2. Use the gold image to understand what the gold object is.
3. Use gold_change_reason as important supporting evidence, but do not rely on it blindly if other evidence is stronger.
4. Pick exactly one major reason code.
5. Pick all contributing reason codes that materially apply.
6. The contributing_reason_codes list must include the major_reason_code.
7. Return JSON only.

Return exactly one JSON object with this schema:
{
  "reasoning": "...",
  "major_reason_code": "A1|A2|A3|B1|B2|C1|C2",
  "major_reason_label": "...",
  "contributing_reason_codes": ["..."],
  "contributing_reason_labels": ["..."]
}
\end{lstlisting}
\end{tcolorbox}

\subsection{Case Study Details}
\label{apdx:case_analysis_details}

\begin{tcolorbox}[title={Case A: Wall Protection from a Metal Hook}, breakable, colback=white, colframe=black!60]
\begin{lstlisting}[style=jsonTiny]
{
  "case": "wall_protection",
  "task": "I am in my bathroom trying to stop a small metal hook from leaving a mark on the painted wall next to the sink. The hook is for hanging a hand towel, but it keeps pressing into the same spot and I do not want the paint to get dented or scraped. What can I use?",
  "gold": {
    "entity": "curved tension shower curtain rod",
    "part": "non_slip_end_pads"
  }
}
\end{lstlisting}

\vspace{2pt}
{\centering
\includegraphics[width=0.6\linewidth]{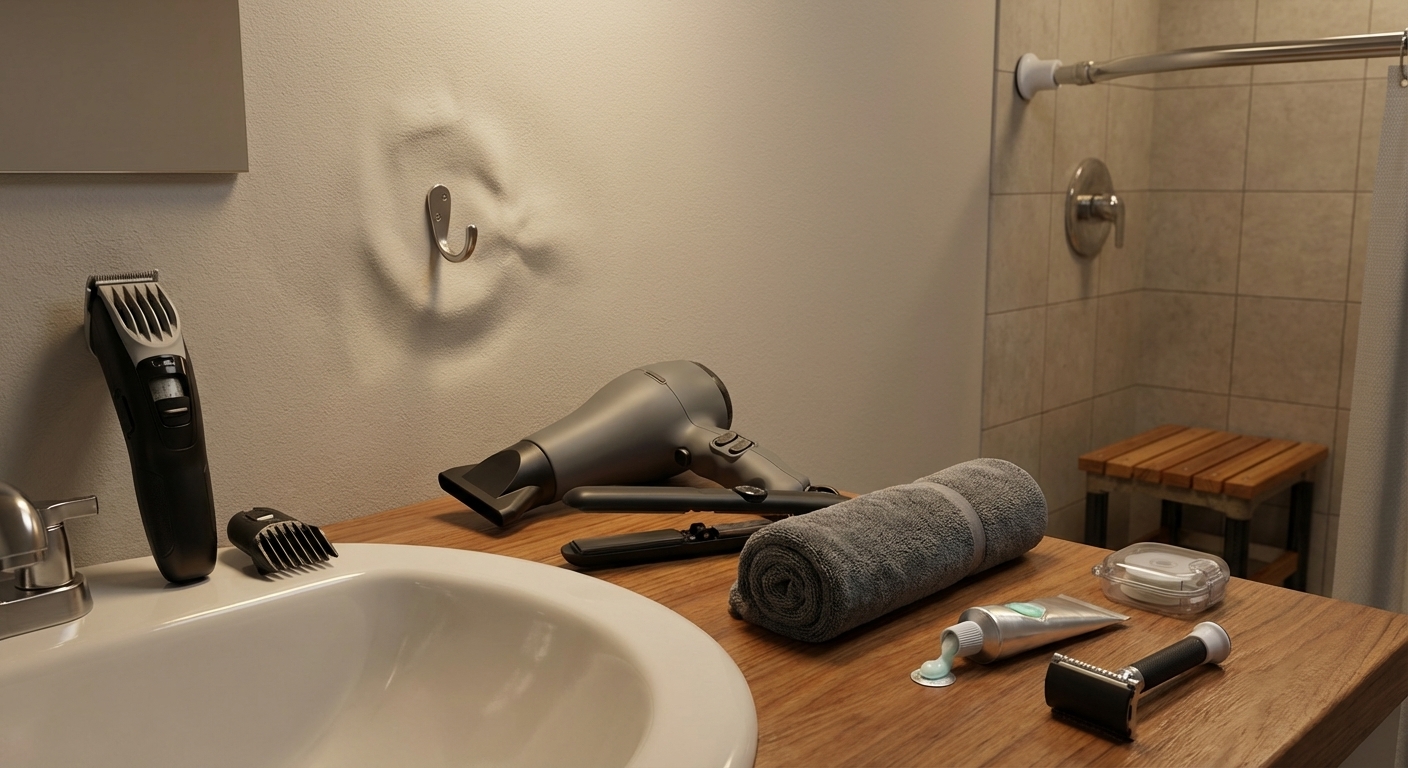}
\par}
\vspace{2pt}

\begin{lstlisting}[style=jsonTiny]
{
  "trace": "bad_base_8b",
  "round": 2,
  "model_response": {
    "reasoning": "The microfiber_pile_surface is the most relevant part for this task because its soft, dense, and absorbent texture makes it ideal for acting as a buffer between the metal hook and the wall. It can conform to the hook's shape and distribute pressure, preventing paint damage. I will inspect this part to confirm its physical properties and how it can be used.",
    "action": "inspect_part",
    "part": "microfiber_pile_surface"
  },
  "feedback": {
    "type": "PART INSPECTION",
    "part": "microfiber_pile_surface",
    "entity": "microfiber hand towel",
    "physical": "microfiber blend (polyester/polyamide) with embedded carbon/antimicrobial additive fibers; sturdy; stretchable; very plush, high-absorbency, slightly higher drag; reduced static cling and reduced odor retention compared to standard pile; moderate",
    "state": "partially visible; partially blocked; slightly wet; slightly hot"
  }
}
\end{lstlisting}

\vspace{2pt}
{\centering
\includegraphics[width=0.6\linewidth]{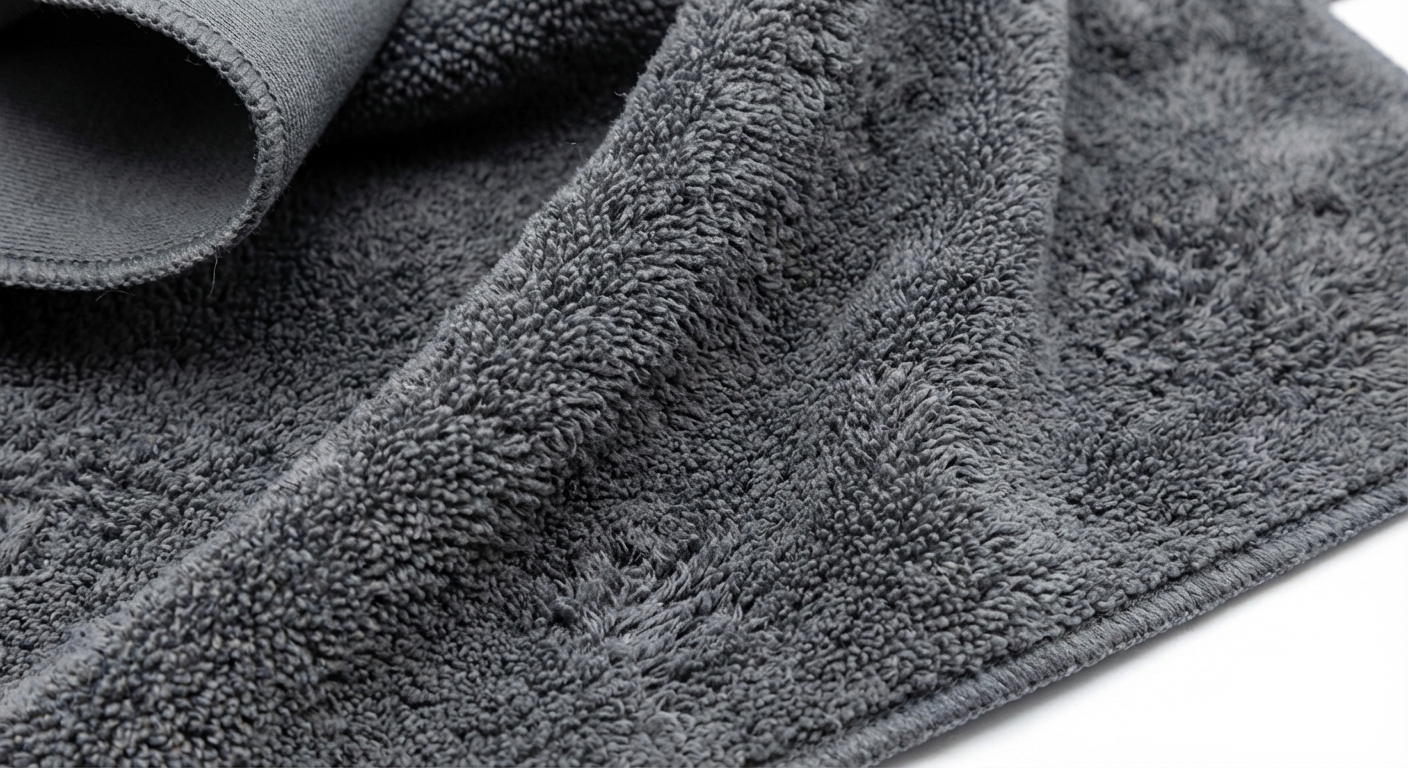}
\par}
\vspace{2pt}

\begin{lstlisting}[style=jsonTiny]
{
  "trace": "bad_base_8b",
  "round": 3,
  "model_response": {
    "reasoning": "The microfiber_pile_surface is highly suitable for this task due to its soft, plush, and absorbent nature, which can cushion the hook's pressure on the wall. However, to ensure this is the best option, I should explore other entities that might offer a more rigid or durable buffer. ... The curved tension shower curtain rod is rigid and would not help. ... Therefore, the microfiber_pile_surface remains the best candidate. I will now answer.",
    "action": "answer",
    "answer_entity": "microfiber hand towel",
    "answer_part": "microfiber_pile_surface",
    "answer_how_to_use": "Place a small piece of the microfiber_pile_surface directly between the metal hook and the wall, ensuring it conforms to the hook's shape. This will act as a soft buffer, distributing pressure and preventing the hook from pressing into the paint."
  },
  "why_bad": "The model generalizes from softness and dismisses the shower rod as rigid without inspecting the rubber end pads."
}
\end{lstlisting}

\begin{lstlisting}[style=jsonTiny]
{
  "trace": "good_sft_dpo_8b",
  "round": 8,
  "model_response": {
    "reasoning": "The knurled_handle is made of rubberized material and is textured for grip, but it is not soft or pliable enough to cushion the hook. ... None of the parts on the double-edge safety razor are suitable for this task. I will now inspect the curved tension shower curtain rod to see if any part can be repurposed as a buffer.",
    "action": "inspect_entity",
    "entity": "curved tension shower curtain rod"
  },
  "feedback": {
    "type": "ENTITY INSPECTION",
    "entity": "curved tension shower curtain rod",
    "parts": ["curved_outer_tube", "inner_telescoping_tube", "twist_lock_collar", "compression_spring", "end_cap_housings", "non_slip_end_pads"]
  }
}
\end{lstlisting}

\vspace{2pt}
{\centering
\includegraphics[width=0.6\linewidth]{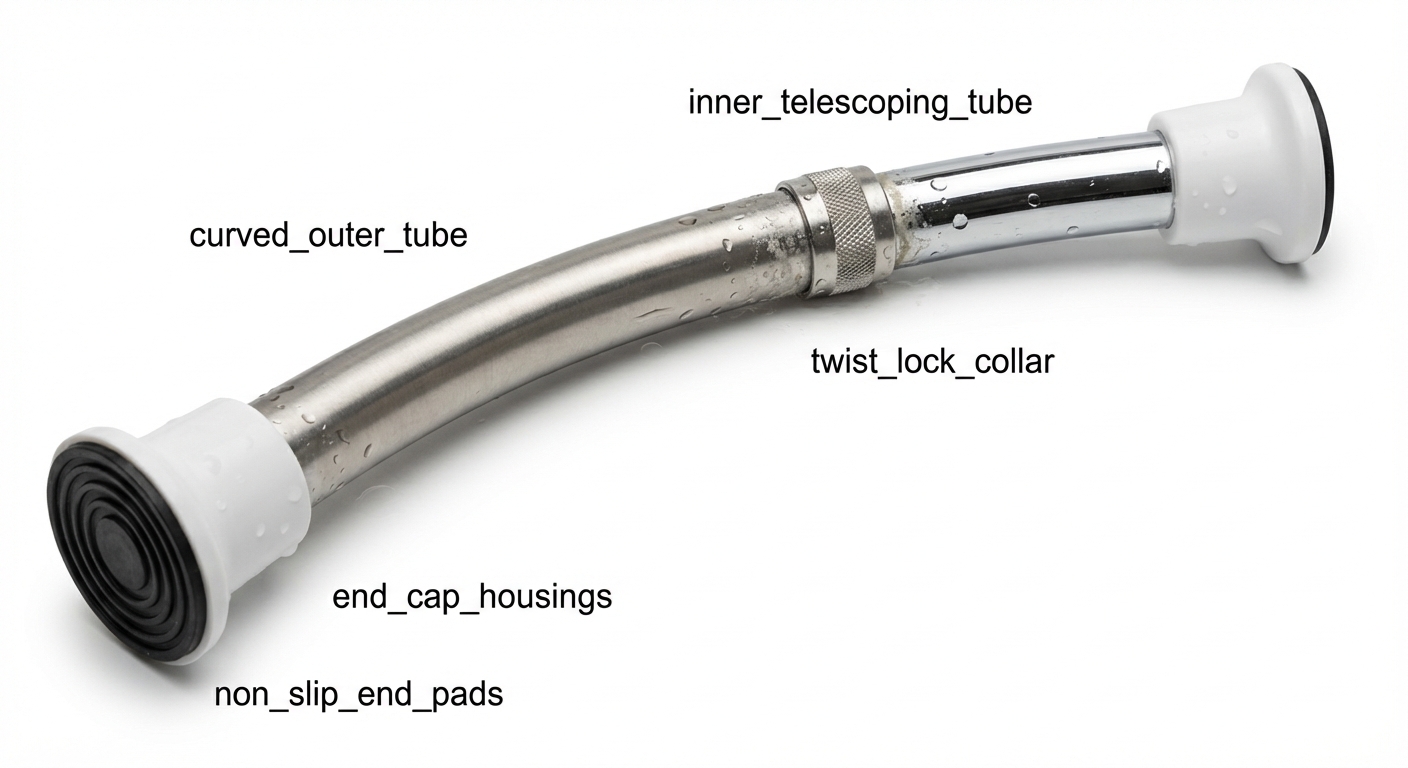}
\par}
\vspace{2pt}

\begin{lstlisting}[style=jsonTiny]
{
  "trace": "good_sft_dpo_8b",
  "round": 9,
  "model_response": {
    "reasoning": "The non_slip_end_pads are made of soft rubber and are designed to grip surfaces without slipping. They are pliable and can be repurposed to cushion the hook against the wall. ... The non_slip_end_pads are the best candidate so far, as they are soft, pliable, and designed to absorb pressure.",
    "action": "inspect_part",
    "part": "non_slip_end_pads"
  },
  "feedback": {
    "type": "PART INSPECTION",
    "part": "non_slip_end_pads",
    "entity": "curved tension shower curtain rod",
    "physical": "EPDM rubber; soft; sturdy; stretchable; high-friction, slightly tacky; water-tolerant; very light; exact diameter in mm",
    "state": "partially visible; partially blocked"
  }
}
\end{lstlisting}

\vspace{2pt}
{\centering
\includegraphics[width=0.6\linewidth]{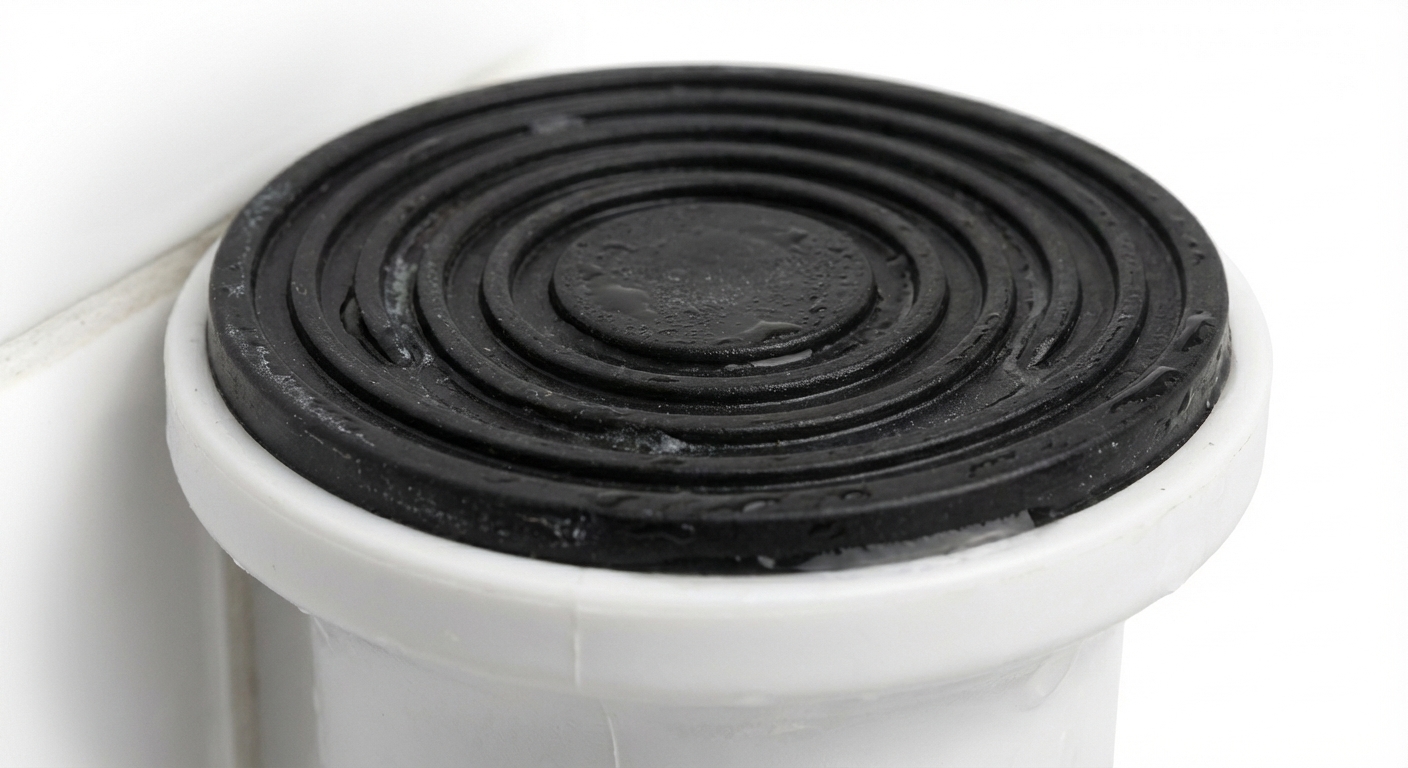}
\par}
\vspace{2pt}

\begin{lstlisting}[style=jsonTiny]
{
  "trace": "good_sft_dpo_8b",
  "round": 10,
  "model_response": {
    "reasoning": "The non_slip_end_pads are made of soft EPDM rubber, which is highly pliable, water-tolerant, and designed to absorb pressure without slipping. This makes them ideal for cushioning the metal hook against the painted wall. While the microfiber_pile_surface is also soft and absorbent, the non_slip_end_pads are more durable and specifically designed for pressure absorption in a high-friction environment. ... The non_slip_end_pads are more suitable because they are designed for cushioning and pressure distribution, and their rubber material is more resilient than the microfiber, which may compress or deform over time.",
    "action": "answer",
    "answer_entity": "curved tension shower curtain rod",
    "answer_part": "non_slip_end_pads",
    "answer_how_to_use": "Place one of the non_slip_end_pads over the metal hook, pressing it firmly against the wall. The soft, pliable rubber will absorb the pressure from the hook, preventing it from pressing into the painted wall and causing dents or scrapes. The pad's high-friction surface will also ensure it stays in place without slipping."
  },
  "capability_shown": "attribute grounding: material, friction, pressure distribution, and contact-point geometry"
}
\end{lstlisting}
\end{tcolorbox}

\begin{tcolorbox}[title={Case B: Wrapping Paper Straight-Edge and Backing Board}, breakable, colback=white, colframe=black!60]
\begin{lstlisting}[style=jsonTiny]
{
  "case": "wrapping_paper_straight_edge",
  "task": "I am in my bedroom packing a gift for a friend, and I need to trim a sheet of wrapping paper so it fits neatly around a small box. I do not have a ruler or cutting mat handy, and I want to keep the paper from slipping while I mark it. What can I use?",
  "gold": {
    "entity": "under-bed storage bin with zipper lid",
    "part": "lid_panel"
  }
}
\end{lstlisting}

\vspace{2pt}
{\centering
\includegraphics[width=0.6\linewidth]{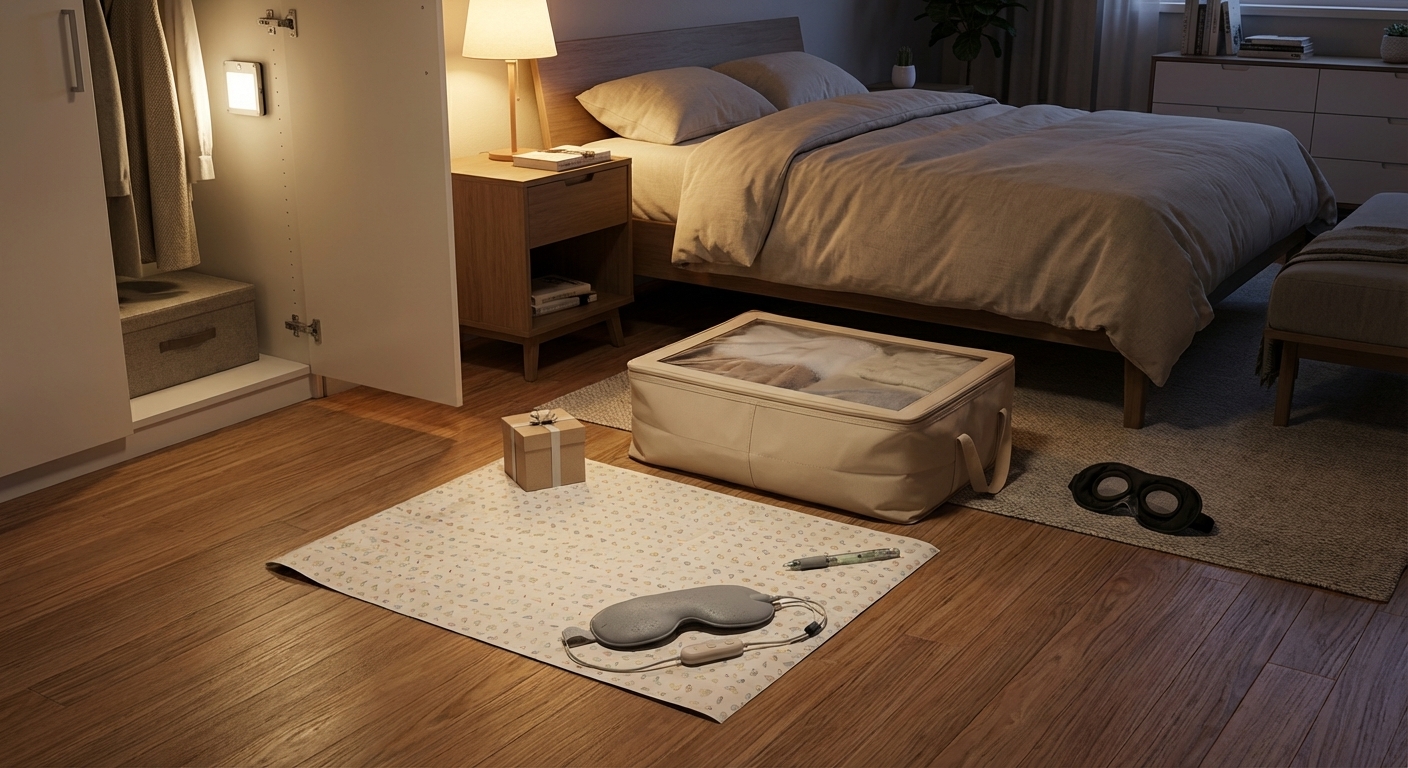}
\par}
\vspace{2pt}

\begin{lstlisting}[style=jsonTiny]
{
  "trace": "bad_base_4b",
  "round": 6,
  "model_response": {
    "reasoning": "The rubber grip sleeve is textured and designed to provide a secure, non-slip hold, making it ideal for stabilizing the wrapping paper while marking. Its flexible yet firm material can conform to the paper's surface, preventing it from slipping. ... I will now inspect the rubber grip sleeve to confirm its physical properties and suitability for the task.",
    "action": "inspect_part",
    "part": "rubber_grip_sleeve"
  },
  "feedback": {
    "type": "PART INSPECTION",
    "part": "rubber_grip_sleeve",
    "entity": "gel ink pen with rubber grip",
    "physical": "closed-cell foam layer bonded to a thin elastomer skin; anti-slip indexing function; EVA foam core with polyurethane (PU) elastomer skin; normal; springy; soft, grippy stipple; very light",
    "state": "visible; free; slightly hot"
  }
}
\end{lstlisting}

\vspace{2pt}
{\centering
\includegraphics[width=0.6\linewidth]{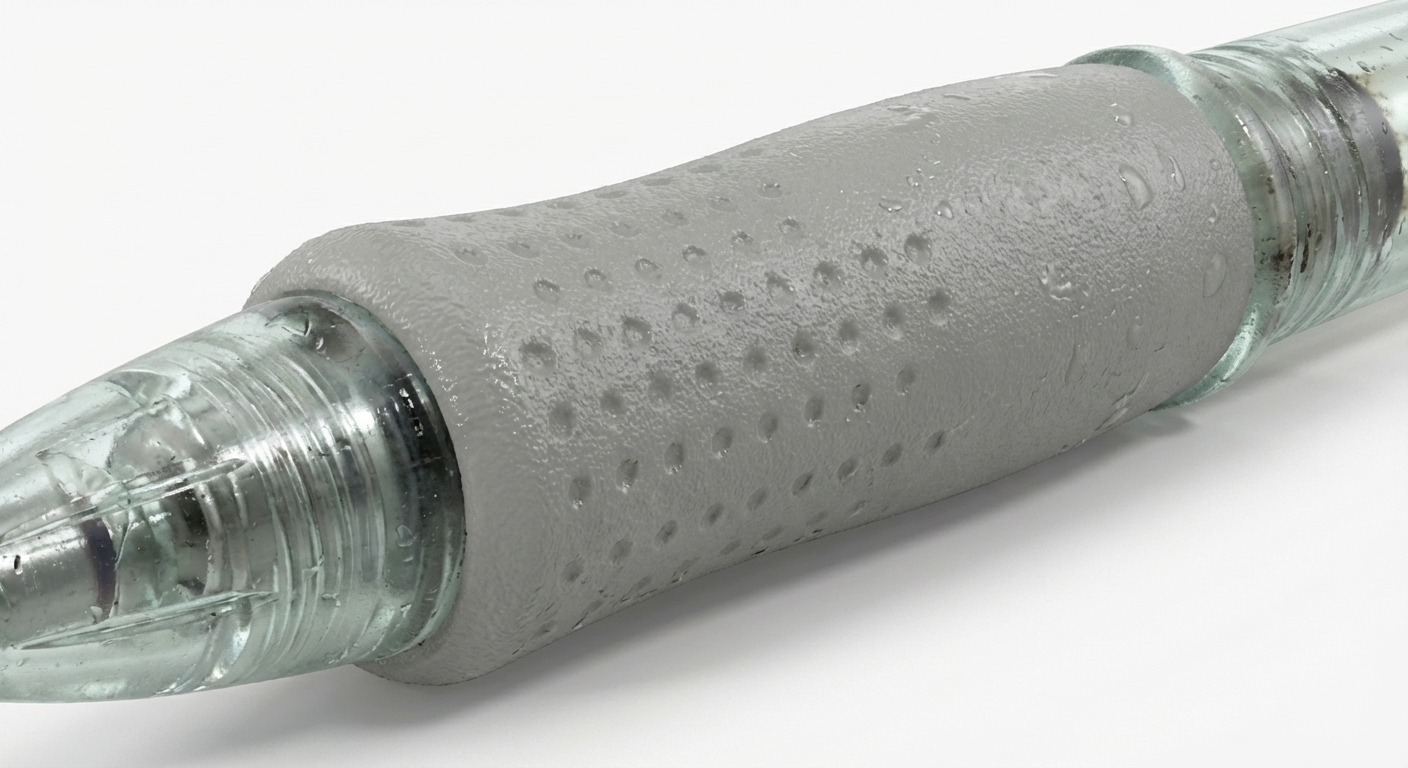}
\par}
\vspace{2pt}

\begin{lstlisting}[style=jsonTiny]
{
  "trace": "bad_base_4b",
  "round": 10,
  "model_response": {
    "reasoning": "The elastic strap is made of a woven, slightly grippy fabric with a stretchy, adjustable design, which makes it ideal for securing the wrapping paper to the mask. ... I will now inspect the gel ink pen's rubber grip sleeve to see if it can be repurposed to hold the paper in place.",
    "action": "inspect_part",
    "part": "rubber_grip_sleeve"
  },
  "feedback": {
    "summary": "Same part and same feedback as Round 6."
  }
}
\end{lstlisting}

\vspace{2pt}
{\centering
\includegraphics[width=0.6\linewidth]{figures/images_case/wrapping_base_rubber_grip_sleeve.png}
\par}
\vspace{2pt}

\begin{lstlisting}[style=jsonTiny]
{
  "trace": "bad_base_4b",
  "round": 50,
  "model_response": {
    "reasoning": "After thorough inspection of all candidate parts, the rubber grip sleeve of the gel ink pen is the most suitable for preventing the wrapping paper from slipping while marking. It is soft, grippy, and flexible, with a stippled surface that provides excellent friction against the paper. ... The rubber grip sleeve is the optimal choice for securing the paper during marking.",
    "action": "answer",
    "answer_entity": "gel ink pen with rubber grip",
    "answer_part": "rubber_grip_sleeve",
    "answer_how_to_use": "Use the rubber grip sleeve to press against the wrapping paper, positioning it along the edge of the box to prevent slippage while marking the dimensions with a pencil or marker."
  },
  "why_bad": "The model loops on local friction. It never finds a large flat straight-edge/backing surface."
}
\end{lstlisting}

\begin{lstlisting}[style=jsonTiny]
{
  "trace": "good_sft_dpo_8b",
  "round": 1,
  "model_response": {
    "reasoning": "The task requires a flat, stable surface or straight edge to keep wrapping paper from slipping while marking. The under-bed storage bin with zipper lid has a large, flat lid panel that could serve as a makeshift cutting mat or straight edge. I will inspect this entity to reveal its parts and confirm whether the lid panel is suitable.",
    "action": "inspect_entity",
    "entity": "under-bed storage bin with zipper lid"
  },
  "feedback": {
    "type": "ENTITY INSPECTION",
    "entity": "under-bed storage bin with zipper lid",
    "parts": ["bin_body", "base_stiffener_panel", "zipper_track", "zipper_slider_pull", "lid_panel", "clear_window_panel", "carry_handles"]
  }
}
\end{lstlisting}

\vspace{2pt}
{\centering
\includegraphics[width=0.6\linewidth]{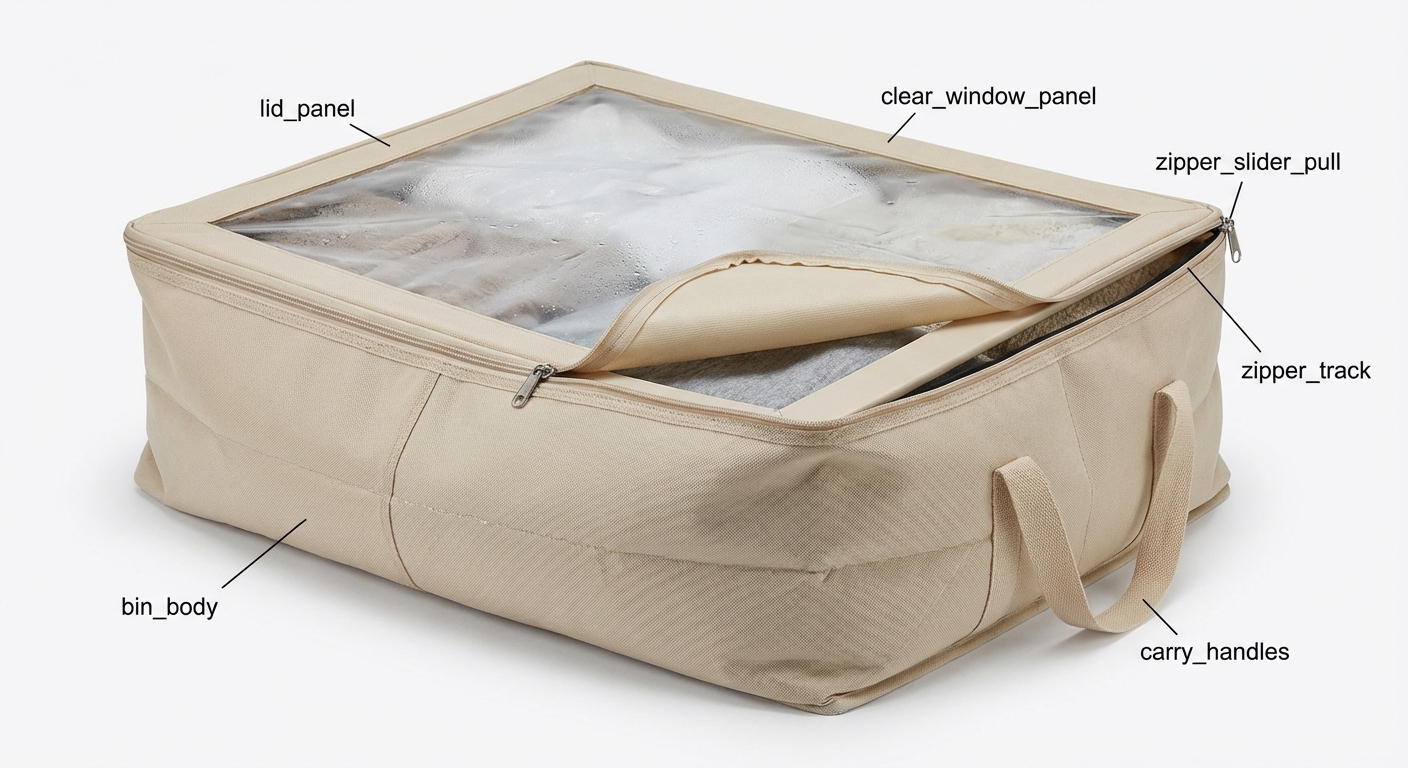}
\par}
\vspace{2pt}

\begin{lstlisting}[style=jsonTiny]
{
  "trace": "good_sft_dpo_8b",
  "round": 2,
  "model_response": {
    "reasoning": "The lid_panel is the most relevant part because it is likely flat, rigid, and large enough to serve as a stable backing or straight edge for marking and trimming wrapping paper. The clear_window_panel might be transparent but may be flexible; the base_stiffener_panel could also be rigid but may be less accessible. I will inspect the lid_panel to confirm its rigidity and surface properties.",
    "action": "inspect_part",
    "part": "lid_panel"
  },
  "feedback": {
    "type": "PART INSPECTION",
    "part": "lid_panel",
    "entity": "under-bed storage bin with zipper lid",
    "physical": "medium thickness; internal stiffener layer; window cutout with double-stitch topstitching; laminated polyester/nylon fabric with an internal polypropylene sheet stiffener; semi-rigid; very sturdy; non-elastic; smooth laminated wipe-clean finish with slight sheen; moderate",
    "state": "partially visible; partially blocked (easily freed by hand); full"
  }
}
\end{lstlisting}

\vspace{2pt}
{\centering
\includegraphics[width=0.6\linewidth]{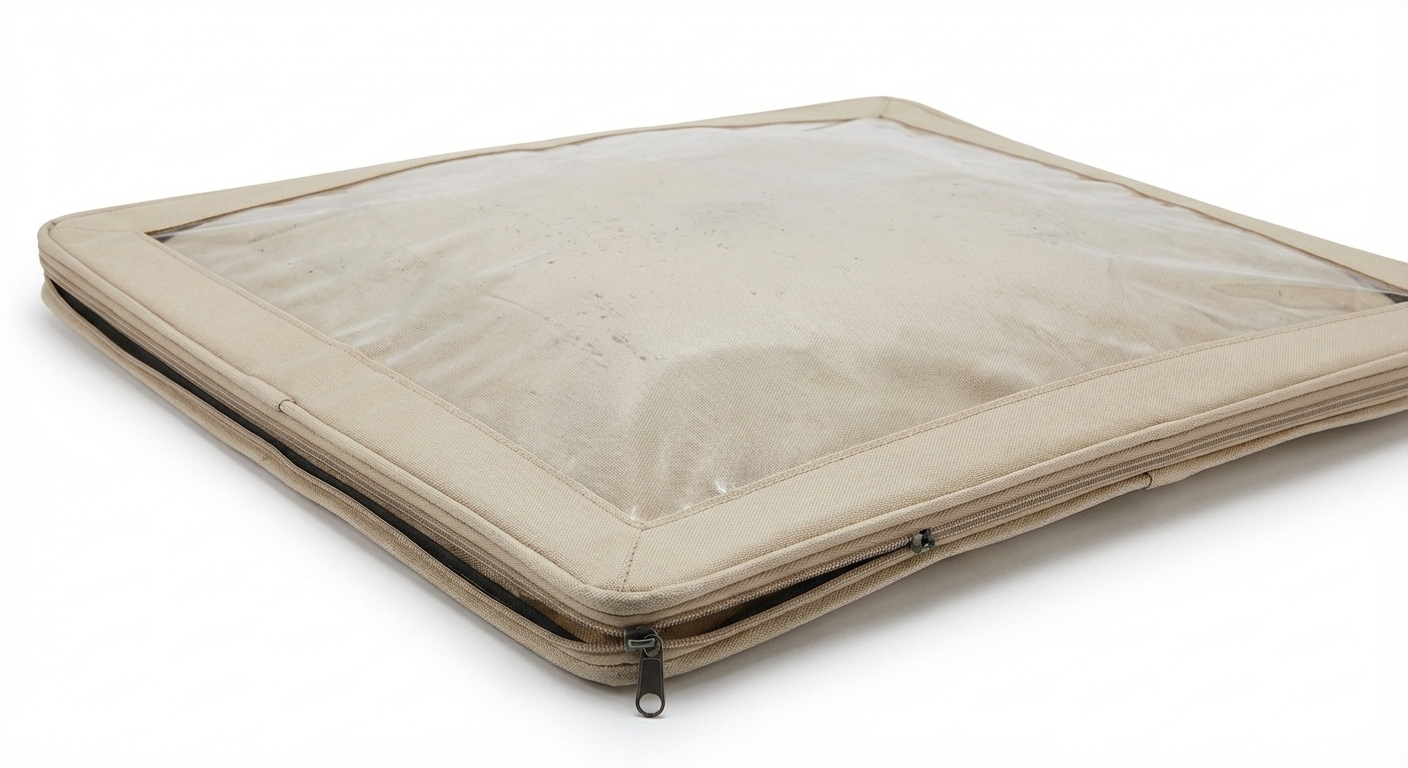}
\par}
\vspace{2pt}

\begin{lstlisting}[style=jsonTiny]
{
  "trace": "good_sft_dpo_8b",
  "round": 3,
  "model_response": {
    "reasoning": "The lid_panel is semi-rigid, sturdy, smooth, and wipe-clean, making it ideal for marking wrapping paper without slipping. Its laminated finish and internal stiffener ensure it won't bend or warp under pressure, and its size is large enough to accommodate a sheet of wrapping paper. ... The gel ink pen with rubber grip is a candidate for marking, but I need a stable surface, not a tool. ... Therefore, the lid_panel is the best available option.",
    "action": "answer",
    "answer_entity": "under-bed storage bin with zipper lid",
    "answer_part": "lid_panel",
    "answer_how_to_use": "Place the sheet of wrapping paper on the lid_panel, which is flat, sturdy, and non-slip. Use the gel ink pen to measure and mark the paper accurately, as the lid's smooth, rigid surface will prevent the paper from shifting while you work."
  },
  "capability_shown": "part-level geometry: backing-board and straight-edge reasoning instead of local anti-slip reasoning"
}
\end{lstlisting}
\end{tcolorbox}

\begin{tcolorbox}[title={Case C: Sink Overflow Slot Cleaning}, breakable, colback=white, colframe=black!60]
\begin{lstlisting}[style=jsonTiny]
{
  "case": "sink_overflow_slot_cleaning",
  "task": "I am in my bathroom and the small opening in the sink overflow slot is packed with damp hair-and-soap gunk. Water is starting to smell bad, and I want to loosen the mess enough to rinse it out, but I do not have any proper cleaning picks nearby. What can I use and how?",
  "gold": {
    "entity": "electric beard trimmer with adjustable guard",
    "part": "adjustable_guard_comb"
  }
}
\end{lstlisting}

\vspace{2pt}
{\centering
\includegraphics[width=0.6\linewidth]{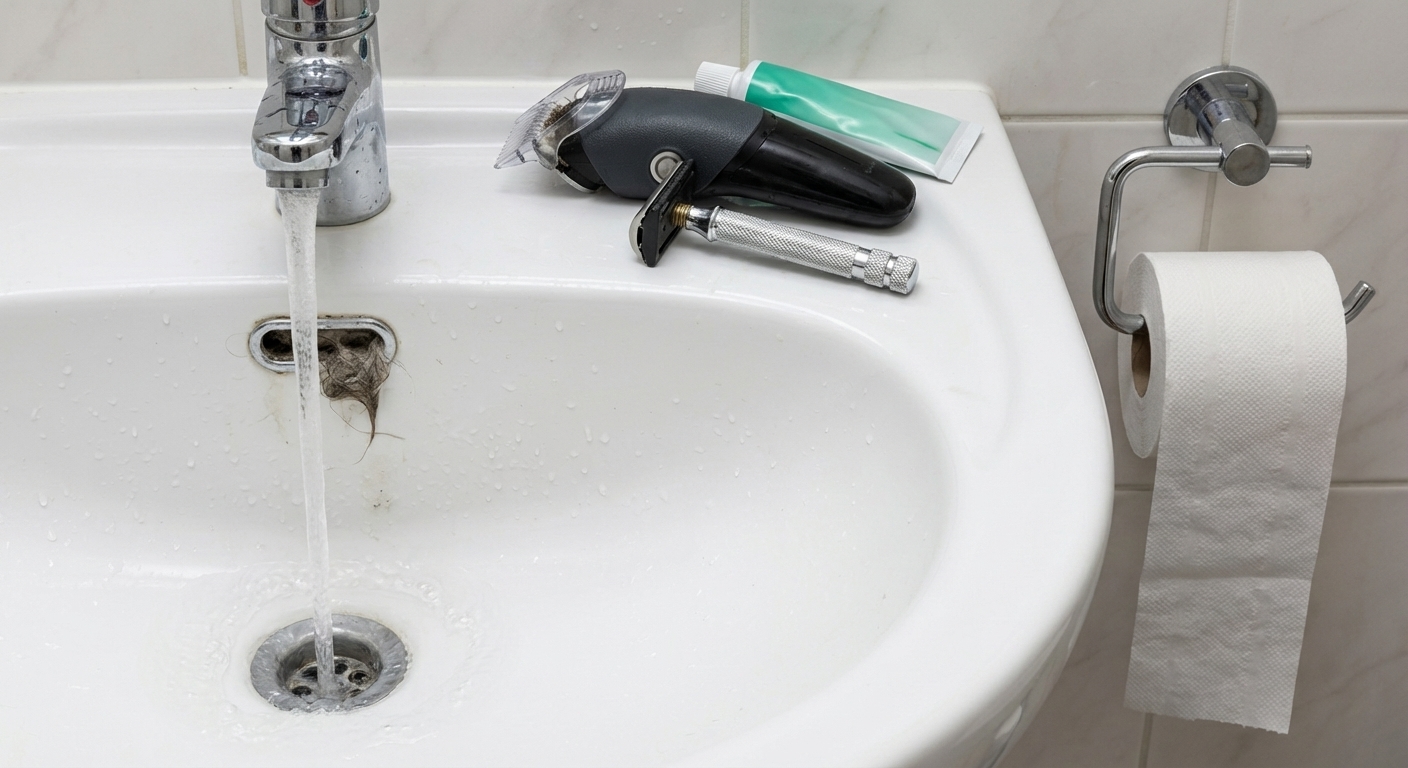}
\par}
\vspace{2pt}

\begin{lstlisting}[style=jsonTiny]
{
  "trace": "bad_base_4b",
  "round": 2,
  "model_response": {
    "reasoning": "The cutting_blade_head is the most suitable part to use for loosening the gunk in the sink overflow slot. It is made of metal and has a sharp, rigid edge that can effectively scrape or dislodge debris. The adjustable_guard_comb is plastic and not suitable for scraping. ... The cutting_blade_head is the only part that meets the physical requirements for this task.",
    "action": "inspect_part",
    "part": "cutting_blade_head"
  },
  "feedback": {
    "type": "PART INSPECTION",
    "part": "cutting_blade_head",
    "entity": "electric beard trimmer with adjustable guard",
    "physical": "rear coupling slot and molded alignment details; exact width in mm; exact thickness; integrated alignment posts, rear coupling slot, molded ceramic shoulder; zirconia ceramic fixed blade, stainless steel moving cutter, polymer/metal backing frame; sturdy; low-friction sliding interface, thermally insulating behavior; light",
    "state": "partially visible; blocked; slightly cold"
  }
}
\end{lstlisting}

\vspace{2pt}
{\centering
\includegraphics[width=0.6\linewidth]{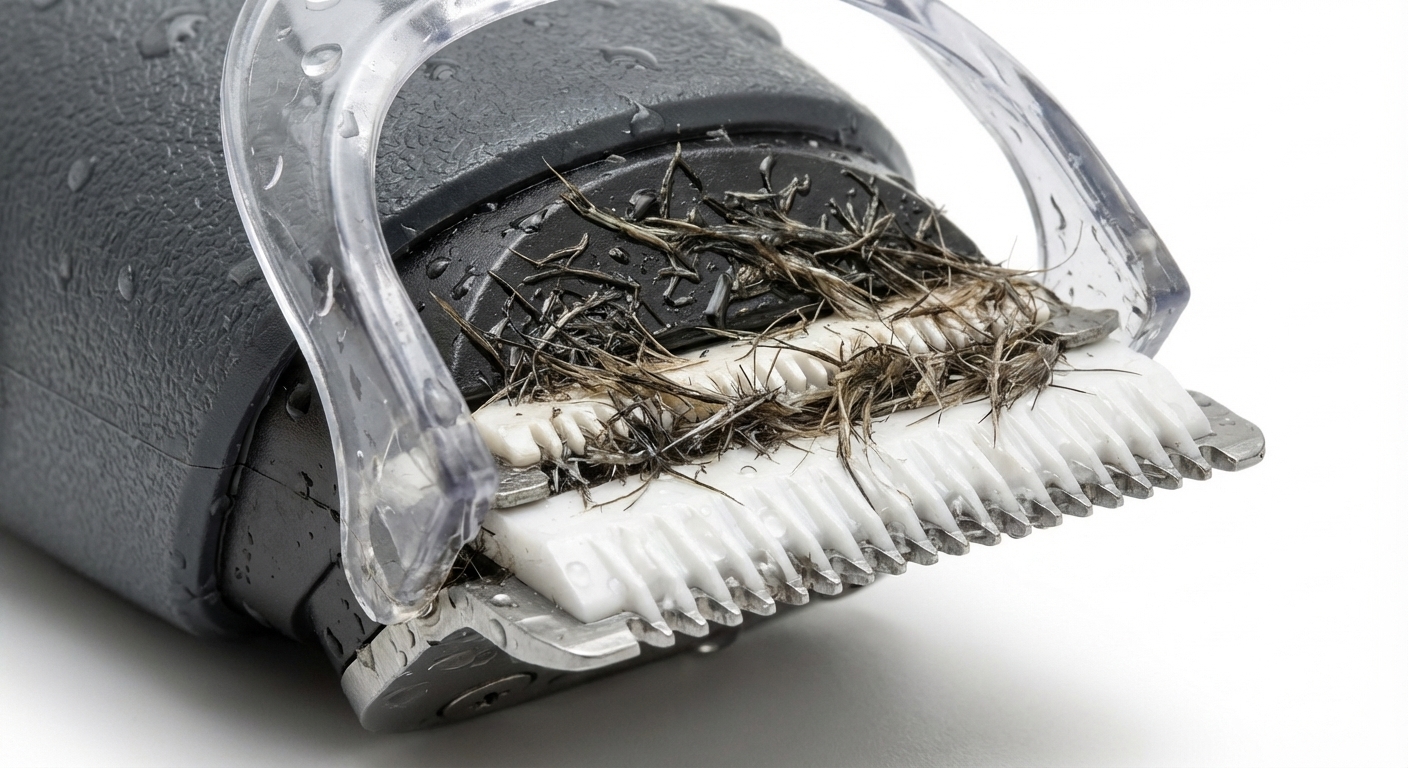}
\par}
\vspace{2pt}

\begin{lstlisting}[style=jsonTiny]
{
  "trace": "bad_base_4b",
  "round": 4,
  "model_response": {
    "reasoning": "The double_edge_blade is the most suitable part for loosening the gunk in the sink overflow slot. It is made of stainless steel and is sharp, rigid, and designed for scraping. ... The double_edge_blade is the only part that meets the physical requirements for this task.",
    "action": "inspect_part",
    "part": "double_edge_blade"
  },
  "feedback": {
    "type": "PART INSPECTION",
    "part": "double_edge_blade",
    "entity": "double-edge safety razor with knurled handle",
    "physical": "high-carbon steel, uncoated or lightly coated; normal durability; springy; more reactive and can develop patina/rust spots if left wet",
    "state": "partially visible and blocked; slightly hot"
  }
}
\end{lstlisting}

\vspace{2pt}
{\centering
\includegraphics[width=0.6\linewidth]{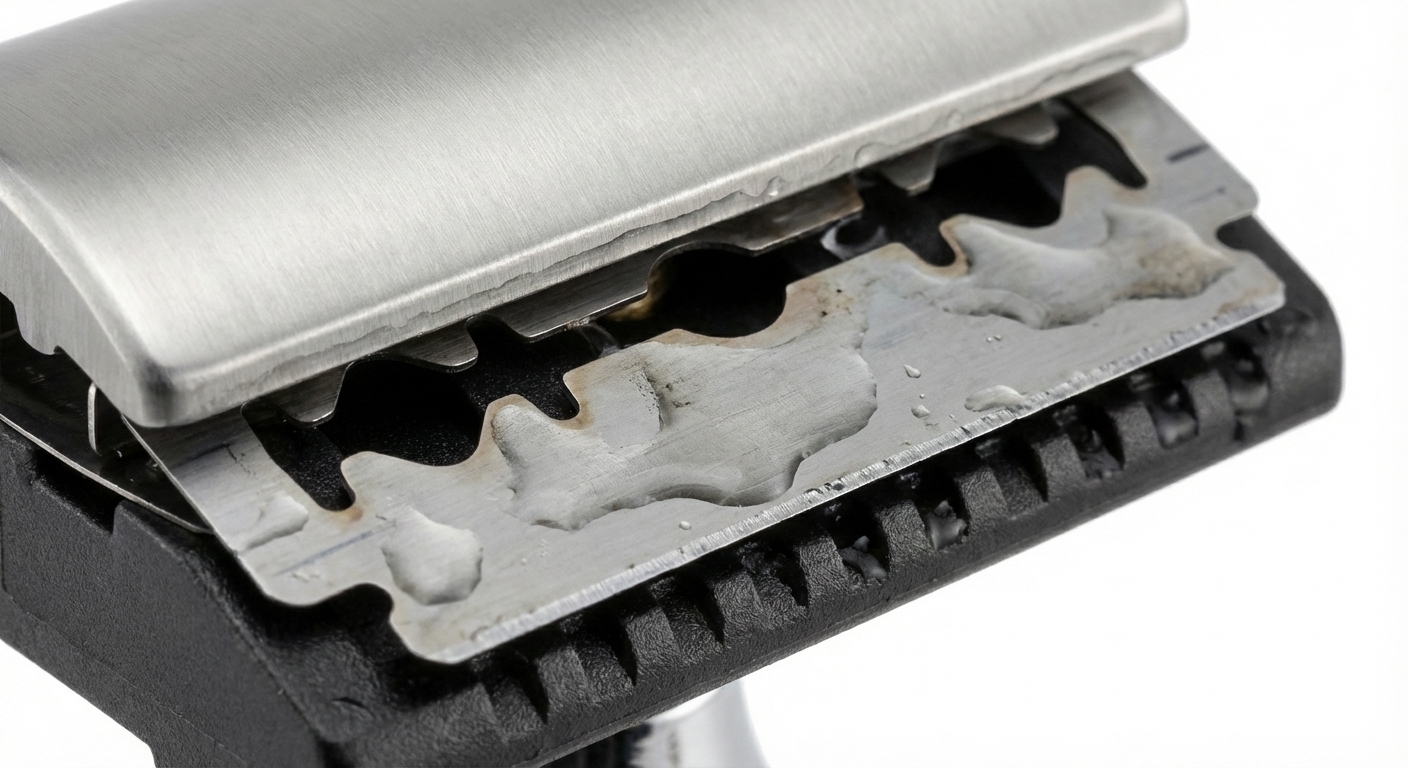}
\par}
\vspace{2pt}

\begin{lstlisting}[style=jsonTiny]
{
  "trace": "bad_base_4b",
  "round": 5,
  "model_response": {
    "reasoning": "The double_edge_blade is highly suitable for loosening the gunk in the sink overflow slot. It is made of high-carbon steel, which is sharp, rigid, and durable, making it ideal for scraping debris. The blade is partially visible and slightly hot, indicating it is not obstructed and can be used safely. ... The double_edge_blade is the best choice.",
    "action": "answer",
    "answer_entity": "double-edge safety razor with knurled handle",
    "answer_part": "double_edge_blade",
    "answer_how_to_use": "Use the double_edge_blade to gently scrape the gunk from the sink overflow slot by inserting the blade into the slot and moving it back and forth to loosen the debris. The sharp, rigid edge will effectively dislodge the hair and soap gunk, allowing water to rinse it out."
  },
  "why_bad": "The model prefers sharp scraping despite blocked/hot state cues and despite the soft obstruction being better handled by nonmetal teeth."
}
\end{lstlisting}

\begin{lstlisting}[style=jsonTiny]
{
  "trace": "good_sft_dpo_4b",
  "round": 2,
  "model_response": {
    "reasoning": "The double-edge blade is sharp and designed for cutting, making it ideal for scraping gunk from the narrow overflow slot. ... I should inspect the blade next to confirm its suitability for loosening gunk, then compare with other candidates like the electric beard trimmer's small parts or the toilet paper roll's edge for leverage.",
    "action": "inspect_part",
    "part": "double_edge_blade"
  },
  "feedback": {
    "type": "PART INSPECTION",
    "part": "double_edge_blade",
    "entity": "double-edge safety razor with knurled handle",
    "physical": "high-carbon steel, uncoated or lightly coated; normal durability; springy; more reactive and can develop patina/rust spots if left wet",
    "state": "partially visible and blocked; slightly hot"
  },
  "note": "The trained model still explores the tempting blade, but does not terminate there."
}
\end{lstlisting}

\vspace{2pt}
{\centering
\includegraphics[width=0.6\linewidth]{figures/images_case/sink_base_double_edge_blade.png}
\par}
\vspace{2pt}

\begin{lstlisting}[style=jsonTiny]
{
  "trace": "good_sft_dpo_4b",
  "round": 4,
  "model_response": {
    "reasoning": "The knurled handle is chrome-plated brass, corrosion-resistant, and heavy -- providing strong grip and leverage. ... However, as a standalone part, it lacks a cutting edge and cannot scrape gunk alone. ... I should now inspect the electric beard trimmer's adjustable guard, as its small, rigid shape might fit into the slot and help dislodge gunk.",
    "action": "inspect_entity",
    "entity": "electric beard trimmer with adjustable guard"
  },
  "feedback": {
    "type": "ENTITY INSPECTION",
    "entity": "electric beard trimmer with adjustable guard",
    "parts": ["housing_body", "cutting_blade_head", "adjustable_guard_comb", "power_switch", "rechargeable_battery_pack", "motor_drive_unit", "control_and_charging_electronics"]
  }
}
\end{lstlisting}

\vspace{2pt}
{\centering
\includegraphics[width=0.6\linewidth]{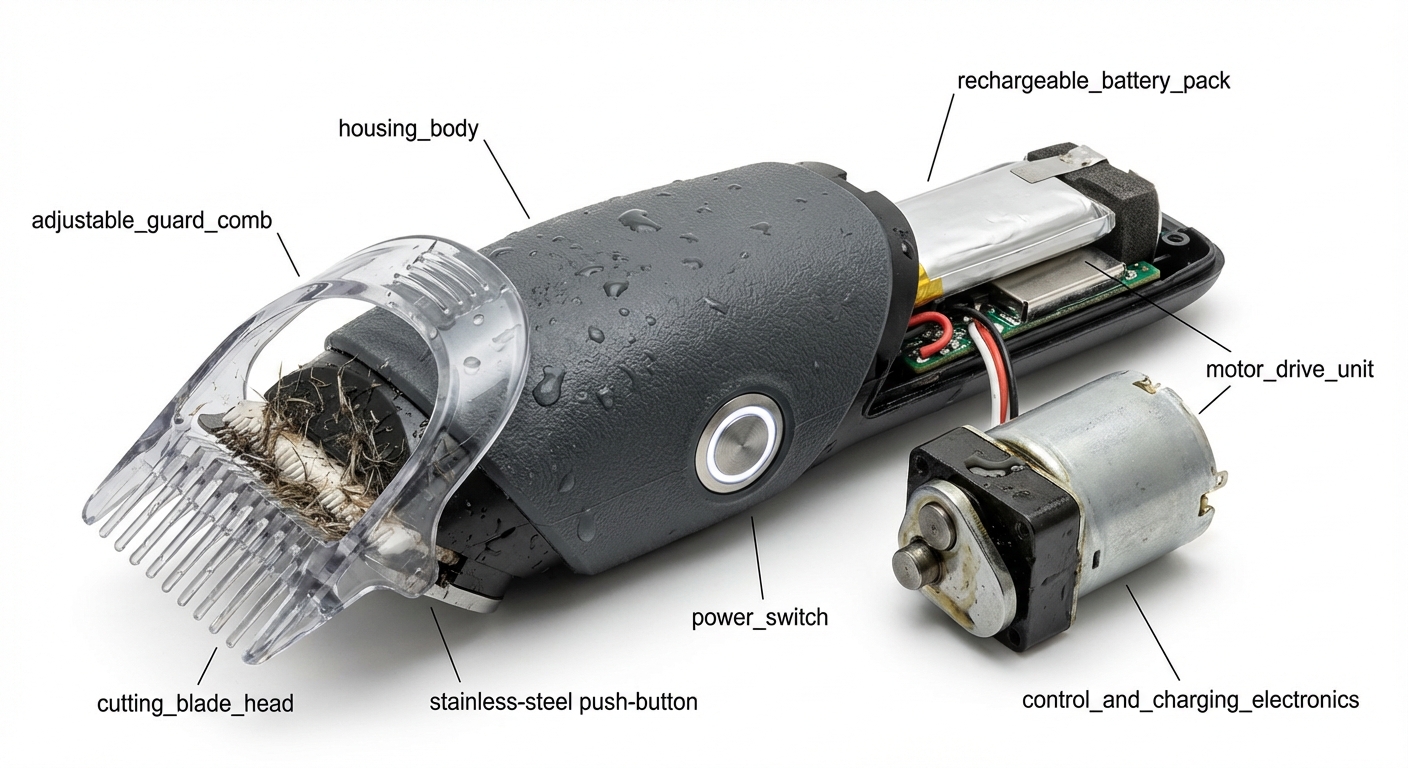}
\par}
\vspace{2pt}

\begin{lstlisting}[style=jsonTiny]
{
  "trace": "good_sft_dpo_4b",
  "round": 5,
  "model_response": {
    "reasoning": "The adjustable_guard_comb is made of clear plastic with fine, rigid teeth and is currently clogged with hair and gunk -- similar to the overflow slot. Its design allows it to be inserted into narrow spaces and its teeth can scrape or dislodge debris. The cutting_blade_head is sharp and designed for cutting, but it is not suitable for scraping gunk from a slot without leverage. ... The adjustable_guard_comb is the most promising candidate because its shape and material are ideal for loosening gunk in a confined space.",
    "action": "inspect_part",
    "part": "adjustable_guard_comb"
  },
  "feedback": {
    "type": "PART INSPECTION",
    "part": "adjustable_guard_comb",
    "entity": "electric beard trimmer with adjustable guard",
    "physical": "exact millimeter dimensions; integrated slider track, detent clicks, and side guide rail height-marking function; specific plastic type; durability/sturdiness; springiness/elasticity; low-friction property",
    "state": "visible; free"
  }
}
\end{lstlisting}

\vspace{2pt}
{\centering
\includegraphics[width=0.6\linewidth]{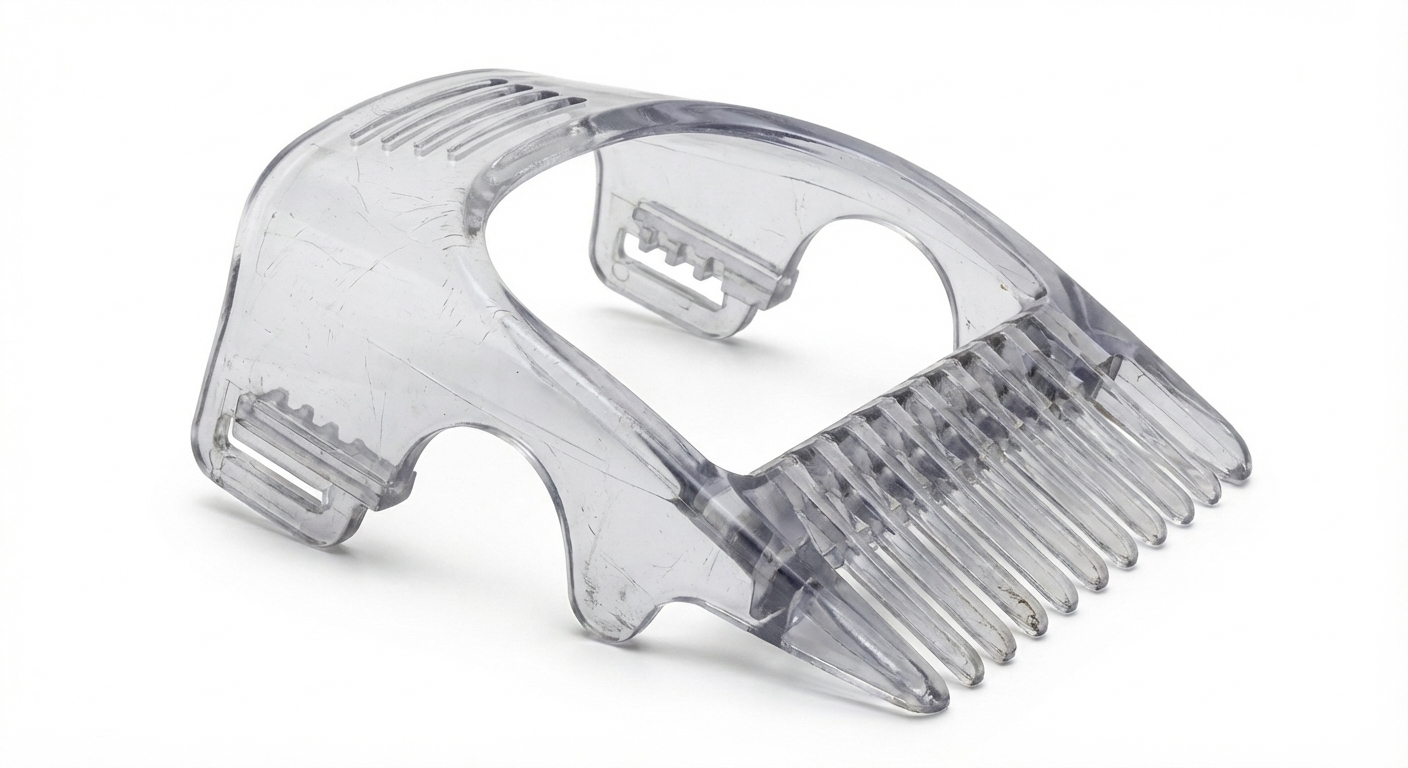}
\par}
\vspace{2pt}

\begin{lstlisting}[style=jsonTiny]
{
  "trace": "good_sft_dpo_4b",
  "round": 6,
  "model_response": {
    "reasoning": "The adjustable_guard_comb is made of durable, low-friction plastic with rigid, fine teeth designed to scrape and dislodge debris. Its shape is ideal for fitting into the narrow overflow slot, and its springiness allows it to flex and loosen gunk without breaking. Unlike the double-edge blade + handle, which requires manual leverage and may not fit snugly, the comb is specifically designed for such tasks and is free and visible. It is also non-abrasive and safe to use in a wet environment.",
    "action": "answer",
    "answer_entity": "electric beard trimmer with adjustable guard",
    "answer_part": "adjustable_guard_comb",
    "answer_how_to_use": "Insert the adjustable_guard_comb into the overflow slot and gently wiggle it back and forth to loosen the gunk. Use the fine teeth to scrape and dislodge the hair-and-soap buildup, then rinse with water to clear the slot."
  },
  "capability_shown": "exact-part discrimination under safety and state constraints"
}
\end{lstlisting}
\end{tcolorbox}

\section{Use of LLMs}
In this work, LLMs are used strictly for research support rather than as sources of substantive content. Their use falls into three categories: (i) serving as automatic pipeline annotation helper, (ii) providing tested results on MM-CreativityBench, and (iii) assisting with language refinement during paper writing. For writing support, we used ChatGPT solely to polish text (improving coherence and grammar) while all ideas, logic, results, and technical contributions originate from the authors. To safeguard rigor, we have carefully reviewed all LLM-refined texts to confirm that no hallucinated content was introduced and that the original arguments, findings, and perspectives were faithfully preserved.


\newpage

\end{document}